\documentclass[a4paper,fleqn]{cas-dc}

\usepackage[numbers]{natbib}
\usepackage{amsthm}
\usepackage{graphicx}

\usepackage{times}
\usepackage{soul}
\usepackage{url}
\usepackage{xcolor}
\usepackage{svg}
\usepackage{amsmath}
\usepackage{cleveref}
\crefformat{section}{\S#2#1#3} 
\crefformat{subsection}{\S#2#1#3}
\crefformat{subsubsection}{\S#2#1#3}
\usepackage[utf8]{inputenc}
\usepackage[small]{caption}
\usepackage{svg}
\usepackage[final]{pdfpages}
\usepackage{graphicx}
\usepackage{amsthm}
\usepackage{booktabs}
\usepackage{algorithm}
\usepackage{algorithmic}
\usepackage{booktabs}
\usepackage[colorinlistoftodos]{todonotes}
\usepackage[inline]{enumitem}
\usepackage{subcaption}

\newcount\Comments  
\newcommand{\kibitz}[2]{\ifnum\Comments=1\textcolor{#1}{#2}\fi}

\theoremstyle{definition}

\def\tsc#1{\csdef{#1}{\textsc{\lowercase{#1}}\xspace}}
\tsc{WGM}
\tsc{QE}
\tsc{EP}
\tsc{PMS}
\tsc{BEC}
\tsc{DE}

\begin{document}
\let\WriteBookmarks\relax
\def\floatpagepagefraction{1}
\def\textpagefraction{.001}
\shorttitle{Robust and Explainable Identification of Logical Fallacies in Natural Language Arguments}
\shortauthors{Sourati et~al.}

\title [mode = title]{Robust and Explainable Identification of Logical Fallacies in Natural Language Arguments}



\author[1,2]{Zhivar Sourati} 
\cormark[1]
\ead{souratih@isi.edu}

\author[1,2]{Vishnu Priya Prasanna Venkatesh}
\ead{vprasann@isi.edu}

\author[1,2]{Darshan Deshpande}
\ead{darshang@isi.edu}

\author[1,2]{Himanshu Rawlani}
\ead{hrawlani@isi.edu}

\author[1,2]{Filip Ilievski}
\cormark[1]
\ead{ilievski@isi.edu}

\author[3]{Hông-Ân Sandlin}
\ead{hongan.sandlin@ar.admin.ch}

\author[3]{Alain Mermoud}
\ead{alain.mermoud@ar.admin.ch}

\address[1]{Information Sciences Institute, University of Southern California, Marina del Rey, CA, USA}
\address[2]{Department of Computer Science, University of Southern California, Los Angeles, CA, USA}
\address[3]{Cyber-Defence Campus, armasuisse Science and Technology, Switzerland}

\cortext[cor1]{Corresponding author}

\begin{abstract}
The spread of misinformation, propaganda, and flawed argumentation has been amplified in the Internet era. Given the volume of data and the subtlety of identifying violations of argumentation norms, supporting information analytics tasks, like content moderation, with trustworthy methods that can identify logical fallacies is essential. In this paper, we formalize prior theoretical work on logical fallacies into a comprehensive three-stage evaluation framework of detection, coarse-grained, and fine-grained classification. We adapt existing evaluation datasets for each stage of the evaluation. We employ three families of robust and explainable methods based on prototype reasoning, instance-based reasoning, and knowledge injection. The methods combine language models with background knowledge and explainable mechanisms. Moreover, we address data sparsity with strategies for data augmentation and curriculum learning. Our three-stage framework natively consolidates prior datasets and methods from existing tasks, like propaganda detection, serving as an overarching evaluation testbed. We extensively evaluate these methods on our datasets, focusing on their robustness and explainability. Our results provide insight into the strengths and weaknesses of the methods on different components and fallacy classes, indicating that fallacy identification is a challenging task that may require specialized forms of reasoning to capture various classes. We share our open-source code and data on GitHub to support further work on logical fallacy identification.

\end{abstract}




\maketitle


\section{Introduction}

The purpose of constructing an argument is to prove conclusions that are in some way unknown or doubtful or that have been challenged and called into question~\cite{barker1965elements}. A \textit{logical fallacy} is a logical mistake in the reasoning used to transition from one proposition to the next, which results in a faulty argument~\cite{almossawi2014illustrated}. Logical fallacies form a broad category of violations of argumentation norms, including structure, consistency, clarity, order, relevance, and completeness. Detecting whether an argument is fallacious and the corresponding actual violation, is in practice a subtle task. Detecting one or more fallacies in an argument, however, does not prove its conclusion to be false - they merely detect a flaw in the reasoning that attempted to prove that the conclusion is true. 

Logical fallacies have been of interest to social science since the early days of mathematics and philosophy~\cite{Aristotle1989-jz}. 
More recently, the societal relevance of logical fallacies has been greatly amplified due to the wide adoption of the World Wide Web, which enabled a free exchange of large amounts of information, including an easy spread of misinformation \cite{wang2019systematic,wu2019misinformation,allcott2019trends} and propaganda~\cite{da2019fine,barron2019proppy,gundapu}. Misinformation and propaganda are thorny issues for social media platforms on the Web and have been increasingly addressed
through the growing teams of moderators~\cite{ganesh2020countering,morrow2022emerging}, and are under the scrutiny of different organizations and governmental bodies, such as the UN~\cite{khan_2021}. Similarly, the EU plans to ratify addressing misinformation as part of the Digital Services Act~\cite{eu-2022-2065}, 
as the spread of harmful and incorrect arguments can sway the population and lead to political shifts and civil unrests~\cite{lazer2018science}. 

Considering the \textit{subtlety} and the \textit{volume} of fallacious arguments, manually checking each by a human has become impossible. Moreover, the very \textit{subjective} nature of the tasks tends to open room for disagreement on the classification when multiple annotators or moderators are involved.
This motivates the need for automated methods that can quickly process an argument, understand its intent, and detect possible flaws in the reasoning. The algorithms need to be \textit{robust}, i.e., work well for an argument in an open domain, and \textit{explainable}, i.e., provide an explicit trace of their reasoning for human collaborators like social media moderators. Prior work on taxonomizing logical fallacies \cite{Copi1954-COPITL-6,Aristotle1989-jz,barker1965elements} and the initial efforts to develop logical fallacy benchmarks~\cite{logical_fallacy_main_paper} has set the ground for comprehensive and trustworthy logical fallacy methods. However, as these works have been attempted in isolation, comprehensive methods and tasks are lacking.

Building a comprehensive evaluation setup and methods for logical fallacy identification has several key challenges. First, while prior work has provided a list of taxonomies for organizing logical fallacies, it is unclear how they can be organized and aligned with existing benchmarks. Second, logical fallacies require an abstraction from syntax to high-level semantics revolving around structure and soft logic. This makes pure language model-based methods insufficient for fully solving the task. Third, arguments rely heavily on background factual and commonsense knowledge. A robust and explainable method needs mechanisms to make implicit (assumed) knowledge in fallacies explicit. Fourth, given the large set of fallacies and the relatively small amount of annotated examples for supervised learning, data sparsity is a serious issue. To build robust and explainable methods, it is essential to devise scalable mechanisms that can combat data sparsity.



In this paper, we consider the research question: \textit{How can we build methods for robust and explainable identification of logical fallacies in natural language arguments?} We consolidate prior work on taxonomizing logical fallacies into a three-stage framework of logical fallacy identification tasks, ranging from deciding whether there is a logical fallacy in an argument (logical fallacy detection), performing classification in high-level classes (coarse-grained classification), and finally performing classification into a wider range of specific classes (fine-grained classification). To deal with the need for abstraction and to fill knowledge gaps, we experiment with three families of methods: prototype-based reasoning, instance-based reasoning, and knowledge injection. We combat data sparsity through suitable methods for data augmentation and curriculum learning.

The contributions of this paper are as follows:
\begin{enumerate}
    \item We design a three-stage framework of logical fallacy identification tasks, inspired by fallacy classification theories. We map and enhance existing datasets into this pipeline to provide a well-motivated and representative evaluation set.
    \item Our framework includes
    a wide range of methods with a focus on robustness and explainability: prototype-based reasoning, instance-based reasoning, and knowledge injection. We complement these methods with strategies for distant learning from more data based on data augmentation and curriculum learning.
    \item We conduct an extensive evaluation of these methods on our datasets, focusing on their robustness and explainability. Our results provide insight into the strengths and weaknesses of the methods on different components and fallacy classes, indicating that fallacy identification is a challenging task that may require specialized forms of reasoning to capture various classes.
\end{enumerate}


The rest of this paper is structured as follows. A comprehensive study of different classification schemas on logical fallacies, together with our three-stage framework, is presented in Section \ref{sec:framework}. Prior work that detects logical fallacies or uses related methods to ours is reviewed in Section \ref{sec:related-work}. We describe the adopted methods in Section \ref{sec:method} and the experimental setup in Section \ref{sec:experimental-setup}. Our results accompanied by the extra ablation studies are presented in Section \ref{sec:results}. We discuss our findings and conclude the paper in Sections \ref{sec:discussion} and \ref{sec:conclusions}.

We make all our code and data available on GitHub at \url{https://github.com/usc-isi-i2/logical-fallacy-identification}.

\section{Organizing Logical Fallacies}
\label{sec:framework}


There are two broad categories of fallacies: 
\textit{formal}, involving the error in the logical structure of the argument, and \textit{informal}, mostly concerned with the content of the argument or the latent error in their expression of logic \cite{Gibbs2010-dp}. In this study, we focus on the latter.
Within informal fallacies, 
various definitions and categorizations of logical fallacies have been proposed since antique Greek philosophers such as 
\citet{Aristotle1989-jz}. Aristotle’s Sophistical Refutations \cite{Aristotle1989-jz} and John Locke’s An Essay Concerning Human Understanding \cite{Locke1997-co} can be considered as the cornerstones of works on logical fallacies, followed by notable contributions by others, especially Copi \cite{Copi1954-COPITL-6}, Barker \cite{barker1965elements}, and Watts \cite{Watts2021-an}. We elaborate on 
each of the aforementioned philosophical theories in Section~\ref{subsec:existing-theories}. Then, in Section~\ref{subsec:logical-fallacy-framework}, we devise our logical fallacy framework that is rooted in these philosophical theories, and it formalizes them into three stages: fallacy detection, coarse-grained classification, and fine-grained classification. We describe the coarse- and the fine-grained classes that constitute our taxonomy of logical fallacies. 


\subsection{Existing Theories of Categorization for Logical Fallacies}
\label{subsec:existing-theories}

\citet{Aristotle1989-jz}
distinguishes several kinds of deductions (syllogisms) in \cite{Aristotle1989-jz}. Broadly, he groups the fallacies into the ones dependent on language (\textit{In Dictione}) and the ones not dependent on language (\textit{Extra Dictionem}).
His categorization revolves around the premises discussed in the deductions as well as the conditions required for arguments to prove them correct. 
According to Aristotle, an argument satisfies three conditions, and “is based on certain statements made in such a way as necessarily to cause the assertion of things other than those statements and as a result of those statements.”
Thus an argument may fail to be a syllogism in three different ways: 
(1) the premises may fail to necessitate the conclusion, (2) the conclusion may be the same as one of the premises, and (3) the conclusion may not be caused by (or grounded in) the premises. 
Aristotle’s fallacies are primarily fallacious deductions that appear to be correct on the surface. 
There are six classes of fallacies dependent on language: \textit{Equivocation}, \textit{Amphiboly}, \textit{Combination of Words}, \textit{Division of Words}, \textit{Accent}, and \textit{Form of Expression}. Additionally, there are seven kinds of logical fallacies (\textit{sophistical refutation} in Aristotle's words) that can occur in the category of fallacies not dependent on language: \textit{Accident}, \textit{Secundum Quid}, \textit{Consequent}, \textit{Non-Cause}, \textit{Begging the Question}, \textit{Ignoratio Elenchi} and \textit{Many Questions}. In summary, Aristotle classifies fallacies into thirteen classes. 


\citet{barker1965elements} classifies logical fallacies based on the validity of the assumptions made when transitioning from premises to conclusions, as well as the validity of the premise and the conclusion themselves.
Barker defines validity as follows.
First, a valid argument would comprise premises that are all true. Second, it would not need the conclusions to satisfy their validity. And finally, its conclusions can be directly derived from the premises. 
This view is closely similar to Aristotle's, as well as the requirements that \cite{Ireneous_Nakpih_2020} have analogously proposed.
Neglect of the third requirement gives rise to the fallacies of \textit{Non Sequitur} that are fallacies that have an insufficient link between premises and conclusions. 
Neglect of the second requirement gives rise to fallacies of \textit{Petitio Principii} in which "the premises are related to the conclusion in such an intimate way that the speaker and his hearers could not have less reason to doubt the premises than they have to doubt the conclusion". 
Neglect of the first requirement gives rise to the remaining category of fallacies in which premises are present that are not necessarily true all at once, even if the link between premises and conclusions is as rigorous as can be. 
In summary, identifying fallacious arguments would boil down to analyzing the validity and soundness of the claims as well as the sufficiency and necessity of the premise of arguments to satisfy the needs of the conclusion to be true. There are three levels of classification proposed in \cite{barker1965elements} that, on the finest level, would sum up to twenty classes of fallacies, although his categorization allows for more as well and he does not argue for a bounded definition or particular number. 


\citet{Locke1997-co} can be credited with the contribution of \textit{Ad-Arguments}, which are arguments "that men, in their reasoning with others, do ordinarily make use of to prevail on their assent; or at least so to awe them as to silence their opposition." Locke discusses three kinds of such arguments: \textit{Ad Verecundiam}, \textit{Ad Ignorantiam}, and \textit{Ad Hominem}. According to him, these are not fallacies, but have been developed beyond his conception and have been named as such \cite{Goodwin1998}.
\textit{Ad Verecundiam}, or \textit{Appeal to Authority} is a fallacy when it is either on the ground that authorities (experts) are fallible or for the reason that appealing to authority is an abandonment of an individual’s epistemic responsibility \cite{sep-fallacies}.
\textit{Ad Ignorantiam}, or \textit{Appeal to Ignorance}, happens when one demands “the adversary to admit what they allege as a proof, or to assign a better.” In other words, the \textit{Ad Ignorantiam} fallacy happens when the argument claims a proposition to be true because there is no evidence against it.
According to Locke, \textit{Ad Hominem} was a way “to press a man with consequences drawn from his own principles or concessions.” That is, to argue that an opponent’s view is inconsistent, logically or pragmatically, with other things he has said or to which he is committed to \cite{sep-fallacies}.

\citet{Copi1954-COPITL-6} defines fallacies as “a form of argument that seems to be correct but which proves, upon examination, not to be so.” Copi discusses both deductive invalidities and inductive weaknesses as sufficient reasons for arguments to be fallacious. From the eighteen \textit{informal fallacies} he categorizes, eleven are borrowed from \cite{Aristotle1989-jz} and the other seven can be traced back to \cite{Locke1997-co}. He breaks down fallacies into \textit{formal fallacies} and \textit{informal fallacies}. With his definition over \textit{formal fallacies} pertaining to the deductive fallacies, he classifies \textit{Affirming the Consequent}, \textit{Denying the Antecedent}, \textit{The Fallacy of Four Terms}, \textit{Undistributed Middle}, and \textit{Illicit Major} as \textit{formal fallacies}. Focusing on the \textit{informal fallacies}, Copi defines two broad categories as \textit{Fallacies of Relevance} and \textit{Fallacies of Ambiguity}. \textit{Fallacies of Relevance} include \textit{Accident}, \textit{Converse Accident}, \textit{False Cause}, \textit{Petitio Principii}, \textit{Complex Question}, \textit{Ignoratio Elenchi}, \textit{Ad Baculum}, \textit{Ad Hominem Abusive}, \textit{Ad Hominem Circumstantial}, \textit{Ad Ignorantiam}, \textit{Ad Misericordiam}, \textit{Ad Populum}, and \textit{Ad Verecundiam}, while \textit{Fallacies of Ambiguity} include \textit{Equivocation}, \textit{Amphiboly}, \textit{Accent}, \textit{Composition} and \textit{Division}.

We conclude that the described categorizations \cite{Aristotle1989-jz,Copi1954-COPITL-6,barker1965elements,Locke1997-co} mostly agree on the definition of fallacious arguments as well as the broad categorizations of fallacies. 
The main difference lies in the fine-grained categorizations: \citet{Aristotle1989-jz} discusses the thirteen ways arguments can be fallacious, while \citet{Copi1954-COPITL-6} proposes eighteen different fallacy groups. \citet{barker1965elements} categorizes fallacies into twenty classes although he does not delineate the exact categorization or the number of classes, and all presumably borrow \textit{Ad Fallacies} from \citet{Locke1997-co}.
These discrepancies require computational approaches for logical fallacy identification to choose between the proposed theories.
For our experimental work, we adopt the broad categorization of \cite{Copi1954-COPITL-6}, and the fine-grained classification by \cite{hansen_2020} and \cite{logical_fallacy_main_paper}. We describe our categorization further in Section \ref{subsec:logical-fallacy-framework}.

\begin{figure*}[!t]
    \centering
    \includegraphics[width=0.7\linewidth]{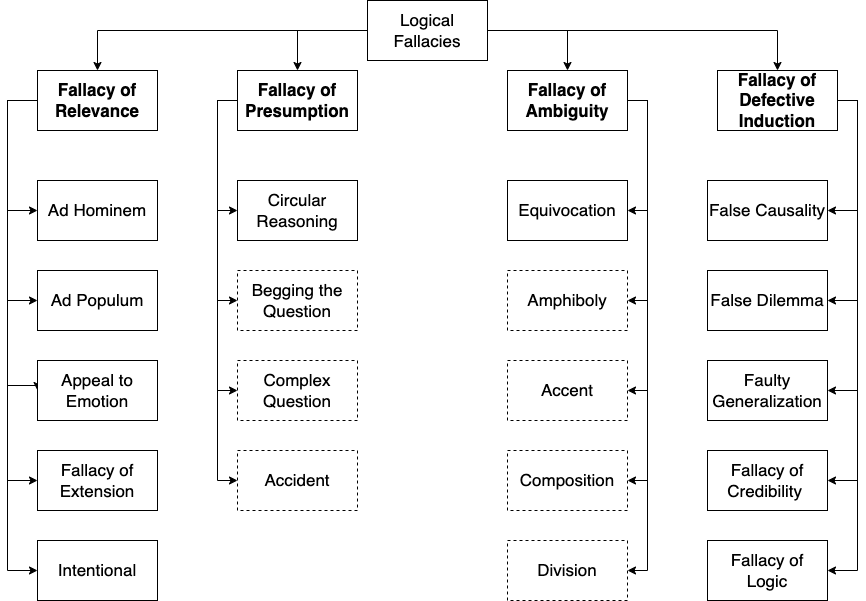}
    \caption{Three-stage taxonomy of logical fallacy identification. Coarse-grained classes are shown in boldface, while regular font is used to show fine-grained classes. We use solid and dotted boundaries to distinguish between fine-grained classes that we include and exclude in our experimental study, respectively.}
    \label{fig:taxonomy}
\end{figure*}

\subsection{Logical Fallacy Framework}
\label{subsec:logical-fallacy-framework}

We design a three-stage framework (Figure~\ref{fig:taxonomy}) as an overarching testbed for prior research on logical fallacies.
The first stage of the \textit{logical fallacy detection} aims
to identify whether a logical statement contains a logical fallacy or not. The detection is formalized as 
a binary classification task to identify the arguments that are logically fallacious in any sense. 
If a fallacy has been detected, the goal of the second stage is to categorize the fallacy into one of a few 
broad classes (e.g., \textit{Fallacy of Relevance}). 
In the third stage, the aim is to further
classify a fallacy into a fine-grained class (e.g., \textit{Ad Populum}).


\begin{table*}[!t]
\centering
\small
\caption{Examples for fallacious arguments belonging to different coarse-grained and fine-grained classes covered in our work.
}
\begin{tabular}{p{0.2\textwidth} | p{0.175\textwidth} | p{0.5\textwidth}}
\toprule
\textbf{Coarse-Grained Class} & \textbf{Fine-Grained Class} &  \textbf{Example}\\
\midrule
\textit{Fallacy of Relevance} & \textit{Ad Hominem} & Boris is not qualified to make suggestions about our penal system. As an ex-convict, he would always take the criminals’ side. \\

& \textit{Ad Populum} & Aliens must exist because most people believe in them.\\

& \textit{Appeal to Emotion} & Luke didn't want to eat his vegetables, but his father told him to think about the poor, starving children in a third world country who don't have anything to eat. \\

& \textit{Fallacy of Extension} & If you don't drive a car, you hate the Earth. \\

& \textit{Fallacy of Relevance} & I know you want to imprison me for having murdered my parents, but judge, have mercy on me, I’m an orphan! \\ 

& \textit{Intentional} & A woman decides to visit a certain doctor after only asking advice on the best doctors from ONE friend. \\ 

\midrule
\textit{Fallacy of Defective Induction} & \textit{False Causality} & The temperature has dropped this morning, and I also have a headache. The cold weather must be causing my headache. \\ 
& \textit{False Dilemma} & Subscribe to our streaming services, or get stuck with cable! \\ 
& \textit{Faulty Generalization} & My friend said her Math class was hard, and the one I’m in is hard, too. All Math classes must be hard! \\ 
& \textit{Fallacy of Credibility} & My uncle is a mechanic and he says you shouldn't spank children. He says it's ineffective. \\ 
& \textit{Fallacy of Logic} & Employees are like nails. Just as nails must be hit in the head in order to make them work, so must employees. \\
\midrule

\textit{Fallacy of Presumption} & \textit{Circular Reasoning} & Quinoa is a delicious, plant-based source of protein because it tastes so darn good. \\ 
\midrule
\textit{Fallacy of Ambiguity} & \textit{Equivocation} & The officer told me to freeze but it was too hot out to be freezing, so I was justified in running away. \\

\bottomrule
\end{tabular}
\label{tab:fallacy-examples}
\end{table*}

Following~\cite{Copi1954-COPITL-6}, we consider the following four coarse-grained classes:
\textit{Fallacy of Relevance}, \textit{Fallacy of Defective Induction}, \textit{Fallacy of Presumption}, and \textit{Fallacy of Ambiguity}. Figure~\ref{fig:taxonomy} shows the sub-categorizations we make from these coarse-grained classes to fine-grained classes described in~\cite{logical_fallacy_main_paper}. To perform the mapping, we use the definitions of fine- and coarse-grained classes given in~\cite{Copi1954-COPITL-6}. We next describe our fallacies in detail.

\textbf{\textit{Fallacy of Relevance}} occurs for arguments with premises that are logically irrelevant to the conclusion. 
\textit{Fallacy of Relevance} subsumes the fine-grained classes \textit{Ad Hominem, Ad Populum, Appeal to Emotion, Fallacy of Extension, Intentional Fallacy}. All of these fallacy classes present different means for using peripheral premises as support for claims. \textit{Ad Hominem} contains sentences where an attack over the subject acts as a premise for the claim made in those sentences, while \textit{Appeal to Emotion} involves manipulating the recipient's emotions to prove a claim. \textit{Ad Populum} involves affirming claims based on popular belief, and \textit{Fallacy of Extension} uses exaggeration for affirming claims based on the corresponding sentences. \textit{Intentional Fallacy} is directed towards using subconscious choices to incorrectly support an argument.


Within the broad class of \textbf{\textit{Fallacy of Defective Induction}}, the premises seemingly provide ground for the conclusion but upon analysis prove to be insufficient and weak for supporting the claim made. \textit{Fallacy of Defective Induction} is specified via five fine-grained categories, namely \textit{False Causality, False Dilemma, Faulty Generalization, Fallacy of Logic, and Fallacy of Credibility.} Arguments that jump to a conclusion without implying a causal relationship between the premise and the claim fall under \textit{False Causality}. If the specific causal relationship between the premise and the claim is generalized to a wider category of subjects, the argument is categorized as \textit{Faulty Generalization}. Arguments that cast doubt regarding the credibility of the subject making a claim constitute for \textit{Fallacy of Credibility}. When an argument presents a premise that erroneously limits the options available, it constitutes a \textit{False Dilemma}. When the logical construct of the argument is inaccurate and misleading, it constitutes a \textit{Fallacy of Logic}. 

\textbf{\textit{Fallacy of Presumption}} takes place when the inference to the conclusion depends mistakenly on unwarranted assumptions. \textit{Fallacy of Presumption} includes the following fine-grained classes. \textit{Circular Reasoning} occurs for arguments that come back to the beginning without proving themselves. Other classes that fall within \textit{Fallacy of Presumption} are: \textit{Begging the Question}, where the conclusion is treated like an assumption from the premise of the statement; \textit{Complex Question}, where the argument is framed as a loaded question that intends to prove another latent unproved assumption; and \textit{Accident}, where generalization is applied to specific cases that are out of scope. 

\textbf{\textit{Fallacy of Ambiguity}} 
occurs when words or phrases are used in an equivocal way, thus causing ambiguity in the logic that connects the premise and the conclusion.
The fallacy class \textit{Equivocation} is a \textit{Fallacy of Ambiguity} due to the presence of phrases in arguments that are used interchangeably in different parts of the sentence, leading to ambiguity in logic. Other classes in \textit{Fallacy of Ambiguity} include \textit{Amphiboly, Accent, Composition, and Division}. In the case of \textit{Amphiboly}, the usage of words that could be used interchangeably leads to a false interpretation in the grammatical construction of the sentences. \textit{Accent} fallacy is one, where a specific phrase or word carries a different contextual meaning in the premise and the conclusion. Mistaken inferences about parts of a whole argument for drawing inferences about attributes for that argument  constitute the \textit{Composition} fallacy.
\textit{Division} fallacy is the reverse of the \textit{Composition} fallacy, where mistaken inferences about the whole argument are used for drawing inferences about attributes of parts of it. 

We provide examples for each of the fine-grained and coarse-grained classes 
in Table \ref{tab:fallacy-examples}. A simplifying assumption we make in this work is that each fallacious argument belongs to exactly one broad class and exactly one fine-grained class. Prior work~\cite{da2019fine,logical_fallacy_main_paper} has shown that this assumption does not always hold, for example, \textit{“Drivers in Richmond are terrible. Why does everyone in a big city drive like that?”} as cited in \cite{logical_fallacy_main_paper}, is an example that belongs to \textit{Ad Hominem} but does have flavors of \textit{Faulty Generalization} as well. This gives room for arguments to be categorized into different fallacy classes simultaneously. Our simplifying assumption restricts our classification task to a multi-class task rather than a multi-label task. 

\section{Related Work}
\label{sec:related-work}

In this section, we review prior computational work on logical fallacy detection and the related task of propaganda detection. We also review related work that leverages the methods of case-based reasoning, knowledge injection, and curriculum learning.

\textbf{Logical Fallacy.}
Prior computational work formalizes arguments containing logical fallacies to make them suitable for ingestion by rule-based systems and theoretical frameworks.
\citet{gibson2007computational} formalize and identifies \textit{formal logical fallacies} using Argument Markup Language (AML) and discusses the theoretical questions that arise in the study of fallacy.  
\citet{Yaskorska2013} adopt a structure-aware approach to identify, include, and eliminate \textit{formal fallacies} in natural dialogues. 
\citet{Ireneous_Nakpih_2020} present a model that discovers \textit{non sequitur fallacies} in legal argumentation using Prolog language and check the validity, soundness, sufficiency, and necessity of argumentation using logical rules. These works mostly focus on \textit{formal fallacies}, which are defined in terms of their structure. In our work, we focus on \textit{informal fallacies}, 
whose detection and classification rely on linguistic and world knowledge. 

One of the few studies done on \textit{informal fallacies}~\cite{logical_fallacy_main_paper} proposes the task of logical fallacy detection, where arguments are classified into thirteen fine-grained fallacies.
This work evaluates the effect of using large pretrained language models on two datasets, called LOGIC and LOGIC Climate. Apart from using large pretrained language models, \citet{logical_fallacy_main_paper} try to abstract away from the surface of the arguments by exploiting coreference resolution and entity linking, in order to identify logical fallacies that are structurally fallacious in their arguments.
Similarly, \citet{goffredo2022fallacious} alongside presenting an annotated dataset of 31 political debates from the U.S. Presidential Campaigns, use transformer-based language models and process four parts of arguments, i.e., the dialogue context, argument components (premise and claim), fallacious argument snippet, argument relation (attack or support) separately, classify them, and train all the models jointly. They show that detecting argument components, relations, and context (see also \cite{sahai-etal-2021-breaking}) in debates is a necessary step to improve the model's performance. The main difference between \cite{goffredo2022fallacious} and our study is the fact that we do not need and use any context to classify logical fallacies. Furthermore, our framework does not assume any specific structure for text, and hence can be more generalizable. In our work, we reuse the dataset from \cite{logical_fallacy_main_paper}, and also extend its evaluation framework by: 
(1) introducing a binary detection and coarse classification stage,
(2) employing methods with robust properties to satisfy the needs of classification of logical fallacies that go beyond language understanding brought by vanilla language models, 
(3) adapting our methods with native explainability, and
(4) carrying out a more extensive set of experiments and analyses.


\textbf{Propaganda Detection.}
Recent research has developed benchmarks and techniques for propaganda detection in natural language documents. A significant portion of these works focuses on extracting better features as well as novel methods that would help the model boost its performance \cite{oliinyk2020propaganda,https://doi.org/10.48550/arxiv.1909.06162,https://doi.org/10.48550/arxiv.2009.05289,kiesel-etal-2019-semeval,vorakitphan-etal-2021-dont}. There has also been a surge focusing on the interpretability of models in propaganda detection \cite{https://doi.org/10.48550/arxiv.2108.12802,yoosuf-yang-2019-fine,ferreira-cruz-etal-2019-sentence}.
\citet{dimitrov-etal-2021-detecting} show that propaganda techniques function as shortcuts in the argumentation process that connect to the emotions of the audience and often include logical fallacies. In \cite{hamilton2021towards}, logical fallacies are called "hallmarks of propagandist messaging", which implies that logical fallacies can be seen as components within the broader task of propaganda detection. However, as pointed out by \citet{logical_fallacy_main_paper}, the two tasks overlap but are distinct, since propaganda detection focuses on arguments that aim to influence people's opinions often using misinformation as a tool~\cite{Luceri2020,Jiang2020}, while logical fallacy detection aims to understand gaps in argumentation. There is also a practical difference between the formalization of these two tasks, as propaganda detection data has typically focused on longer input documents, while logical fallacy datasets have generally relied on focused and isolated text inputs. In our study, we utilize 
the overlap between some of the propaganda techniques and fallacy classes, by augmenting the training data for logical fallacy classification with a dataset gathered explicitly around propaganda detection~\cite{https://doi.org/10.48550/arxiv.2007.08024}. 



\textbf{Case-Based Reasoning.}
The case-based reasoning framework 
has been used to learn from past experiences explicitly in medical applications \cite{OYELADE2020100395,pantazi2004case} and mechanical engineering \cite{bardasz1993dejavu,qin_regli_2003}.
One of the most important aspects of case-based reasoning is its inherent interpretability. \citet{8776909} use case-based reasoning as an interpretation model for Word Sense Disambiguation, while \citet{bruninghaus2006progress} apply case-based reasoning to predict the outcome of legal cases. \citet{https://doi.org/10.48550/arxiv.2009.06349,Ge_Mao_Cambria_2022,https://doi.org/10.48550/arxiv.2209.07494} advocate for the increase in comprehension of the black-box models and their explainability as well as transparency using example-based explanations by the end-users. Similar to our work, \citet{spensberger2022effects} explore the effect of case-based reasoning on the student social workers and their fallacy recognition abilities and find that those who have access to worked examples perform better during the experiment. In this paper, we adopt two complementary case-based reasoning methods.
First, we adopt the instance-based reasoning method proposed by
\cite{sourati-cbr} that enriches the inputs with similar cases and with different case enrichments (e.g., based on counterarguments), and evaluates the impact of different modeling decisions and case representations on the model performance. We apply this method to our three-stage evaluation framework and perform further 
ablation studies to understand its performance in relation to modeling decisions and against other systems. Second, we include a prototype-based reasoning method, that maps novel examples to 
prototypical ones to classify logical fallacies.
With both of these methods, we use case-based reasoning both as a means to enhance the performance of our model and simultaneously as a proxy to explain the behavior of the model classifying logical fallacies.

\textbf{Knowledge Injection.}
The challenge of generalizability and transferability for logical fallacy classifiers has been discussed in \cite{logical_fallacy_main_paper}, by testing the model on a dataset containing unseen domain-specific subjects. This motivates the need for the injection of background knowledge.
Injection of background, especially commonsense knowledge in language models has been proposed within tasks of multiple-choice question answering. Combining neural language models with commonsense knowledge graphs (KGs) like ConceptNet~\cite{speer2017conceptnet} or ATOMIC~\cite{sap2019atomic} can be done by lexicalizing knowledge into task-targetted evidence paths and combining them with the task input~\cite{ma-etal-2019-towards,mitra2019exploring}. The idea in K-BERT \cite{k-bert} is similar - here a multi-head attention layer is used to combine evidence from background knowledge and the input task. 
Other forms of knowledge injection have been popular as well, such as using graph and relation networks~\cite{lin-etal-2019-kagnet,10.1007/978-3-030-32233-5_2}, or introducing the entire KG 
at training time regardless of the task at hand~\cite{peters-etal-2019-knowledge,ma2021knowledge}.
Notably, prior work has shown that the impact of the injected knowledge strongly depends on the overlap between the knowledge in these graphs and the downstream question answering task~\cite{ma2021knowledge,ilievski2021dimensions}. 
Due to the nature of logical fallacies, they can cover daily-life matters and events spreading throughout social media, and this calls for domain-specific knowledge for comprehension of certain logical fallacies. However, to our knowledge, exploiting external knowledge has not yet been fully explored in logical fallacy detection.
Trying to fill in the gap and utilize commonsense knowledge in the detection of logical fallacies, we use \cite{k-bert} to incorporate knowledge from arbitrary knowledge bases and benefit from potential enhancements.

\textbf{Curriculum Learning.}
Curriculum learning has been proposed in \cite{curriculum_learning_bengio} and \cite{doi:10.1046/j.0963-7214.2003.01267.x} from the computer science and psychology perspectives respectively. The key idea of curriculum learning is that starting from simple examples and learning from examples in an organized and meaningful way can contribute positively to the learning process. Using pure language model-based methods does not suffice for a reliable classification of logical fallacies~\cite{logical_fallacy_main_paper}, due to known issues of robustness and induction capabilities of vanilla language models on unseen data~\cite{https://doi.org/10.48550/arxiv.2106.11533,https://doi.org/10.48550/arxiv.2109.02837}. This motivates us to leverage continual curriculum learning to attempt to improve the convergence and robustness capabilities of models, an idea that has not yet been explored in logical fallacies. The application of curriculum learning to logical fallacies in our work is facilitated by the availability of datasets at different granularity levels. 

\begin{figure*}[!t]
    \centering
    \includegraphics[width=0.7\linewidth]{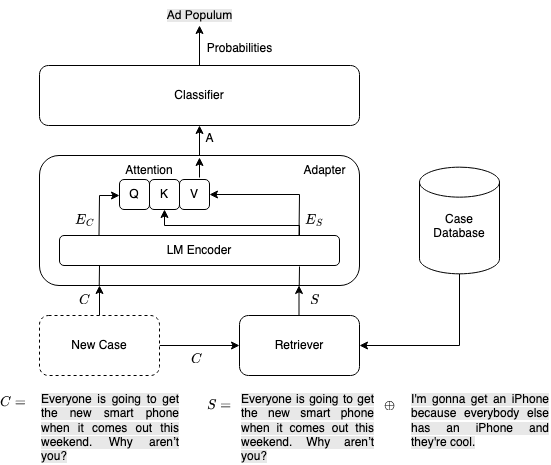}
    \caption{Three stages of the IBR pipeline. 
    Using the new Case $C$, retriever finds $k$ similar examples $\{ S_1, S_2, ..., S_k \}$, and creates $S = C\; \oplus <SEP> S_1 \oplus S_2 \oplus ... \oplus S_k $. The adapter encodes these two inputs and tries to adapt $S$ based on the new case $C$. Finally, the classifier uses the rectified information from the adapter to classify the new case by outputting the probabilities corresponding to belonging to each class of fallacies (in the example shown above, $k = 1$).
    }
    \label{fig:cbr}
\end{figure*}

\section{Method}
\label{sec:method}

Due to the difficulty, as well as the contention over the categorization and classification of logical fallacies \cite{hansen_2020}, we use methods that humans usually adopt when faced with problems that require complex reasoning.
According to \cite{10.2307/1165776,sciencedaily_2009,ROSCH1973328}, people use similar or prototypical examples of a situation or problem to solve or approach a new one.
The alluded similarity can be in the various levels, namely, coarse-grained features such as the whole argument or statements, but also in the more fine-grained features and in terms of the extra knowledge one might have about concepts or entities discussed in the sentences as discussed by \cite{arora-etal-2022-metadata}. 
Having in mind the simplicity as well as explainability of using similar examples or experiences to reason about and solve new problems or situations, we adapt methods for Instance-based Reasoning, Prototype Learning, and Knowledge Injection (\cref{subsec:method-neuro-symbolic-reasoning}). 
Another approach that humans follow for learning how to solve problems is starting from easy or simpler tasks and gradually shifting to harder ones to learn \cite{doi:10.1046/j.0963-7214.2003.01267.x}, which has been shown to work even better than other learning strategies by \citet{https://doi.org/10.1111/1467-9817.12022}.
This has been shown to be the case for neural networks as well \cite{ELMAN199371}, not as a barrier, but as a way of training more robust models referred to as Curriculum learning (\cref{subsec:method-cl}). 
Finally, we devise data augmentation strategies to address data sparsity and improve the stability of our models \cite{https://doi.org/10.48550/arxiv.1604.04326} (\cref{subsec:method-da}).

\subsection{Explainable Reasoning Methods}
\label{subsec:method-neuro-symbolic-reasoning}

\subsubsection{Instance-Based Reasoning}
\label{subsubsec:method-ibr}

Instance-based reasoning (IBR) \cite{daelemans_van_den_bosch_2005} is the process of solving new problems based on the solutions of similar past problems \cite{Aamodt1994}. IBR is reported to resemble the way humans think and approach new problems to save time and effort instead of starting from scratch \cite{10.2307/1165776}. IBR is a formalization of the general idea of Case-based reasoning (CBR) \cite{Aamodt1994}. Within CBR, rather than comparing new problem instances with instances seen before like in IBR, we use past similar problems and experiences and attempt to perform explicit generalization or induction.\footnote{We cover another variant of CBR, prototype theory, in \cref{subsec:method-pl}.} 


IBR starts with a set of cases or training examples; it forms generalizations of these examples, albeit implicit ones, by identifying commonalities between a retrieved case and the target case, and tries to approach the new case using known solutions to past cases. 
Our IBR formulation (Figure \ref{fig:cbr}) follows the three-stage pipeline proposed by~\cite{sourati-cbr} consisting of:
(1) \textit{Retriever} - given a target problem, retrieve similar cases with known solutions from memory, 
(2) \textit{Adapter} - adapt the retrieved similar cases to help the decision on the new case, and
(3) \textit{Classifier} - classify the new case based on the adapted exemplars. 
The last step in this pipeline corresponds to two steps in the formulation by \cite{Aamodt1994}: classify the new case based on the previous examples, and retain the new problem alongside its adapted solution and resulting experience in memory for later use in a more explicit way. 
We next describe the design of the retriever, the adapter, and the classifier.

\textbf{Retriever} 
is responsible for finding similar cases $S_i$ to the new case $C$ from a database and passing them to the adapter together with the new case ($S = C \;\oplus <SEP> S_1 \oplus S_2 \oplus ... \oplus S_k $ extracting $k$ similar examples). The retriever uses language model encoders to get the feature vectors for each new case as well as all the previous cases in the retriever database and uses these features to compute their cosine similarity.\footnote{We also experiment with encoding the input examples as either AMR graphs~\cite{banarescu2012abstract}, using explanation graphs~\cite{saha2021explagraphs}, or their combination, however, we do not pursue this direction further due to poor performance and explainability.} The retriever obtains the $k$ most similar examples from the database, which are then passed on to the adapter module.

We experiment with SimCSE \cite{https://doi.org/10.48550/arxiv.2104.08821}, a Transformer-based retriever that is optimized for capturing overall sentence similarity using a contrastive loss function. We also include sentence encoders that are reportedly able to manipulate a wide range of concepts, by using Sentence-BERT \cite{reimers-2019-sentence-bert} based on MiniLM \cite{wang2020minilm}.
We also include Transformer models that have been trained to distinguish emotional expressions, since it has been shown that emotions can be used to manipulate masses \cite{Bernays2004-fp} and they are intuitively important to detect certain logical fallacies, such as \textit{Appeal to Emotion}. To capture the usage of empathetic and emotional terminology, we use a RoBERTa model \cite{https://doi.org/10.48550/arxiv.1907.11692} fine-tuned on the WASSA 2022 Shared Task dataset \cite{barriere-etal-2022-wassa}. 




\textbf{Adapter} transforms the retrieved cases $\{S_1, S_2, ..., S_k\}$ together with the new case $C$ denoted as $S$ as well as the new case $C$, and prioritizes earlier cases that are most helpful. The adapter consists of two parts: an encoder and an attention mechanism. As an encoder, we use a language model that takes as input $C$ and $S$ and produces a set of raw hidden states $E_C$ and $E_S$ respectively without a head layer on top. 

\begin{figure*}[!t]
    \centering
    \includegraphics[width=0.8\linewidth]{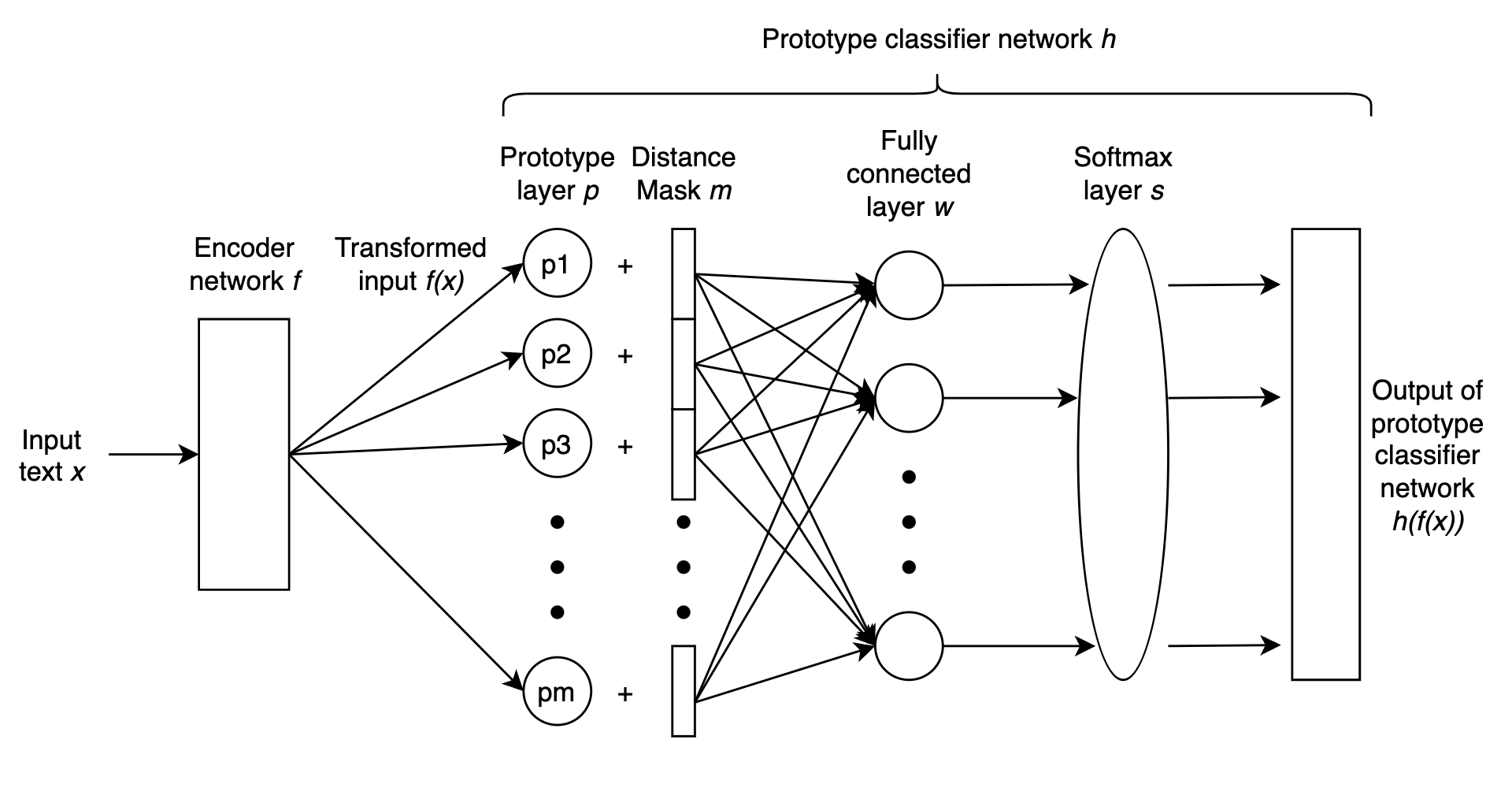}
    \caption{Our PBR method, using an adaptation of the Prototex architecture.}
    \label{fig:prototex}
\end{figure*}

The attention mechanism selects the most important information to be considered from similar cases. 
Based on the second step of the pipeline by \cite{Aamodt1994}, after the similar cases are retrieved, some of these similar cases should be manipulated or adapted to help the classifier at the end of the pipeline, since not all similar cases will be equally helpful for the model. 
We formalize this step with an attention mechanism on top of the encoded cases ($E_{S}$ and $E_{C}$) to filter the retrieved cases or shift the attention to where it helps the model best to reason about new cases. 
More concretely, we use a Multi-headed attention component \cite{https://doi.org/10.48550/arxiv.1706.03762} that fetches the \textit{new case} embedding $E_{C}$ as the query and the combined embeddings $E_{S}$ as both keys and values. We include both the new case as well as similar cases in $S$ to avoid losing information from the new case. The output of this component, i.e., the attention output $A$ has the same shape as $E_{C}$ and $E_{S}$ and is fed to the last step of IBR, i.e., the classifier. 

\textbf{Classifier} layer at the end of the pipeline is applied on top of the adapter output $A$ to predict the labels. As a classifier, we use a two-layer perceptron with a $gelu$ \cite{https://doi.org/10.48550/arxiv.1606.08415} activation function. Given a number of classes $C$, we compute $C$ logits and their corresponding probabilities of belonging to each class $c$. We use cross-entropy loss as our learning objective. 

Overall, the IBR method is similar to a language model with a classification head on top with an important distinction. By using a retriever and finding similar examples to the new case and integrating these new examples in the classification process, we benefit in two ways:
(1) we use similar examples of an argument to help the model classify the argument more accurately, 
and simultaneously, (2) enhance the explainability of the model, showing the end-users similar examples of an argument to lift end-users' understanding of the capabilities and acquired knowledge of the model \cite{https://doi.org/10.1111/cogs.12086}.

\subsubsection{Prototype-Based Reasoning}
\label{subsec:method-pl}

Prototype theory \cite{ROSCH1973328} is a theory of categorization in psychology and cognitive linguistics, in which there is a graded degree of belonging to a conceptual category, and some members are more central than others. In prototype theory, any given concept in any given language has a real-world example that best represents this concept, i.e., its \textit{prototype}. Like IBR, prototype-based reasoning (PBR) is also an instance of case-based reasoning, and
there has been some controversy about the superiority of one over the other. There are both claims about the superiority of prototypical examples over normal examples \cite{Johansen2005-nu}, as well as their counterparts \cite{Medin1978-je} who state that a context theory of classification, which derives concepts purely from exemplars works better than a class of theories that included prototype theory (\cref{subsec:results-analysis-of-method-performance}).

We build on the deep learning adaptation of the prototype theory by the Prototex \cite{das-etal-2022-prototex} method. 
The architecture of Prototex is shown in Figure~\ref{fig:prototex}. 
Prototex is based on the Prototype Classification Network proposed in \cite{https://doi.org/10.48550/arxiv.1710.04806}. The Prototex architecture contains an encoder \emph{f} and a special prototype layer \emph{p}, where each unit of that layer stores a weight vector that resembles a prototypical example. The prototype layer includes both positive and negative prototypes, aiming to help the models distinguish between the presence and absence of features that support any given class. 
The input \emph{x} is first encoded into a latent representation that is shared between the input data and the prototype layer \emph{p}. 
This representation is used to calculate the euclidean distance with the prototype layer \emph{p}, resulting in a distance vector $d$. We mask the distance vector with a distance mask layer \emph{m}. The role of the distance mask \emph{m} is to make the model only optimize the proximity of input examples of a particular class to a fixed set of prototypes. In other words, the distance mask directs the prototypes to represent prototypical examples of a particular class instead of a mixture of arbitrary classes. 
The masked distance vectors between input examples and prototypes are further fed to a fully connected layer \emph{w} followed by a softmax layer \emph{s} to classify a particular data point. To have interpretable prototype vectors, the model is optimized with auxiliary loss terms that bring the embeddings of the training examples closer to the prototypes and also the embeddings of the prototypes closer to the input examples. 

The Prototex method was originally designed for binary classification between propagandistic and non-propagandistic sentences.
We modify the Prototex architecture to support a multi-class classification setup. 
Moreover, the original architecture uses a sequence-to-sequence model, BART~\cite{lewis2019bart}. For a fair comparison to our other methods and inspired by the best results on logical fallacy reported in~\cite{logical_fallacy_main_paper}, we replace the BART encoder model in Prototex with a self-supervised language model, Electra~\cite{electra}.
We do not use the decoder network and instead focus on the learned prototypes and their explanations. 

\begin{figure*}[!t]
    \centering
    \includegraphics[width=0.7\linewidth]{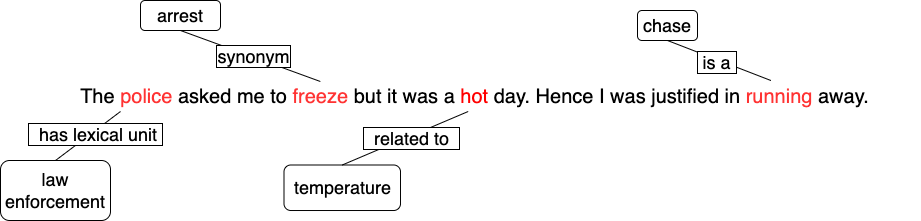}
    \caption{Example Sentence Tree Construction in K-BERT.}
    \label{fig:sentence-tree-example}
\end{figure*}

\label{subsubsec:method-KI}
\begin{figure}[!t]
    \centering
    \includegraphics[width=\columnwidth]{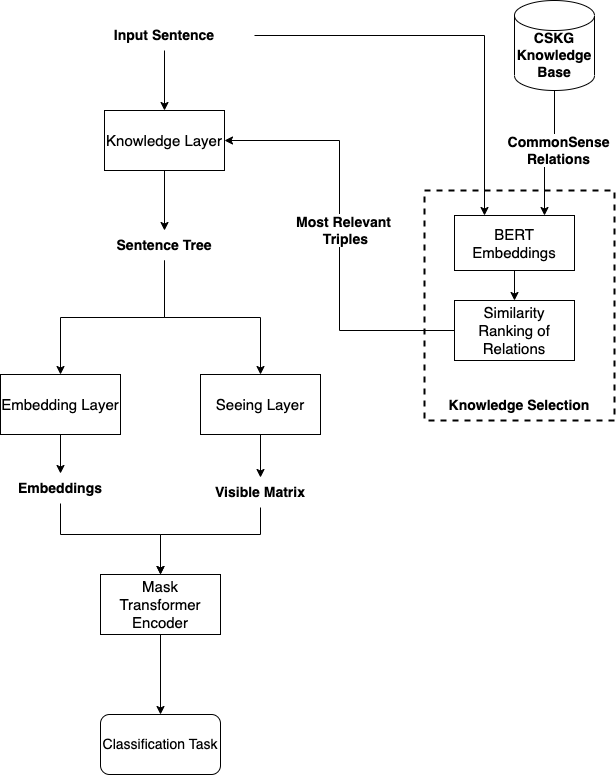}
    \caption{Our Knowledge Injection architecture, which is an adaptation of the K-BERT method.}
    \label{fig:kbert}
\end{figure}
 

\subsubsection{Knowledge Injection}

Many fallacy classes rely on the ambiguous structure of the logical construct in sentences to introduce flaws in arguments. Let us consider the example sentence \textit{The police asked me to freeze, but it was a hot day. So I was justified in running away}, which belongs to the fallacy class \textit{Equivocation} (Figure~\ref{fig:sentence-tree-example}). Here, the word \textit{freeze} is used in two contexts, one for where the \textit{police asked to freeze} and another, where the antonym of \textit{freeze}, i.e, \textit{hot} is used in the sentence. Such sentences, with latent fallacies, illustrate the need for models to have access to commonsense knowledge. 

We propose a knowledge injection (KI) formulation, where background commonsense knowledge is combined with the original input for the language model.
We adopt a popular method for injecting background knowledge in language models, called K-BERT \cite{k-bert}. 
K-BERT introduces knowledge injection to a BERT~\cite{devlin2018bert} model by querying a structured knowledge base. This knowledge base consists of a set of triples of the form (\textit{subject, predicate, object}). In the first layer, i.e., the knowledge layer, triples from the knowledge base are connected along with the tokens of the sentences, forming a sentence tree, as illustrated in Figure~\ref{fig:sentence-tree-example}. 
The embedding layer of K-BERT flattens out the sentence tree by retaining the structural information in the form of a visible matrix. As stated in \cite{k-bert}, a crucial goal of K-BERT is to prevent false semantic changes to the original sentence due to the addition of sentence trees from the knowledge base. 
K-BERT functions similarly to BERT~\cite{devlin2018bert} but uses a masked self-attention mechanism. The masked self-attention mechanism takes the visible matrix calculated by the seeing layer and ensures that the knowledge branches are not isolated from the tokens they are associated with and do not change the context of the general sentence that they are connected to. The classification task in K-BERT uses the Masked Language Modeling objective. 


Our KI adaptation of the K-BERT method focuses on the input of the knowledge layer, as shown in Figure~\ref{fig:kbert}. We adapt K-BERT to leverage knowledge from the Commonsense Knowledge Graph (CSKG)~\cite{ilievski2021cskg}, which consolidates commonly used public commonsense sources like ConceptNet~\cite{speer2017conceptnet}, ATOMIC~\cite{sap2019atomic}, and WordNet~\cite{miller1995wordnet}. The information in CSKG is structured as (\textit{subject, relation, object}) triples. To link to these triples, we extract all non-stopword tokens from the sentences as individual words and we match them with triples in CSKG where the words act as subjects. 


Since CSKG contains multiple relations associated with the same subject, a key question is how to prioritize or select relations (triples) that are most relevant and informative for the input sentence. Following \cite{ma2021knowledge}, we only use the 14 highly semantic relations in CSKG, namely \textit{'Causes', 'UsedFor', 'CapableOf', 'CausesDesire', 'IsA', 'SymbolOf', 'MadeOf', 'LocatedNear', 'Desires', 'AtLocation', 'HasProperty', 'PartOf', 'HasFirstSubevent', 'HasLastSubevent'}. Furthermore, we add a Similarity Ranking component, which ranks the retrieved triples according to their relevance to the original sentence. To do so, we estimate the contextual similarity of the triple to the original sentence by using the cosine similarity of their BERT \cite{https://doi.org/10.48550/arxiv.1706.03762} embeddings as a proxy. The cosine similarity is directly used to order the triples in order of priority. The triples with the highest similarity are injected into the original sentence, thus enriching it with commonsense knowledge. In our experiments with KI, we investigate the impact of the width and depth of the knowledge retrieval procedure. For this purpose, we test different \textit{branching factors} ($b$), representing the maximum number of obtained relations per subject, and different \textit{numbers of hops}, representing the path length between the subject and the subsequent relations discovered iteratively based on the entities of the previously discovered relation. 

\begin{figure}[!t]
    \centering
    \includegraphics[width=\linewidth]{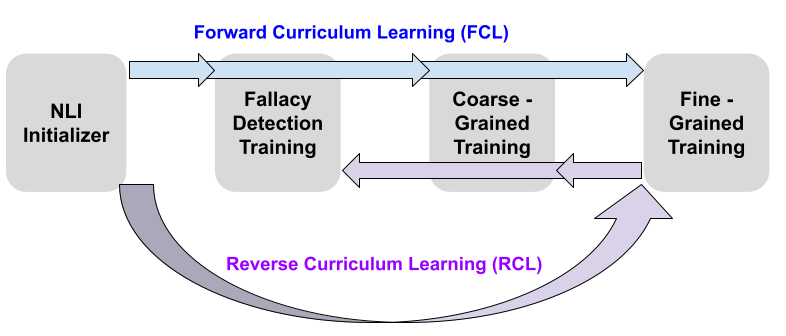}
    \caption{Three-stage curriculum pipelines for Forward Curriculum Learning and Reverse Curriculum Learning.}
    \label{fig:cl_pipeline}
\end{figure}

Returning to our example in Figure~\ref{fig:sentence-tree-example}, we see that the knowledge derived from CSKG helps in providing the context for the word \textit{freeze}, as a \textit{synonym} for arrest. Similarly, background knowledge tells us that the word \textit{hot} is \textit{related to temperature}. On the surface, the words \textit{freeze} and \textit{hot} seem to be used in the same context, but the background information from the knowledge base helps in indicating that they are based on two completely different contexts. The background knowledge for \textit{police} also bolsters that the usage of the word \textit{freeze} was intended for an arrest. This additional knowledge helps in identifying the ambiguous usage of words and connects the terms based on making implicit knowledge explicit. As the additional context (arrest, law enforcement) is not directly connected, we rely on the ability of the BERT LM to estimate the contextual similarity between these terms.
Thus, the combination of CSKG and LMs would lead to the classification of the logical fallacy in the sentence as one of \textit{equivocation}.








\subsection{Curriculum Learning with Language Models}
\label{subsec:method-cl}

\textit{Curriculum learning (CL)} \cite{curriculum_learning_bengio} is a strategy that exploits the varying complexity across ordered tasks in a pipeline to increase performance. CL uses previously learned concepts in the task pipeline and applies this information to more complex tasks in the latter half of the pipeline. We follow prior work~\cite{Rohde1999} to formulate two variants of CL (Figure~\ref{fig:cl_pipeline}). Our Forward Curriculum Learning (FCL) strategy exposes the model to increasingly demanding tasks, similar to how humans learn concepts. We also experiment with the inverse strategy of Reverse Curriculum Learning (RCL), which starts with a difficult task and gradually adapts the model for increasingly easy tasks. 

\textbf{Forward Curriculum Learning (FCL)}
For FCL, we primarily experiment with continuous training of Transformer language model variants. 
We try to induce fallacy knowledge in a discrete, three-stage curriculum pipeline, going from the simplest (binary fallacy detection) to the most complex (fine-grained classification) tasks. Through the binary classification stage, we aim to introduce the structural and topical knowledge required to identify fallacies in arguments. 
The model uses this information in the subsequent (coarse-grained) stage to learn about the broad categories of fallacies. These learned coarse representations are then transferred to and trained further on the fine-grained fallacy classification objective. 

\textbf{Reverse Curriculum Learning (RCL)} 
\citet{Rohde1999} discovered that learning from simple to complex examples is sometimes not as effective as learning complex patterns directly first. Although they revised their claims in a subsequent paper \cite{rohde2004less}, we explore the capabilities of the models trained with a reverse curriculum, i.e., moving inversely from complex to simple examples, which allows us to compare the different curriculum learning strategies for the task of logical fallacy identification. For RCL, we first train on the fine-grained classes and use these weights for the coarse-grained classification task. We ultimately test their applicability on the binary fallacy detection task. 



\begin{table*}[!t]
\centering
\caption{Augmentation examples.}
\begin{tabular}{p{0.2\textwidth} | p{0.21\textwidth} | p{0.5\textwidth}}
\toprule
\textbf{Original Sentence} & \textbf{Augmentation Method} &  \textbf{Augmented Sentence}\\
\midrule

Even without watching the movie, I just know that it would not be as good as the book. 
& WordNet & \textit{Yet} without watching the \textit{picture show}, I just \textit{make love} that it would not be as good as the book. \\
& Word2Vec & Even without watching the \textit{moive}, I just know that it \textit{could} not be as good \textit{regarded} the book.\\
& RoBERTa & Even without \textit{viewing} the movie, \textit{you} just knew that it would not be as good as the book. \\
& Backtranslation (DE-EN) & Even without \textit{seeing} the \textit{film}, \textit{all I} know \textit{is that} it \textit{wouldn't} be as good as the book. \\

\midrule

The news is fake because so much of the news is fake.
& WordNet & The news \textit{be} fake because so much of the \textit{word} is fake. \\
& Word2Vec & The news \textit{becomes} fake \textit{anyway} so much of the news is \textit{bogus}.\\
& RoBERTa & The \textit{data} is fake because so much about the \textit{information} is fake. \\
& Backtranslation (DE-EN) & The \textit{messages} \textit{are} fake because so \textit{many} \textit{messages} \textit{are} fake. \\

\bottomrule
\end{tabular}
\label{tab:augmentation-examples}
\end{table*}

\subsection{Data Augmentation}
\label{subsec:method-da}

Besides curriculum learning, we experiment with using data augmentation for addressing data sparsity. We devise two data augmentation strategies: modifying the original task data and adapting related benchmarks.

\textbf{Augmentation by Modifying the Original Task Data.} 
We apply commonly used text augmentation techniques for improving the performance and enhancing contextual understanding for logical fallacy detection and classification. We begin with a basic WordNet~\cite{miller1995wordnet} similarity-based augmentation. This involves using the synsets to substitute the words in the input with words that have the closest meaning according to the synset. Second, we evaluate word embedding substitution methods based on Word2Vec and transformer embeddings. These substitutions involve finding word vectors that are closest to the input word vector in the embedding space and replacing them. Lastly, we experiment with a more recent technique of back-translation, popularized by \cite{backtranslation} and originally proposed by \cite{backtranslation_at_scale}. This involves translating the input sentence into a language that is syntactically and morphologically dissimilar and subsequently reverse-translating this translation back to the original language. To select languages, we follow the insights from prior work \cite{backtranslation,backtranslation_at_scale,sennrich-etalbacktranslation}.
As the parental tree for a language must be analyzed, languages that have fewer cognates are preferred as they enhance variety. Additionally, the use of two translation models trained on different datasets has been found to usually work better and provide more diversity to the output sentence. The most popular choices for back-translation model pairs are German $\leftrightarrow$ English, Turkish $\leftrightarrow$ English, and French $\leftrightarrow$ English. 

Table~\ref{tab:augmentation-examples} shows representative examples of the obtained augmentations for two input sentences. We observed that the WordNet and Word2Vec techniques introduced excessive noise in our trials, which ended up deteriorating the performance of our models. For the back-translation, we experiment with German $\leftrightarrow$ English translation models for the augmentation because of the syntactical dissimilarity between the two languages. Although the back-translation method was able to broaden the variety of the sentence structure, it occasionally led to the rephrasing of the actual fallacious components of the sentences. Therefore, while we believe that back-translation and transformer-based substitution together would work best with improved translation models, in this work, we focus on augmentation with RoBERTa embedding-based synonym substitution (RESS).

\begin{figure}[!t]
    \centering
    \includegraphics[width=\linewidth]{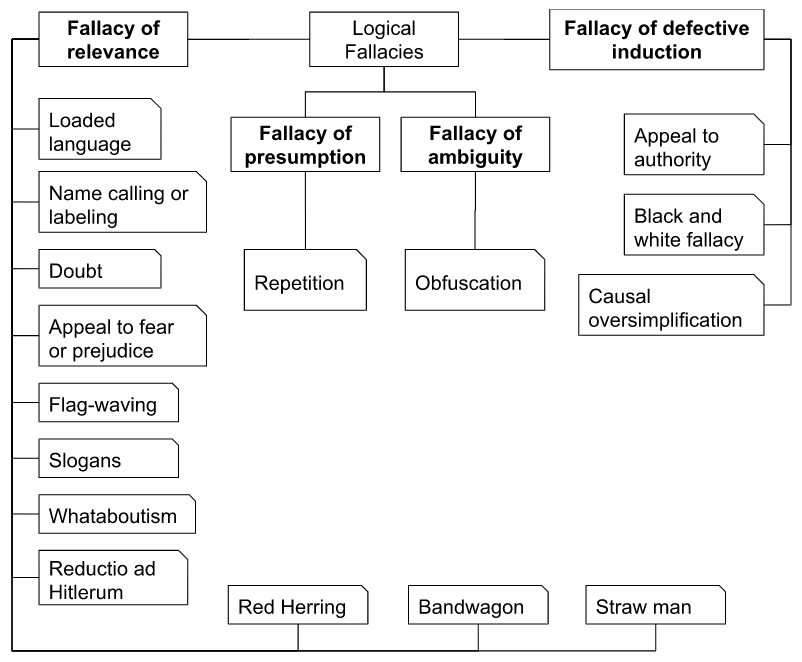}
    \caption{Three-stage taxonomy of propaganda detection.} 
    \label{fig:ptc_taxonomy}
\end{figure}

\textbf{Augmentation by Adapting Related Benchmarks.}
We investigate the possibility of augmenting the training data with human-curated datasets created for the related task of propaganda detection. As discussed in \cref{sec:related-work}, this task applies various logical fallacy techniques including \textit{Ad Hominem}, \textit{Red Herring}, \textit{Appeal to Emotion}, and \textit{Irrelevant Authority}. We adopt the \textit{Propaganda Techniques Corpus (PTC)}~\cite{da2019fine}, which includes techniques that can be found in journalistic articles and can be judged intrinsically, without the need to retrieve supporting information from external resources. The taxonomy of PTC is illustrated in Figure~\ref{fig:ptc_taxonomy}. 

The PTC dataset consists of news articles, where each sentence can have between zero, one or more fallacy annotations. As such, we adapt the PTC dataset for augmentation as follows.
If a sentence contains more than one propaganda technique, then that sentence is duplicated with all its respective labels. 
We also combine one previous sentence, as a context, with the original labeled sentence only if the previous sentence does not belong to another fallacy class.
As some of the fine-grained classes of PTC differ from those of our logical fallacy framework, we use PTC for augmentation after mapping its 18 classes to coarse-grained classes.
To do so, we map the fine-grained classes in the PTC dataset to their closest fine-grained class correspondents in the logical fallacy dataset using the class definitions and descriptions. We then simply apply the broad class mapping created for the logical fallacy dataset and map the PTC fine-grained classes to the logical fallacy coarse classes.
As the goal of this merging is to use the PTC coarse-grained classes for augmentation, we only leverage the training set of PTC and discard its development and test sets.
Since the imbalance of the dataset worsens after merging, we use the RESS-based augmentation to augment the three under-represented classes in the merged training setup to a minimum of $n=2000$ samples. We cap the augmentation to this amount so as to avoid repetitions and noise in the augmented dataset, which become dominant in the case of augmenting until the number of samples in the largest class ($n\approx4,000$). We refer the reader to Table~\ref{tab:ptc_augmentation_statistics} for augmentation statistics.


\begin{table}[t]
\centering
\caption{Training data augmentation statistics for PTC.}
\small
\resizebox{\columnwidth}{!}{%
\begin{tabular}{@{}lrr@{}}
\toprule
\textbf{Fallacy Class}&\textbf{Pre-augmentation}&\textbf{Post-augmentation}\\
\midrule
Relevance & 3950 & 3950\\
Defective induction & 1040 & 2000\\
Presumption & 536 & 2000\\
Ambiguity & 42 & 2000\\
\bottomrule
\end{tabular}
}
\label{tab:ptc_augmentation_statistics}
\end{table}

\begin{figure*}
     \centering
     \begin{subfigure}[b]{0.33\textwidth}
         \centering
         \includegraphics[width=\textwidth]{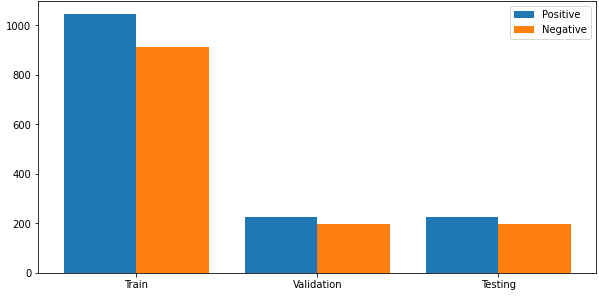}
         \caption{BIG Bench Distribution}
         \label{fig:bb_dist}
     \end{subfigure}
     \hfill
     \begin{subfigure}[b]{0.33\textwidth}
         \centering
         \includegraphics[width=\textwidth]{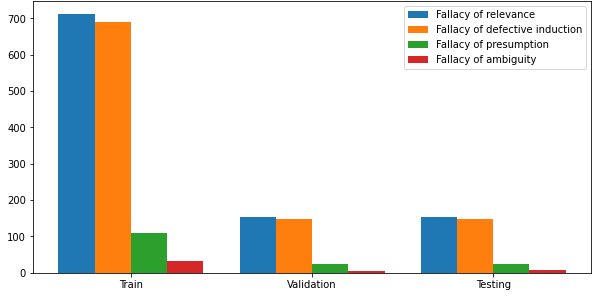}
         \caption{LOGIC Coarse-Grained Distribution}
         \label{fig:coarse_dist}
     \end{subfigure}
     \hfill
     \begin{subfigure}[b]{0.33\textwidth}
         \centering
         \includegraphics[width=\textwidth]{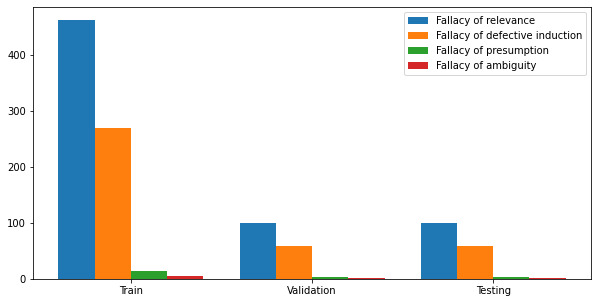}
         \caption{LOGIC Climate Coarse Distribution}
         \label{fig:climate_coarse_dist}
     \end{subfigure}
        
    \begin{subfigure}[b]{0.45\textwidth}
         \centering
         \includegraphics[width=\textwidth]{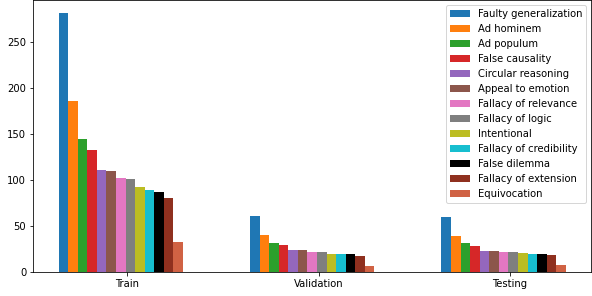}
         \caption{LOGIC Fine-Grained Distribution}
         \label{fig:fine_dist}

     \end{subfigure}
     \hfill
     \begin{subfigure}[b]{0.45\textwidth}
         \centering
         \includegraphics[width=\textwidth]{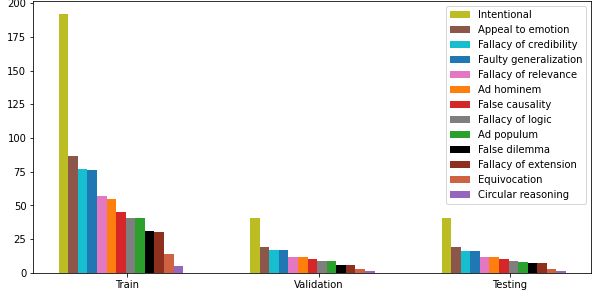}
         \caption{LOGIC Climate Fine-Grained Distribution}
         \label{fig:climate_fine_dist}
     \end{subfigure}
    \caption{Dataset Distributions.} 
    \label{fig:dataset_dist}
    \hfill
\end{figure*}

\section{Experimental Setup}
\label{sec:experimental-setup}

\subsection{Evaluation}

\textbf{Binary Logical Fallacy Detection.}
BIG Bench~\cite{ghazal2013bigbench} is a benchmarking dataset that is used for probing the representations of large language models to check their biases on various sub-tasks. BIG Bench includes two tasks for probing fallacies: binary logical fallacy detection and the formal fallacy syllogism negation. We use the binary fallacy detection dataset for evaluating whether the methods can distinguish between normal and fallacious arguments. We do not use the formal fallacy syllogism negation dataset since its format and purpose involve the deduction of the validity of sentences on the basis of the two provided premises, which is not directly related to the objective of this paper. 

We split the BIG Bench logical fallacy dataset into training, validation, and testing sets, for which the distributions are shown in Figure \ref{fig:bb_dist}. The dataset is balanced and contributes 2,800 samples across all three splits.

\textbf{Fine-Grained Classification.} For the fine-grained classification evaluation, we use the LOGIC and LOGIC Climate datasets introduced in \cite{logical_fallacy_main_paper}. There are thirteen classes within the LOGIC and LOGIC Climate fallacy datasets as described in Table~\ref{tab:fallacy-examples}.
The LOGIC dataset contains everyday fallacious arguments belonging to various topics. We use the cleaned and revised version of this dataset.\footnote{\url{https://github.com/tmakesense/logical-fallacy/tree/main/dataset-fixed}} The LOGIC Climate dataset consists of climate change news articles and fallacious arguments detected in them. We use LOGIC Climate as an evaluation-only dataset. As observed in Figures \ref{fig:fine_dist} and \ref{fig:climate_fine_dist}, the distributions between the two datasets are different, with \textit{Intentional} being the largest class in the Climate dataset, whereas it is one of the under-represented classes in the LOGIC fine-grained dataset. The LOGIC Climate dataset is included to test the ability of our models to learn these under-represented classes as well as the transferability of the model's knowledge to unseen topics. 



\textbf{Coarse-Grained Classification.} 
We evaluate the coarse-grained classification based on data inferred from the LOGIC and LOGIC Climate datasets.
The coarse-grained datasets are curated by mapping fine-grained classes from these two datasets to the coarse-grained categories following Figure~\ref{fig:taxonomy}. In the mapping process, fine-grained classes with $k\le20$ samples were removed from their corresponding coarse class if this coarse class was not under-represented. For LOGIC, we left out the fine-grained classes \textit{Fallacy of Relevance}, \textit{Fallacy of Logic} and  \textit{Intentional}, as their mapping to coarse-grained classes was mostly ambiguous for the data examples.
This resulted in a four-way coarse classification task for LOGIC and LOGIC Climate into:
\textit{Fallacy of Relevance, Fallacy of Defective Induction, Fallacy of Ambiguity, and Fallacy of Presumption.}

The coarse version of the LOGIC dataset shows a clear imbalance. A visual representation of the distribution is shown in Figure~\ref{fig:coarse_dist}. To ensure that the testing and validation splits are representative of this distribution, we sample all our splits using stratification. The splits for LOGIC Climate~\cite{logical_fallacy_main_paper} are created in a similar manner. Their distribution is shown in Figure~\ref{fig:climate_coarse_dist}. 





\textbf{Evaluation Regime and Metrics.} We test the models on the BIG Bench and LOGIC datasets by fine-tuning and curriculum learning. We apply the models trained on the LOGIC dataset for fine- and coarse-grained classification in a zero-shot fashion to the corresponding LOGIC Climate data. We report the average model performance over three runs.
We use weighted precision, recall, F1-score, and accuracy to characterize the performance of different models. Weighted measures are used to assess the per-class scores more accurately for the available unbalanced testing sets. 

\subsection{Implementation Details}

\textbf{Baselines.} We experiment with six NLI/MNLI base version models: BERT \cite{bert-mnli}, DeBERTa \cite{deberta-nli}, DistilBERT \cite{distilbert-nli}, Electra \cite{electra-mnli} and RoBERTa \cite{reimers_2021}. We utilize NLI models because we find that they perform better on the tasks of logical fallacy identification. This can be expected given that they are trained on a larger variety of data than MLM or similar models. NLI models have also been shown to have a better grasp of concepts than their MLM counterparts and to produce embeddings with better semantic representations \cite{nli_fine_tune}.
To contextualize the results, we also evaluate two simple baselines: a random baseline and a baseline that picks classes based on the relative frequency of classes in the training set.

\textbf{Instance-Based Reasoning.} For all the experiments, we use a sweep over the hyperparameters such as weight decay (L2 regularization), learning rate, and feed-forward network dropout rate.
Since we use a threshold to filter the fetched similar examples from the retriever, based on cosine similarity, we use a sweep over the used threshold as well. We then use the best combination on the development set and report the average performance on three runs using the best hyperparameters. We use the NLI-initialized Electra-base LM as the underlying encoder for generating input sentence embeddings and train our models for ten epochs in each experiment. As we observe that the $0.5$ similarity filter for fetched similar cases from the retrievers yields the best results, we apply a similarity filter on top of the retrievers discarding any fetched case whose cosine similarity to the new case is below $0.5$. We use multi-head attention with eight heads. The number of cases ($k$) used in our experiments ranges from $1$ to $10$. We do not experiment with more cases due to the stable trend seen when increasing the number of cases.

\textbf{Prototype-Based Reasoning.} We experiment with a different number of positive and negative prototypes and find that 49 positive prototypes and 1 negative prototype works best for the fine-grained classification task. We keep the same number of prototypes for the binary, coarse-, and fine-grained classification tasks. To train the negative prototype, we also include a ``None'' class, supported by the examples from the negative class in the binary classification task.
We use the NLI-initialized Electra-base as the underlying encoder for generating input sentence embeddings and report the best metrics averaged over three runs. We monitor the validation loss to choose the best model and use early stopping ($patience=10$) to prevent overfitting. We also compute class weights to handle any imbalance in the training dataset.



\textbf{Knowledge Injection.} For the experiments with K-BERT, we perform a grid search and report the results for the best-performing set of parameters.
We use grid search to find the optimal parameters: a learning rate of $2 \times 10^{-5}$ with a dropout of 0.5. We use the BERT-base model by injecting knowledge from CSKG and fine-tune the KI model over different datasets for five epochs.

\textbf{Curriculum Learning.}
For a fair comparison of the curriculum learning pipeline against the baseline model, we report scores on the default hyperparameters of the fine-tuned model, though we expect an overall increase for all metrics of at least 2-3\% when these models are tuned.
We train for 5, 8, and 10 epochs for each tuning stage respectively in the curriculum learning pipeline to avoid loss of knowledge across multiple fine-tuning stages. We fix the batch size to 32, the learning rate to $5 \times 10^{-5}$, and we use the cosine learning rate scheduler while keeping the remaining hyperparameters for our experiments unchanged.

\textbf{Data Augmentation.} We conduct experiments with different augmentation techniques for word-based and sentence-based augmentation using NLPAug \cite{nlpaug}. We experiment with a range of augmentation probabilities and the number of suitable substitutions for RESS, discovering the best results with 5 substitutions, while over 10 substitutions leads to a decrease in performance. Similarly, we obtain the best results with the augmentation threshold set between $80-90\%$, and a maximum of three replacements per argument.

\begin{table}[!t]
\centering
 \small
\caption{The corresponding runtime (in seconds) for the experiments done with each method and model per epoch, for the best models on the binary fallacy detection task.}
\begin{tabular}{@{}llrrr@{}}
\toprule
\textbf{Method}&\textbf{Model}& Binary & Coarse & Fine\\
\midrule
NLI & Electra & 66.0 & 89.0 & 108.5\\
\midrule
IBR & Electra & 103.7 & 147.6 & 191.2\\
\midrule
PBR & Electra & 15.5 & 38.2 & 114.8\\
\midrule
KI & K-BERT & 96.4 & 107.3 & 123.9 \\
\bottomrule
\end{tabular}
\label{tab:runtimes}
\end{table}

\begin{table}[!t]
\centering
\small
\caption{Main results for the best models for each method family on binary logical fallacy detection on the BIG Bench dataset.
}
\begin{tabular}{@{}llrrrr@{}}
\toprule
& &
\multicolumn{4}{c}{ \textbf{BIG Bench (Binary)}}\\
\cmidrule(lr){3-6}
\textbf{Method}&\textbf{Model}& Acc&P&R&F1\\
\midrule
Random & / & 0.499 & 0.508 & 0.499 & 0.499 \\
Frequency & / & 0.501 & 0.501 & 0.501 & 0.501 \\
NLI          & Electra & 0.995 & 0.995 & 0.995 & 0.995 \\ \hline
NLI FCL & Electra & 0.995 & 0.995 & 0.995 & 0.995  \\
IBR & Electra &  \textbf{0.997} &  \textbf{0.997} &  \textbf{0.997} &  \textbf{0.997} \\
PBR & Electra & 0.984 & 0.984 & 0.984 & 0.984\\
KI & BERT & 0.776 & 0.779 &  0.775 & 0.777 \\
\bottomrule
\end{tabular}
\label{tab:main_results_bb}
\end{table}

\begin{table*}[!t]
\centering
\small
\caption{Main results for the best models for each method family on the coarse-grained classification. 
}
\begin{tabular}{@{}llrrrlrrrl@{}}
\toprule
& & 
\multicolumn{4}{c}{ \textbf{LOGIC}} & \multicolumn{4}{c}{ \textbf{LOGIC Climate}} \\
\cmidrule(lr){3-6} \cmidrule(lr){7-10}
\textbf{Type}&\textbf{Model}& Acc&P&R&F1& Acc&P&R&F1\\
\midrule
Random & / & 0.249 & 0.413 & 0.249 & 0.298 & 0.249 & 0.508 & 0.249 & 0.323 \\
Frequency & / &  0.415 & 0.413 & 0.415 & 0.413 & 0.446 & 0.508 & 0.446 & 0.468\\
NLI & Electra & 0.767 & 0.765 & 0.767 & 0.764 $\pm{0.01}$ & 0.509 &  \textbf{0.602}   & 0.509  &  0.498  $\pm{0.01}$ \\ \hline
NLI FCL & DeBERTa & 0.758 & 0.748 & 0.758 & 0.751 $\pm{0.02}$ & 0.491 & 0.552 & 0.491  &  0.490 $\pm{0.02}$ \\
IBR & Electra & \textbf{0.829} &	\textbf{0.827} &	\textbf{0.829} &	\textbf{0.827} $\pm{0.01}$ & 0.459 & 0.585 & 0.459 & 0.466 $\pm{0.01}$ \\
PBR & Electra & 0.708 & 0.711 & 0.708 & 0.695 $\pm{0.03}$ & \textbf{0.578} & 0.570 & \textbf{ 0.578} &  \textbf{0.573} $\pm{0.03}$\\
KI & BERT & 0.787 & 0.781 &  0.782 &  0.781 $\pm{0.03}$ & 0.385 & 0.589 & 0.385 & 0.415 $\pm{0.01}$ \\ 
\bottomrule
\end{tabular}
\label{tab:main_results_coarse}
\end{table*}

\begin{table*}[!t]
\centering
\small
\caption{Main results for the best models for each method family on the fine-grained classification.
}
\begin{tabular}{@{}llrrrlrrrl@{}}
\toprule
& & 
\multicolumn{4}{c}{ \textbf{LOGIC}} & \multicolumn{4}{c}{ \textbf{LOGIC Climate}} \\
\cmidrule(lr){3-6} \cmidrule(lr){7-10}
\textbf{Type}&\textbf{Model}& Acc&P&R&F1& Acc&P&R&F1\\
\midrule
Random & / & 0.076 & 0.094 & 0.076 & 0.079 & 0.077 & 0.124 & 0.077 & 0.085 \\
Frequency & / & 0.094 & 0.094 & 0.094 & 0.093 & 0.079 & 0.120 & 0.079 & 0.080\\
NLI & Electra & 0.602 & 0.614 & 0.602 & 0.599 $\pm{0.02}$ & 0.229 & 0.276  & 0.229 & 0.217 $\pm{0.01}$ \\ \hline
NLI FCL & Electra & 0.613 & 0.624 & 0.613 & 0.610 $\pm{0.04}$ & 0.236 &  0.304  &    0.236 & 0.243 $\pm{0.02}$\\
IBR & Electra &  \textbf{0.631} &  \textbf{0.638} &  \textbf{0.631} & \textbf{0.627} $\pm{0.01} $ &  \textbf{0.254} & 0.281 &  \textbf{0.254} & \textbf{0.245} $\pm{0.01}$ \\
PBR & Electra & 0.574 & 0.600 & 0.574 & 0.574 $\pm{0.01}$ & 0.199 & \textbf{0.330} & 0.199 & 0.166 $\pm{0.01}$\\
KI & BERT &  0.488 & 0.478 & 0.488 & 0.482 $\pm{0.03}$ & 0.106 & 0.092 & 0.106 & 0.090 $\pm{0.02}$ \\
\bottomrule
\end{tabular}
\label{tab:main_results_fine}
\end{table*}



\section{Results}
\label{sec:results}


We run all our experiments on a cluster of A100-PCIE-40GB GPUs. The runtime of our experiments depends on the family of methods used, the dataset, and the size of the model being fine-tuned. We report runtimes for our best models on the binary fallacy detection task as well as coarse- and fine-grained classification task in Table \ref{tab:runtimes}. The recorded times show that the runtime of the models is mostly within the same order of magnitude of tens of seconds for binary, around a hundred seconds for coarse-, and between one and two hundred seconds for fine-grained classification. The PBR model is exceptionally efficient to train - its runtime is lower or comparable to the baseline NLI model. IBR takes the longest to run, taking one order of magnitude longer than PBR for binary classification and twice as long for fine-grained classification. 
As the encoding stage that is part of the retriever in the IBR framework is executed as a preprocessing step and is presented as a look-up table in the training stage, the time that is needed to encode all training examples with an encoder is excluded in this table. 

\subsection{Overview of the Results}
\label{subsec:results-main-results}


Tables \ref{tab:main_results_bb}, \ref{tab:main_results_coarse}, and \ref{tab:main_results_fine} show the obtained results for each method: NLI baseline, NLI with FCL, IBR, PBR, and KI on the tasks of logical fallacy detection, coarse-grained classification, and fine-grained classification. Here, we present the best result per method, indicating the corresponding model, and dive into each method in the subsequent sections. All presented results use augmentation data based on modifying the original task data (RESS). 

We observe that all methods besides KI can solve the logical fallacy detection task with a nearly perfect F1-score ($98.4\% - 99.7\%$), with the IBR method using an NLI-Electra language model reaching the best performance (cf. Table \ref{tab:main_results_bb}). The results on the coarse- and fine-grained tasks show more intriguing patterns. IBR again obtains the best performance on the in-domain task (LOGIC dataset) achieving $82.7\%$ and $62.7\%$ F1-scores on the coarse-grained and fine-grained datasets, respectively (cf. Tables \ref{tab:main_results_coarse},\ref{tab:main_results_fine}). 
However, the trends are more mixed when generalizing to the out-of-domain task of LOGIC Climate. 
The transfer learning F1-score of IBR ($46.6\%$) falls behind the PBR model ($57.3\%$) on the coarse-grained classification of the LOGIC Climate data (cf. Table \ref{tab:main_results_coarse}), while the performance of the NLI method with curriculum learning performs on par with IBR ($\sim24\%$) for the LOGIC Climate fine-grained task outperforming the other models (cf. Table \ref{tab:main_results_fine}). 
Among the different language models, most of our methods achieve the best results when using Electra with NLI initialization. 

All in all, we observe that CBR models (IBR and PBR) perform better than baseline, curriculum learning, and KI, while offering inherent explainability. 
We observe a significant gap between the performance of all the models on the in-domain dataset (LOGIC) and the out-of-domain dataset (Climate LOGIC), particularly in the fine-grained dataset, which indicates the complexity of knowledge transfer in logical fallacies from topic to topic. Zooming in on the performance of the CBR models on the out-of-domain setting, prototypical examples seem to be more helpful for approaching coarse-grained classes, while simply focusing on the semantic similarity of previous cases to approach new ones is performing better for fine-grained logical fallacies.

These results provide insights into the overall trends between the method families, however, many questions remain open. We next investigate the following questions. \textit{Does augmentation help? (\cref{subsec:results-effects-of-augmentation}) Does curriculum learning have a consistent impact across models? (\cref{subsec:results-effect-of-cl})  Does commonsense knowledge and reasoning by cases have a robust and notable effect on the model performance? (\cref{subsec:results-analysis-of-method-performance}) Do instances, prototypes, and commonsense knowledge provide intuitive explanatory mechanisms? (\cref{subsec:results-qualitative-analysis}) Which classes are helped by our methods, and which remain difficult to address? (\cref{subsec:results-per-class-analysis})}




\subsection{Effect of Augmentation}
\label{subsec:results-effects-of-augmentation}

As the LOGIC dataset is highly imbalanced, we hypothesize that data augmentation will help to address this gap, ultimately bringing better performance on this dataset. The challenge with standard augmentation techniques is that logically fallacious statements have a certain structure and arrangement, which we wish to retain even after applying the augmentation technique.
We experiment with modifying the original dataset using our RESS method and including data from the neighboring propaganda dataset, PTC. 

The obtained results for our models using Forward Curriculum Learning are shown in Table~\ref{tab:da_results}. We observe that augmentation is overall helpful on the fine-grained task and harmful on the coarse-grained task. Within the fine-grained task, the RESS augmentation always outperforms the baseline which confirms our expectation that data sparsity is an important challenge and it can be addressed through RoBERTa-based synonym substitution. The PTC augmentation is partially beneficial for some models, owing to the overlap between the propaganda and the logical fallacy data.
However, the effect of augmentation with PTC is dominantly negative, signaling that despite the overlap, this dataset is prohibitively different from the logical fallacy data. On the coarse-grained data, we see that augmentation has a negative impact on four out of five models even for the RESS augmentation method. 

We investigate this further by monitoring the augmentation impact per class. Comparing the performance of the models between pre-augmented data and post-augmented data in the coarse-grained dataset, models trained on the post-augmented data perform slightly better (up to 11\%) on the \textit{Ambiguity} class that is initially under-represented. However, the effects are adversary for the three other classes that initially have much more data points. We attribute this observation to the trade-off between enriching the data and disturbance in the natural distribution that the initial dataset possesses. Although by augmenting the dataset we achieve higher performance on the sparse class, the augmentation has a negative effect on the other classes. This also explains the success of data augmentation on the fine-grained classes, which mostly have a low number of training examples.
In summary, while augmentation does not increase performance on the coarse-grained task variant, its success on the fine-grained task and on sparsely represented classes motivates the need for further analysis and development of data augmentation methods.

\begin{table*}[t]
\centering
\small
\caption{Data augmentation results on the LOGIC dataset: no data augmentation, augmentation with RESS, and augmentation with PTC. All the models in the table are trained using the Forward Curriculum Learning framework -- FCL.
}
\begin{tabular}{@{}llrrrrrrrr@{}}
\toprule
& &
\multicolumn{4}{c}{\textbf{Coarse-grained}}&\multicolumn{4}{c}{\textbf{Fine-grained}}\\
\cmidrule(lr){3-6} \cmidrule(lr){7-10}
\textbf{Model}&\textbf{Augmentation}& Acc&P&R&F1 &Acc &P&R&F1\\
\midrule

BERT & - &   \textbf{0.747} &  \textbf{0.737} &  \textbf{0.747} &  \textbf{0.739} $\pm{0.01}$ & 0.549 & 0.571 & 0.549 & 0.552 $\pm{0.01}$ \\
& RESS  & 0.727 & 0.717 & 0.727 & 0.721 $\pm{0.03}$ & \textbf{0.586} & \textbf{0.613} &  \textbf{0.586} &  \textbf{0.584} $\pm{0.02}$\\
& PTC & 0.696 & 0.651 & 0.696 & 0.667 $\pm{0.01}$ & 0.567 & 0.590 & 0.567 & 0.570 $\pm{0.02}$\\
\midrule
DeBERTa & - &  \textbf{0.765} & \textbf{0.778} & \textbf{0.765} & \textbf{0.766} $\pm{0.03}$ & 0.564 & 0.627 & 0.564 & 0.576 $\pm{0.02}$ \\
& RESS  & 0.758 & 0.748 & 0.758 & 0.751 $\pm{0.02}$ &  \textbf{0.604} & \textbf{0.632} &  \textbf{0.604} &  \textbf{0.608} $\pm{0.01}$\\
& PTC  & 0.710 & 0.675 & 0.710 & 0.683 $\pm{0.01}$ & 0.537 & 0.590 & 0.537 & 0.547 $\pm{0.04}$\\
\midrule
DistilBERT & - & 0.711 & 0.698 & 0.711 & 0.704 $\pm{0.01}$ & 0.507 & 0.529 & 0.507 & 0.509 $\pm{0.01}$\\
& RESS  &  \textbf{0.713} &  \textbf{0.703} &  \textbf{0.713} &  \textbf{0.706} $\pm{0.02}$ &  \textbf{0.520} & \textbf{0.550} &  \textbf{0.520} &  \textbf{0.525} $\pm{0.03}$ \\
& PTC & 0.704 & 0.652 & 0.704 & 0.664 $\pm{0.02}$ & 0.492 & 0.534 & 0.492 & 0.495 $\pm{0.04}$\\
\midrule
RoBERTa & - &  \textbf{0.752} &  \textbf{0.746} &  \textbf{0.752} &  \textbf{0.742} $\pm{0.01}$ & 0.504 & 0.538 & 0.504 & 0.510 $\pm{0.01}$ \\
& RESS & 0.713 & 0.710 & 0.713 & 0.706 $\pm{0.02}$ & 0.569 & 0.578 & 0.569 & 0.565 $\pm{0.02}$\\
& PTC & 0.699 & 0.647 & 0.699 & 0.666 $\pm{0.01}$ & \textbf{0.603} & \textbf{0.620} & \textbf{0.603} & \textbf{0.595} $\pm{0.01}$\\
\midrule
Electra & - &  \textbf{0.758} &  \textbf{0.745} &  \textbf{0.758} &  \textbf{0.749} $\pm{0.02}$ & 0.602 & 0.621 & 0.602 & 0.608 $\pm{0.02}$\\
& RESS & 0.722 & 0.711 & 0.722 & 0.716 $\pm{0.03}$ & \textbf{0.613} &  \textbf{0.624} & \textbf{0.613} & \textbf{0.610} $\pm{0.04}$\\
& PTC & 0.725 & 0.689 & 0.725 & 0.690 $\pm{0.01}$ & 0.578 & 0.596 & 0.578 & 0.581 $\pm{0.02}$\\
\bottomrule
\end{tabular}
\label{tab:da_results}
\end{table*}

\begin{table*}[!t]
\centering
\caption{Curriculum learning results with different NLI and PBR models on Big Bench and the LOGIC coarse- and fine-grained datasets. All models use RESS augmentation.}
\begin{tabular}{@{}llrrrrrrrrr@{}}
\toprule
\small
& & 
\multicolumn{3}{c}{\textbf{Binary (BIG Bench)}}
&\multicolumn{3}{c}{\textbf{Coarse-grained}}&\multicolumn{3}{c}{\textbf{Fine-grained}}\\
\cmidrule(lr){3-5} \cmidrule(lr){6-8} \cmidrule(lr){9-11}
\textbf{Model}&\textbf{CL Type}& P&R&F1 &P&R&F1 &P&R&F1\\

\midrule
BERT & - & \textbf{0.848} & \textbf{0.845} & \textbf{0.845} $\pm{0.01}$& 0.714 & 0.718 & 0.717 $\pm{0.04}$ & 0.583 & 0.583 & 0.583 $\pm{0.01}$ \\
 & FCL  & -& -& -& 0.717 & 0.727 & 0.721 $\pm{0.03}$ & \textbf{0.613} & \textbf{0.586} &\textbf{ 0.584} $\pm{0.02}$\\
 & RCL & 0.826 & 0.827 & 0.826 $\pm{0.00}$ & \textbf{0.783} & \textbf{0.779} & \textbf{0.778} $\pm{0.02}$ & -& -& -\\
\hline
DeBERTa & - & \textbf{0.988} & \textbf{0.988} & \textbf{0.988} $\pm{0.00}$ & 0.746 & 0.740 & 0.741 $\pm{0.03}$  & 0.607 & 0.593 & 0.592 $\pm{0.02}$ \\
& FCL  & -& -& - & 0.748 & 0.758 & 0.751 $\pm{0.02}$ & \textbf{0.632} & \textbf{0.604} & \textbf{0.608} $\pm{0.01}$\\
 & RCL &  0.908 & 0.892 & 0.889 $\pm{0.05}$ & \textbf{0.779} & \textbf{0.785} & \textbf{0.780} $\pm{0.02}$ & -& -& -\\
\hline
DistilBERT & - & \textbf{0.848} & \textbf{0.847} & \textbf{0.847} $\pm{0.01}$ & 0.684 & 0.695 & 0.683 $\pm{0.02}$  & 0.508 & 0.513 & 0.505 $\pm{0.02}$ \\
& FCL & -& -& - & 0.703 & 0.713 & 0.706 $\pm{0.02}$ & \textbf{0.550} & \textbf{0.520} & \textbf{0.525} $\pm{0.03}$ \\
& RCL  & 0.844 & 0.842 & 0.841 $\pm{0.01}$ & \textbf{0.704 }& \textbf{0.719} & \textbf{0.711} $\pm{0.03}$& -& -& -\\
\hline
RoBERTa & - & \textbf{0.983} & \textbf{0.983} & \textbf{0.983} $\pm{0.01}$  & 0.719 & 0.714 & 0.716 $\pm{0.01}$  & 0.560 & 0.545 & 0.545 $\pm{0.02}$ \\
& FCL& -& -& - &  0.710 & 0.713 & 0.706 $\pm{0.02}$  & \textbf{0.578} & \textbf{0.569} & \textbf{0.565} $\pm{0.02}$\\
& RCL & 0.900 & 0.899 & 0.899 $\pm{0.01}$ & \textbf{0.736} &\textbf{ 0.741} & \textbf{0.732} $\pm{0.01}$ & -& -& -\\
\hline
Electra & - & \textbf{0.995} & \textbf{0.995} & \textbf{0.995} $\pm{0.00}$ & 0.765 & 0.767 & 0.764 $\pm{0.01}$  & 0.614 & 0.602 & 0.599 $\pm{0.02}$\\
& FCL & -& -& - & 0.711 & 0.722 & 0.716 $\pm{0.03}$ & \textbf{0.624} & \textbf{0.613} & \textbf{0.610} $\pm{0.04}$\\
& RCL & 0.957 & 0.957 & 0.957 $\pm{0.01}$ & \textbf{0.779} & \textbf{0.782} & \textbf{0.775} $\pm{0.03}$ & -& -& - \\
\bottomrule
\end{tabular}
\label{tab:cl_results_nli}
\end{table*}

\subsection{Effect of Curriculum Learning} 
\label{subsec:results-effect-of-cl}

The effect of curriculum learning for models trained on RESS-augmented data can be seen in Table \ref{tab:cl_results_nli}. We see clear trends for all three tasks that are consistent across the NLI models.\footnote{We observe identical trends when using CL together with the PBR-based Electra model.} 
We observe that curriculum learning is beneficial for the coarse-grained and the fine-grained tasks, whereas it is detrimental for the binary detection task.

Among the two CL variants tested on the coarse-grained task, we see that RCL performs better than FCL. With the reverse curriculum, we notice that using the fine-grained weights for coarse-grained classification improves scores considerably for all models, with DeBERTa performing the best with a $0.78$ weighted F1 score. This means that all models learn more about the coarse-grained task from the fine-grained task compared to learning from the binary fallacy detection task (i.e., when we use the BIG Bench initialization weights instead of NLI). Three out of five models still improve their performance in the FCL setup. However, Electra and RoBERTa decrease their performance and increase their variance between runs in this setup, which can be attributed to their sensitivity to hyperparameter values. 

On the other two tasks, we only compare a single CL variant to the baseline models. We do not test FCL on the binary task, as there is no task that is easier than the binary detection in our pipeline to initialize the weights from. Analogously, we do not test RCL on the fine-grained task, because our pipeline has no task that is more complex than the fine-grained classification to initialize the model weights from. For the fine-grained evaluation, we see that the coarse-grained initialization performs better than the original NLI initialization. We note that using a forward curriculum leads to an increase in at least 1\% F1-scores throughout, with Electra performing the best in this category with $0.61$ weighted F1. As described before, we expected the benefit of FCL on the fine-grained task, as the forward curriculum allows the model to learn in stages of increasing difficulty, which enhances model performance at each granularity. We also observe that increasing the number of epochs at each level of the pipeline helps to reduce the forgetting of knowledge during downstream, fine-grained tasks. 
However, we observe a negative impact when using RCL for binary fallacy detection, which indicates that this task does not benefit from the initialization of models on the fallacy classification tasks. 

Overall, our results reliably
show that the curriculum learning pipeline is capable of improving performance for the logical reasoning task of fallacy detection and the coarse representations are effective in the final stage of tuning even though they do not always outperform the other initializers in the coarse stage.

\subsection{Analysis of Method Sensitivity and Ablations}
\label{subsec:results-analysis-of-method-performance}

Next, we perform ablations of the components of our methods and investigate key parameter settings.


\subsubsection{Instance-Based Reasoning}
\label{subsubsec:results-cbr}



We observe that the IBR method performs the best among the methods across all datasets (cf. Table \ref{tab:main_results_bb}, \ref{tab:main_results_coarse}, and \ref{tab:main_results_fine}). This indicates that the idea of using similar instances to solve a new problem is effective at various levels of granularity. Although considering the common belief about the trade-off between predictive ability and interpretability (\cite{10.1145/2939672.2939874}, \cite{Caruana2015IntelligibleMF}, \cite{https://doi.org/10.48550/arxiv.1608.05745}), IBR models could have not behaved as well as other methods discussed, inline with \cite{https://doi.org/10.48550/arxiv.1602.04938} and \cite{NEURIPS2018_285e19f2}, we observe that IBR models offer good accuracy, as well as potentials for explainability~\cite{https://doi.org/10.1111/cogs.12086}. We investigate the effect of the optimal number of similar cases, and of the designs of the retriever and the adapter on the performance of the method. We do not investigate different design choices for the Classifier, which is currently a feed-forward neural network, and as such, a trivial step in the framework.



\begin{figure}[!t]
\centering
\includegraphics[width=0.9\linewidth]{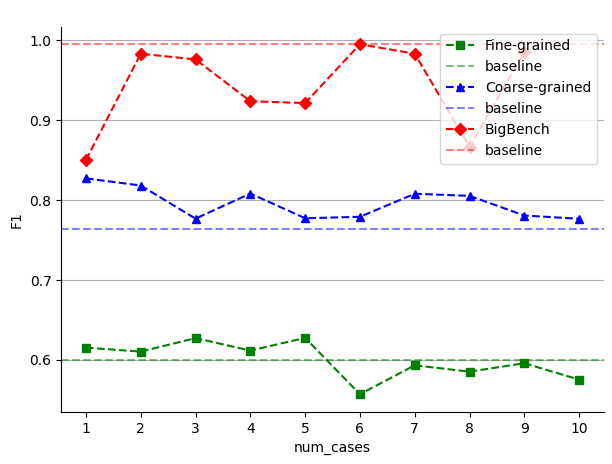}
\caption{Comparing the performance of the model being exposed to different numbers of similar cases in the IBR framework.}
\label{fig:cbr-num-cases}
\end{figure}

\textbf{Optimal number of similar cases.}
Considering the complexity of sentences containing a logical fallacy, as well as the wide range of subjects they cover and revolve around, it is most likely that for some sentences, there would be more than one already-seen sentence that would be useful or essential for the model's reasoning. It is worth mentioning that although similar cases can potentially help the model classify certain sentences better, due to the fact that retrievers are imperfect and also language models can only capture the surface meaning of the sentences (form in the language) and not necessarily understand the meaning \cite{bender-koller-2020-climbing}, adding more similar cases to the model can be considered noise and not useful. On this ground, we check the effect of the different number of cases shown to the model and assess their impact on the model's performance in Figure \ref{fig:cbr-num-cases}. As can be observed, for the coarse-grained and fine-grained datasets, there is a soft downward transition between using fewer examples and more examples that shows using more similar cases does not help the model as much as it hinders the process. This pattern differs further between coarse-grained classes and fine-grained classes. In the coarse-grained classification, regardless of the number of cases, the performance of the IBR model is always superior to the baseline, while in the fine-grained classification, having more than five similar examples would hurt the performance and cause a drop even below the baseline. Considering the fact that fewer similar cases means less noise and more similar cases means better coverage in terms of the potential aid from similar cases, we conclude that higher coverage cannot compensate for the excess noise added to the model. 

\textbf{Design of Retriever.}
For sentences that contain logical fallacies, nuances in meaning are vital to distinguish the actual relevant similar sentences from the ones that are only revolving around the same subject. Building upon this idea, we investigate different pre-trained language models as the retriever's encoder (\cref{subsubsec:method-ibr}). The comparison between these encoders is illustrated in Table \ref{tab:cbr-retriever}. 
We observe the superior performance of SimCSE on all the datasets with different granularity levels. We attribute this to the contrastive learning objective used in SimCSE. MiniLM with six layers, an all-round model tuned for many use cases, comes in the second rank.
Both SimCSE, as well as the all-MiniLM model trained on NLI, show the relevance and effectiveness of NLI for logical fallacy prediction. 
However, the paraphrase models, though trained on similar tasks such as AllNLI (concatenation of SNLI \cite{bowman2015large} and MultiNLI \cite{N18-1101}) and sentence compression, come in the last rank. 


\begin{table}[!t]
\centering
\caption{Comparing the performance of the model using different retrievers to fetch similar cases. 
}
\begin{tabular}{p{1.1cm}p{3cm}rrrr@{}}
\toprule
\textbf{Dataset}&\textbf{Retriever} & P & R & F1\\
\midrule
BIG Bench & empathy	& 0.969 & 	0.969 &	0.969 \\
& all-MiniLM-L6-v2	&  0.983	& 0.983	& 0.983 \\
& paraphrase-MiniLM-L6-v2	& 	0.861 &	0.823	& 0.822 \\
& SimCSE	 & \textbf{0.997} & \textbf{0.997} & \textbf{0.997}	\\
\midrule
Coarse-grained & empathy	 & 	0.815 &	0.813 &	0.808 \\
&all-MiniLM-L6-v2		& 0.807	& 0.801 &	0.796 \\
&paraphrase-MiniLM-L6-v2 &	0.788 &	0.785 &	0.786 \\
& SimCSE	 &	\textbf{0.827} &	\textbf{0.829} &	\textbf{0.827} \\

\midrule
Fine-grained & empathy		& 0.622	& 0.607 &	0.609 \\
& all-MiniLM-L6-v2	& 0.616 &	0.616 &	0.611 \\
& paraphrase-MiniLM-L6-v2 	& 0.588 &	0.567 &	0.567 \\
& SimCSE	& \textbf{0.638}	& \textbf{0.631}	& \textbf{0.627} \\
\bottomrule
\end{tabular}
\label{tab:cbr-retriever}
\end{table}

\begin{table}[t]
\centering
\small
\caption{Comparing the performance of the IBR model with and without using the attention mechanism.
}
\begin{tabular}{@{}llrrrr@{}}
\toprule
\textbf{Dataset}&\textbf{Attn}& Acc & P & R & F1\\
\midrule
BIG Bench & w & 	\textbf{0.997} & \textbf{0.997} & \textbf{0.997} & \textbf{0.997} \\
& w/o & 0.826 & 0.829 & 0.826 & 0.824 \\
\midrule
Coarse-grained & w &  \textbf{0.829} &	\textbf{0.827} &	\textbf{0.829} &	\textbf{0.827}\\
& w/o & 0.768 &	0.762 &	0.768	&  0.764 \\
\midrule
Fine-grained & w  & \textbf{0.631} &	\textbf{0.638} &	\textbf{0.631} &	\textbf{0.627} \\
& w/o  & 0.620 & 0.631 & 0.620 & 0.619 \\
\bottomrule
\end{tabular}
\label{tab:cbr-attention}
\end{table}

\textbf{Design of Adapter.} 
We compare our results on three datasets with and without using the attention mechanism in the third stage (adaptation). The results of this ablation study are presented in Table \ref{tab:cbr-attention}. Confirming our hypothesis, we note better performance in the presence of an attention mechanism to adjust the weights on similar cases when reasoning about the new case $C$. This observation is consistent across all datasets, which means that attention is a robust adaptation mechanism that helps the model to attend to relevant cases regardless of the granularity of the task.




\subsubsection{Prototype Learning}
\label{subsubsec:results-prototype-learning}

We dive deeper into the connection between prototypes and classes, and the sensitivity of our PBR model on the number of prototypes.

\begin{figure}[!t]
\centering
\includegraphics[width=0.8\linewidth]{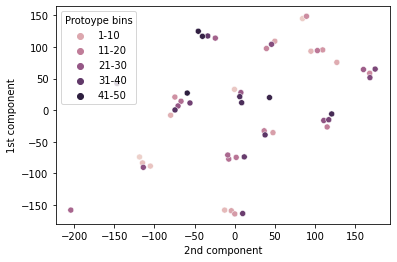}
\caption{T-SNE clustering (perplexity=2) of the 50 prototype tensors used for fine-grained classification. We flatten the prototypes thereby reducing the 98,304-dimensional data to just 2 dimensions.} 
\label{fig:tsne_clustering}
\end{figure}

\textbf{Prototypes Characterizing Classes.}
We find the prototypes responsible for the classification of each training example and assign them to the respective labels. We observe that the masking mechanism, which we introduce to the PBR method, helps to associate certain prototypes to particular classes. While we expect to see a distinct set of prototypes representative of each class, we observe a mix of distinct and common prototypes representing a particular label. For example, for the class \textit{Fallacy of Logic}, we get prototypes 6, 13, 38, and 7 as the strongest representatives. However, we observe prototype 38 to be a strong representative for five other class labels as well. We believe this is because of the nature of the overlap of fallacy classes, e.g., a fallacious sentence might have flavors of both \textit{Appeal to Emotion} and \textit{Ad Populum}, even if only one of them is annotated as the correct class. Further, we cluster the 50 prototype tensors used for the benchmarking of the fine-grained classification task and color code the prototypes based on their indices, as shown in Figure \ref{fig:tsne_clustering}.
Here, prototypes 1-10 have a light color and as we go towards prototypes 40-50, the shades get darker. We observe a certain grouping of prototype tensors, which may indicate unique features captured by the prototypes per class. 


\begin{figure}[!t]
\centering
\includegraphics[width=0.4\textwidth]{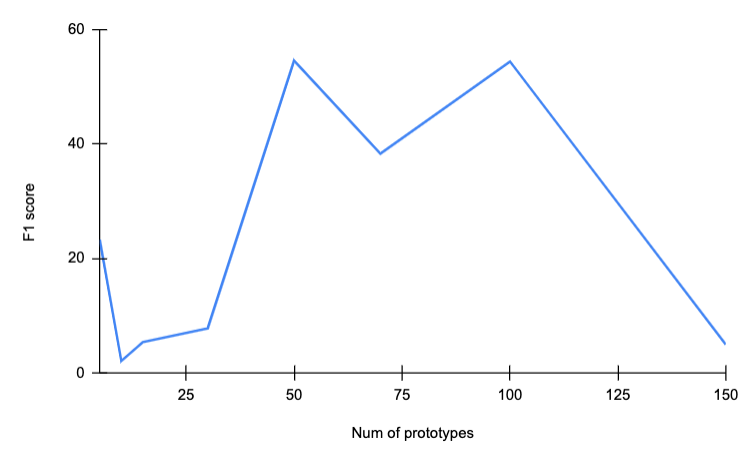}
\caption{Comparison of the performance of the PBR model for different numbers of prototypes on the LOGIC fine-grained classification task.} 
\label{fig:num_of_prototypes}
\end{figure}

\textbf{Prototypes Characterizing Classes.}
 Figure \ref{fig:num_of_prototypes} shows the trend of F1-score on the fine-grained classification task for a different number of prototypes. We assign 10\% of the prototypes to the negative class for this specific benchmarking. We observe a high sensitivity of the Prototex model to the number of prototypes, where having a too low or too high number of prototypes yields suboptimal results. We find that having a total of 50 (5 negatives) or 100 (10 negatives) prototypes yields the best performance. The PBR method is highly sensitive to the number of prototypes, and, thus, it is important to tune this hyperparameter for new datasets. Moreover, we investigate whether introducing negative prototypes is beneficial to the PBR model. Similar to \cite{das-etal-2022-prototex}, we find that including negative prototypes together with a ``None'' prediction class brings better performance on the logical fallacy coarse- and fine-grained classification tasks, though the performance gain in our case is more limited (2-3\% increase in absolute F1-scores).

\begin{table}[t]
\centering
\caption{Comparing the performance of the KI method with and without using similarity ranking of relations.
}
\resizebox{\columnwidth}{!}{%
\begin{tabular}{@{}llrrrr@{}}
\toprule
\textbf{Dataset}&\textbf{Similarity ranking}& Acc & P & R & F1\\
\midrule 
BIG Bench & w & 	\textbf{0.776} &	\textbf{0.779} &	\textbf{0.775}  & \textbf{0.777}\\
& w/o & 0.750  &	0.770	& 0.740  & 0.739 \\
\midrule
Coarse-grained & w & \textbf{0.787} & \textbf{0.781}	& \textbf{0.782} & \textbf{0.781}\\
& w/o & 0.760 &	0.706 &	0.746	&  0.721 \\
\midrule
Fine-grained & w  & \textbf{0.488} & \textbf{0.478} & \textbf{0.488} & \textbf{0.482} \\
& w/o  & 0.468 & 0.489 & 0.407 & 0.419 \\
\bottomrule
\end{tabular}
}
\label{tab:ki-priority}
\end{table}

\begin{figure}[!t]
    \centering
    \includegraphics[width=0.8\linewidth]{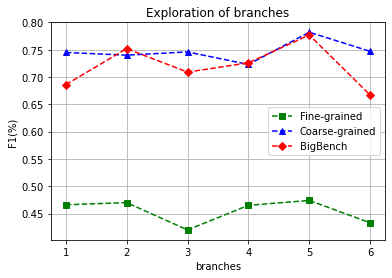}
    \caption{F1-scores for different branching factors for KI. }
    \label{fig:kbert_branching}
\end{figure}


\subsubsection{Knowledge Injection}
\label{subsec:results-KI}

We assess the performance of K-BERT, \cite{k-bert} on identifying logical fallacies in terms of the decisions made when injecting knowledge from the external KG (namely, CSKG).
The information gained from CSKG is used for forming sentence trees that are used as the primary points for knowledge injection. 
In the process of knowledge injection with CSKG, the tokens of the sentences are broken down and the triples containing the token are appended to the token to form a sentence tree.
By default, the KI method creates a sentence tree by using a maximum of two such branches per token. The exploration depth of the relation is limited to a single hop. With regards to picking useful relations, the KI method uses a brute force method for choosing triples for tokens that have multiple relations present within the knowledge base. We investigate the effect of different knowledge selection strategies, numbers of branches, and hops.

\textbf{Effect of Similarity Ranking of Relations.}
While information is appended to the sentence tree, we hypothesize that it is more meaningful to have a selection strategy in effect to select relations that add relevant knowledge to the sentence, and this serves as a point for an ablation study. 
This similarity ranking strategy enhances the performance of the KI method consistently over three different tasks for the different datasets, as observed in Table \ref{tab:ki-priority}. The performance gain is around six F1-score points for each of the three datasets, confirming our hypothesis that selecting knowledge based on relevance is important, as also shown in \cite{ma2021knowledge}. This result also motivates the need for more advanced methods for context-dependent knowledge selection.


\textbf{Branching Factor Size.} In the knowledge layer of K-BERT, the default branching factor is 2. Here, we analyze the performance of the model for different branching factors chosen (with similarity ranking of relations). As observed from Figure \ref{fig:kbert_branching}, a branching factor of 5, gives better performance over the other branching factors. We take this branching factor to represent a sweet spot between providing K-BERT with too little additional knowledge ($b<5$) and too much additional knowledge ($b>5$).   

\textbf{Number of hops.}
The base KI model uses only 1 hop of knowledge. A single hop corresponds to discovering the first relation and entity connected with the token, while by using multiple hops, we discover subsequent depths of relations based on the entities associated with them. Our analysis shows that, in the multi-hop setup, the performance of K-BERT decreases by 3-4\%. The drop in performance can be explained by the noise introduced by including multiple hops without careful filtering of the expansion. This finding is consistent with the finding of the best branching factor size that the KI model works better when presented with a smaller set of relevant relations. We look closer at the quality of the retrieved knowledge in the next section.


\begin{table*}[!t]
\centering
\caption{Input arguments with their fetched similar cases. We mark the exemplars from the same class as the input in bold.} 
\begin{tabular}{p{0.1\textwidth} | p{0.27\textwidth} | p{0.27\textwidth} | p{0.27\textwidth}}
\toprule
Class & Input Sentence & Similar Cases (IBR) & Prototypical Cases (PBR)\\
\midrule
Ad Populum & Everyone is going to get the new smart phone when it comes out this weekend. Why aren’t you? & (1) \textbf{I'm gonna get an iPhone because everybody else has an iPhone and they're cool.} & (1) \textbf{Everyone seems to support the changes in the vacation policy, and if everyone likes them, they must be good.}\\
& & (2) \textbf{Everyone wants the iPhone 11 because it's the best phone on the market!} & (2) \textbf{Everyone is buying the new iPhone that’s coming out this weekend. You have to buy it too.}\\
\midrule
Fallacy of Logic & surgeons have X-rays to guide them during an operation, lawyers have briefs to guide them during a trial, carpenters have blueprints to guide them when they are building a house. Why, then, shouldn’t students be allowed to look at their textbooks during an examination? & (1) \textbf{Doctors refer to medical books all the time when they are treating patients. In the same way, I should be allowed to use a textbook in my medical exam.} & (1) \textbf{All Paul Newman movies are great. All great movies are Oscar winners. Therefore, all Oscar winners are Paul Newman movies.}\\
& & (2) \textbf{If I say that a surgeon should be allowed to use a guidebook to carry out surgery like a student can use open notes on a test, I have made a ...} & (2) \textbf{The lady in the pink dress is Julia Roberts. The reporter thinks Julia Roberts drives a Prius. Therefore, the reporter thinks the lady in the pink dress drives a Prius.}\\
\midrule
Faulty Generalization & Everyone knows that teenagers are lazy & (1) \textbf{If we let teenagers wear whatever they want to school, they will no longer respect the rules and academic performance will decline.} &  (1) \textbf{If we allow a housing development to be built on Sunny Lake, a resort will come next, and soon we won’t have any wilderness left!}\\
& & (2) If we don’t teach teens to work harder, the human race is doomed &  (2) \textbf{Michael is part of the Jackson Five.  Without Tito and company, he will never make it.}\\
\midrule
Faulty Generalization & If you forget to floss, you will get cavities, and if you get cavities, you will lose all your teeth by the time you're 30 &  (1) \textbf{If you don’t eat breakfast, you’ll slouch in your desk. If you slouch in your desk, you’ll hurt your back. If you hurt your back, you’ll never become President.} &  (1) \textbf{If we allow gay people to get married, then the next thing you know people will be wanting to marry their pets!}\\
& & (2) four out of five dentists agree that brushing your teeth makes your life meaningful & (2) \textbf{You smoke pot? If you keep doing that, you’ll be a heroin addict within two years.}\\

\bottomrule
\end{tabular}
\label{tab:cbr-examples}
\end{table*}

\subsection{Qualitative Analysis}
\label{subsec:results-qualitative-analysis}

We analyze four cases for which the base model predicts an incorrect class, and our IBR and PBR methods change the prediction to the correct class. The KI method predicts the last two examples correctly as well.

\textbf{Quality of the Retrieved Cases.} For these four exemplars, Table \ref{tab:cbr-examples} shows the retrieved instances by IBR and prototypical examples by PBR. 
For PBR, we show the two nearest training examples to the nearest prototype for a given input.
We note that 6 out of 8 examples for IBR and all 8 examples for PBR come from the same class, which indicates that the modified decision in these cases correlates with obtaining helpful (or even representative) examples from the same class. We note, however, this is not always the case - the retrieved examples for IBR and PBR can also be from different classes. 
We observe that the corrected prediction of IBR and PBR is based on two scenarios. The first situation, shown with the first three examples for IBR in Table \ref{tab:cbr-examples}, is when the retrieved examples reflect surface similarity, which curiously still helps the model to change its decision. The second situation, observed for the last example of IBR and most PBR examples, is when the model captures the structural similarity and more abstract semantics. As we hypothesize that informal fallacies require a mixture of both aspects, observing that IBR and PBR capture them to different extents is encouraging for future work. At the same time, we also observe cases where the model correctly changes its prediction even though some of the retrieved cases belong to different classes and it is not clear how they help the model prediction. This shows the impact of the other components of our methods (the Adapter and Classifier components for IBR, or the rest of the neural architecture in PBR), but also motivates the need for future work on better models for retrieving semantically and pragmatically similar cases. 



\textbf{Quality of the Retrieved Knowledge.} Table \ref{tab:triples-examples} shows the commonsense triples retrieved by KI as background knowledge. 
As mentioned above, the KI method predicts the first two examples incorrectly, and the last two examples correctly. In example 1, we see that while the retrieved triples focus on the word \textit{phone}, it is the word \textit{everyone} in the sentence that is the main clue to the fact that the sentence belongs to the class \textit{Ad Populum}. The second example shows a case where the background knowledge misleads the model about the subject of the sentence, thus hindering it to perform a correct classification. 
In the third and fourth examples, the model is able to correctly classify the example as \textit{Faulty Generalization}, and we believe that this correlates with the quality of the retrieved knowledge. For instance, in the last example, BERT receives relevant knowledge such as flossing being used for good oral hygiene and floss related to teeth, which may have helped the model to overturn the wrong prediction into a correct one.
\begin{table*}[!t] 
\centering
\small
\caption{Examples of extracted triples with our KI method. We use an asterisk `*' to indicate the examples that have been classified correctly by KI. }
\resizebox{\textwidth}{!}{%
\begin{tabular}{p{0.2\textwidth} | p{0.4\textwidth} | p{0.27\textwidth} }
\toprule
class & Input Sentence & Sample Triples\\
\midrule
Ad Populum & Everyone is going to get the new smart phone when it comes out this weekend. Why aren’t you? & (phone, \textit{able to}, communicate), (phone, \textit{intent to}, give or get information), (weekend, \textit{related to}, relax) \\
\midrule
Fallacy of Logic & surgeons have X-rays to guide them during an operation, lawyers have briefs to guide them during a trial, carpenters have blueprints to guide them when they are building a house. Why, then, shouldn’t students be allowed to look at their textbooks during an examination? & (surgeons, \textit{related to}, operation), (operations, \textit{related to}, surgery), (student, \textit{related to}, education), (lawyers, \textit{related to}, law), (students, \textit{able to}, give exams) \\ 
\midrule 
Faulty Generalization & *Everyone knows that teenagers are lazy  & (teenager, \textit{capable of}, looking), (teenager, \textit{capable of}, performing), (teenager, \textit{is a}, juvenile person), (teenager, \textit{located near}, street)  \\ 
\midrule 
Faulty Generalization & *If you forget to floss, you will get cavities, and if you get cavities, you will lose all your teeth by the time you're 30  & (floss, \textit{used for}, good oral hygiene), (floss, \textit{related to}, teeth), (floss, \textit{related to}, dental floss), (floss, \textit{related to}, mouth) \\

\bottomrule
\end{tabular}
}
\label{tab:triples-examples}
\end{table*}

\begin{table*}[t]
\centering
\small
\caption{F1 Scores per class for LOGIC test dataset using the models trained on the augmented train split with each class having 281 data points (The number of the data points shown for the training split in the table is before augmentation).}
\begin{tabular}{@{}lllrrrrrr@{}}
\toprule
& & 
\multicolumn{5}{c}{ \textbf{F1-Scores}}\\
\cmidrule(lr){3-7}
\textbf{fine-grained class} & \textbf{coarse-grained class} & Baseline & NLI CL&IBR&ProtoTex & KI & \# test & \# train \\
\midrule
Faulty Generalization & Defective Induction & 0.656 & 0.614 & \textbf{0.660} & 0.612 & 0.549  & 60 & 281\\
Ad Hominem & Relevance & 0.596 & \textbf{0.633} & 0.627 & 0.624 & 0.607 & 39 & 185 \\
Ad Populum & Relevance& 0.812 & \textbf{0.844} & 0.814 & 0.751 & 0.656 & 31 & 144\\
False Causality & Defective Induction & 0.596 & \textbf{0.727} & 0.708 & 0.698 & 0.526 & 28 & 132\\
Circular Reasoning & Presumption & 0.524 & 0.708 & \textbf{0.719} & 0.686 & 0.450 & 23 & 110\\
Appeal to Emotion & Relevance & 0.426 & 0.473 & \textbf{0.624} & 0.445 & 0.300 & 23 & 109\\
Fallacy of Relevance & Relevance & 0.512 & 0.436 & \textbf{0.526} & 0.374 & 0.286 & 22 & 102\\
Fallacy of Logic & Defective Induction & 0.322 & 0.619 & \textbf{0.622} & 0.453 & 0.138 & 22 & 101\\
Intentional & Relevance & 0.482 & 0.356 & \textbf{0.500} & 0.419 & 0.345 & 20 & 92 \\
Fallacy of Credibility & Defective Induction & 0.400 & 0.390 & \textbf{0.486} & 0.473 & 0.231 & 19 & 89\\
False Dilemma & Defective Induction & 0.800 & 0.765 & \textbf{0.824} & 0.791 & 0.636 & 19 & 87\\
Fallacy of Extension & Relevance & 0.482 & 0.629 & 0.541 & 0.598 & \textbf{0.649} & 18 & 80\\
Equivocation & Ambiguity & 0.000 & 0.000 & 0.000 & \textbf{0.065} & 0.000 & 7 & 32\\
\bottomrule
\end{tabular}
\label{tab:main_results_perclass}
\end{table*}

\subsection{Per-Class Analysis}
\label{subsec:results-per-class-analysis}

Table \ref{tab:main_results_perclass} shows the per-class performance of our models on the fine-grained LOGIC task. Across the different classes, IBR performs best for eight out of thirteen classes and CL comes second, which is consistent with the overall results (cf. Table \ref{tab:main_results_fine}).
While we do not observe a clear pattern in terms of the superiority of methods in terms of coarse-grained classes, we do observe that the classes with more data points (top rows in Table 
\ref{tab:main_results_perclass}) are handled better by the CL model, showing that the CL model is able to reach its best performance when more data is available. This is somewhat counterintuitive, as we expect that CL can help the classes with more sparse data. However, we do observe that qualitatively CL has the best performance on the \textit{Ad} classes: \textit{False Causality}, \textit{Ad Populum}, and \textit{Ad Hominem}, indicating that the CL models are able to benefit from transferring knowledge within the same class from the coarse- to the fine-grained task. The fact that the two least populated classes are handled best by the methods KI and PBR indicates a potential for data-efficient reasoning with these methods. 

Curiously, we do not see a significant improvement using any of the models on the \textit{Equivocation} class. We attribute the consistent poor performance in this class to two important factors: 
(1) lack of training data: although we perform augmentation, this augmentation only modifies the original data slightly and does not add substantial variety to help our models understand this class better.
(2) as \textit{Equivocation} is the only fine-grained class that belongs to the broader class of \textit{Ambiguity}, our models do not have enough data points to distinguish ambiguous arguments from arguments belonging to the other classes. 
\section{Discussion}
\label{sec:discussion}



Our evaluation shows that the methods perform relatively well across tasks and even on out-of-domain arguments, while further analysis shows that curriculum learning and data augmentation are promising components of a robust methodology for identifying logical fallacies in natural language (\cref{subsec:results-effect-of-cl} \& \cref{subsec:results-effects-of-augmentation}). While our methods rely heavily on language models, the additional components such as retrieval, attention-based mechanisms, and prototype networks, provide a consistent advantage of the models over their corresponding baselines (\cref{subsec:results-analysis-of-method-performance}). Looking closer at the retrieved exemplars in the IBR and PBR methods, we observe that they are often from the same class even when they are not syntactically similar to the input case (\cref{subsec:results-qualitative-analysis}), which contributes to both the accuracy and the explainability of our models. Commonsense knowledge is also useful in particular cases, and potentially misleading in others, signifying the need for better grounding and path retrieval or generation. Looking at the performance per fallacy class, we observe that curriculum learning is able to benefit from knowledge transfer between \textit{Ad} classes, while KI and PBR perform best in the most sparse classes.

This paper pursues robust and explainable methods for reasoning about fallacies in arguments, a task that is not only understudied but also vital to support critical thinking in an educational setting~\cite{scheffer2000consensus,facione1990critical}. Our study points to research paths that should be addressed in future work.

\textbf{Further Innovation on Robust and Explainable Methods.} We observe that our models are often unable to perform abstraction and comprehend the classes in a more general sense. This has been apparent from the mixed prototype of PBR (see \cref{subsubsec:results-prototype-learning}), the mixed relevance of the examples of IBR (see \cref{tab:cbr-examples}), and the occasionally confusing triples retrieved by KI (see \cref{tab:triples-examples}). We note, however, that detecting and classifying logical fallacies is a challenging task both for modern-day AI as well as for humans, as it requires a complex (and possibly ambiguous) combination of a wide range of knowledge, including an understanding of rhetorical structures and inclusion of background knowledge about affordances and symbolism of concepts~\cite{hitchcock2017fallacies}. We see two parallel streams of AI methods that should be explored in depth for logical fallacies. On the one hand, a promising new stream relies on neural language models through methods like chain-of-thought reasoning \cite{https://doi.org/10.48550/arxiv.2201.11903}, self-rationalization \cite{https://doi.org/10.48550/arxiv.2004.14546}, and prompt decomposition \cite{cui-etal-2021-template}, coupled with large language models like GPT-3 \cite{https://doi.org/10.48550/arxiv.2005.14165} and Codex \cite{chen2021evaluating}. On the other hand, neuro-symbolic methods that, e.g., pose reasoning as a soft logic problem~\cite{https://doi.org/10.48550/arxiv.2002.05867} may provide an alternative approach to generalizable reasoning. We invite future work to explore these directions, as well as their intersection, for the challenge of logical fallacy identification.

\textbf{Focused Evaluation in Realistic and Open-Ended Settings.} The task of logical fallacy identification, and even its related task of propaganda detection, has been introduced relatively recently in the field of AI. As such, not only the methods but also the evaluation settings for these tasks are limited at present. In this study, we take a broad perspective, starting from theories of logical fallacies from social science disciplines, and we provide a unified framework that can support a more comprehensive evaluation of fallacies. We plan to extend the evaluation datasets in this paper by further annotation of data for the remaining categories like \textit{Begging the Question} and \textit{Amphiboly} in Figure~\ref{fig:taxonomy}. Moreover, beyond identifying fallacies in the context of propaganda and misinformation, we also propose that logical fallacy identification should be considered in a broader set of use cases, such as forecasting~\cite{morstatter2019sage}, where detecting wrong or misleading arguments may be central to the judgment of the trustworthiness of predictions. It is also important to consider the relation of (formal) logical fallacies to boolean satisfiability (SAT) problems~\cite{marques2008practical}, which have been proven to be NP-complete. 


\textbf{Application and Misuse of This Work.} Logical fallacies hold the promise to prevent the spread of propaganda, misinformation, and wrong argumentation among the very expansive content circulating daily on social media platforms. This could benefit both industry and governments, and ultimately ordinary social media users. However, strong logical fallacy identification models may also be misused to increase or enhance the diffusion of manipulative discourse~\cite{de2005manipulation,de2005manipulation,tymbay2022manipulative}.
We believe that analogously to the idea that encryption algorithms can be made robust if published and tested by the community~\cite{shannon1949communication}, our social media systems and communication channels will become more resilient with the progress in developing methods and evaluation tasks for logical fallacy identification.


\section{Conclusions}
\label{sec:conclusions}


This paper presented an effort to consolidate social science work on logical fallacy organization into a formal framework that can be used to develop and evaluate AI methods. The framework consisted of three stages: fallacy detection, coarse-grained classification, and fine-grained classification.
We designed a framework with three methods with native explainability and robustness: instance-based reasoning, prototype learning, and commonsense knowledge injection. To deal with the inherent data sparsity, we paired our methods with approaches for data augmentation and curriculum learning. Extensive experiments on in- and out-of-domain data showed that our methods have the ability to perform robustly across tasks, and retain much of their accuracy on out-of-domain evaluation. Curriculum learning was most helpful for coarse- and fine-grained evaluation, whereas data augmentation brought clear benefits for the most difficult task of fine-grained classification. We found that the explanation by the models in terms of known training instances or structured knowledge is easy to interpret, however, we noticed that the models still largely rely on surface form patterns and similarity in their reasoning. Guided by these insights, we proposed that future research should focus on further innovation in building robust and explainable methods, extending the evaluation to more realistic and open-ended settings, and facilitating open-source applications for social good while minimizing the possibility for misuse of the developed solutions.

\section*{Acknowledgements}
The first five authors have been supported by armasuisse Science and Technology, Switzerland under contract No. 8003532866. Zhivar Sourati has been also partially supported by NSF under Contract No. IIS-2153546, while Filip Ilievski is sponsored in part by the DARPA MCS program under Contract No. N660011924033 with the US Office Of Naval Research.
\bibliographystyle{cas-model2-names}
\bibliography{ijcai22}

\begin{thebibliography}{129}
\expandafter\ifx\csname natexlab\endcsname\relax\def\natexlab#1{#1}\fi
\providecommand{\url}[1]{\texttt{#1}}
\providecommand{\href}[2]{#2}
\providecommand{\path}[1]{#1}
\providecommand{\DOIprefix}{doi:}
\providecommand{\ArXivprefix}{arXiv:}
\providecommand{\URLprefix}{URL: }
\providecommand{\Pubmedprefix}{pmid:}
\providecommand{\doi}[1]{\href{http://dx.doi.org/#1}{\path{#1}}}
\providecommand{\Pubmed}[1]{\href{pmid:#1}{\path{#1}}}
\providecommand{\bibinfo}[2]{#2}
\ifx\xfnm\relax \def\xfnm[#1]{\unskip,\space#1}\fi
\bibitem[{Aamodt and Plaza(1994)}]{Aamodt1994}
\bibinfo{author}{Aamodt, A.}, \bibinfo{author}{Plaza, E.},
  \bibinfo{year}{1994}.
\newblock \bibinfo{title}{Case-based reasoning: Foundational issues,
  methodological variations, and system approaches}.
\newblock \bibinfo{journal}{AI Communications} \bibinfo{volume}{7},
  \bibinfo{pages}{39--59}.
\newblock \URLprefix \url{https://doi.org/10.3233/AIC-1994-7104},
  \DOIprefix\doi{10.3233/AIC-1994-7104}. \bibinfo{note}{1}.
\bibitem[{Allcott et~al.(2019)Allcott, Gentzkow and Yu}]{allcott2019trends}
\bibinfo{author}{Allcott, H.}, \bibinfo{author}{Gentzkow, M.},
  \bibinfo{author}{Yu, C.}, \bibinfo{year}{2019}.
\newblock \bibinfo{title}{Trends in the diffusion of misinformation on social
  media}.
\newblock \bibinfo{journal}{Research \& Politics} \bibinfo{volume}{6},
  \bibinfo{pages}{2053168019848554}.
\bibitem[{Almossawi(2014)}]{almossawi2014illustrated}
\bibinfo{author}{Almossawi, A.}, \bibinfo{year}{2014}.
\newblock \bibinfo{title}{An illustrated book of bad arguments}.
\newblock \bibinfo{publisher}{The Experiment}.
\bibitem[{{Aristotle}(1989)}]{Aristotle1989-jz}
\bibinfo{author}{{Aristotle}}, \bibinfo{year}{1989}.
\newblock \bibinfo{title}{On sophistical refutations: On Comin to be passing
  away - on the cosmos v. 3}.
\newblock Loeb Classical Library, \bibinfo{publisher}{LOEB},
  \bibinfo{address}{London, England}.
\bibitem[{Arora et~al.(2022)Arora, Wu, Liu and Re}]{arora-etal-2022-metadata}
\bibinfo{author}{Arora, S.}, \bibinfo{author}{Wu, S.}, \bibinfo{author}{Liu,
  E.}, \bibinfo{author}{Re, C.}, \bibinfo{year}{2022}.
\newblock \bibinfo{title}{Metadata shaping: A simple approach for
  knowledge-enhanced language models}, in: \bibinfo{booktitle}{Findings of the
  Association for Computational Linguistics: ACL 2022},
  \bibinfo{publisher}{Association for Computational Linguistics},
  \bibinfo{address}{Dublin, Ireland}. pp. \bibinfo{pages}{1733--1745}.
\newblock \URLprefix \url{https://aclanthology.org/2022.findings-acl.137},
  \DOIprefix\doi{10.18653/v1/2022.findings-acl.137}.
\bibitem[{Banarescu et~al.(2012)Banarescu, Bonial, Cai, Georgescu, Griffitt,
  Hermjakob, Knight, Koehn, Palmer and Schneider}]{banarescu2012abstract}
\bibinfo{author}{Banarescu, L.}, \bibinfo{author}{Bonial, C.},
  \bibinfo{author}{Cai, S.}, \bibinfo{author}{Georgescu, M.},
  \bibinfo{author}{Griffitt, K.}, \bibinfo{author}{Hermjakob, U.},
  \bibinfo{author}{Knight, K.}, \bibinfo{author}{Koehn, P.},
  \bibinfo{author}{Palmer, M.}, \bibinfo{author}{Schneider, N.},
  \bibinfo{year}{2012}.
\newblock \bibinfo{title}{Abstract meaning representation (amr) 1.0
  specification}, in: \bibinfo{booktitle}{Parsing on Freebase from
  Question-Answer Pairs. In Proceedings of the 2013 Conference on Empirical
  Methods in Natural Language Processing. Seattle: ACL}, pp.
  \bibinfo{pages}{1533--1544}.
\bibitem[{Bardasz and Zeid(1993)}]{bardasz1993dejavu}
\bibinfo{author}{Bardasz, T.}, \bibinfo{author}{Zeid, I.},
  \bibinfo{year}{1993}.
\newblock \bibinfo{title}{Dejavu: Case-based reasoning for mechanical design}.
\newblock \bibinfo{journal}{AI EDAM} \bibinfo{volume}{7},
  \bibinfo{pages}{111--124}.
\bibitem[{Barker(1965)}]{barker1965elements}
\bibinfo{author}{Barker, S.F.}, \bibinfo{year}{1965}.
\newblock \bibinfo{title}{The Elements of Logic}.
\newblock \bibinfo{publisher}{New York: Mcgraw-Hill}.
\bibitem[{Barriere et~al.(2022)Barriere, Tafreshi, Sedoc and
  Alqahtani}]{barriere-etal-2022-wassa}
\bibinfo{author}{Barriere, V.}, \bibinfo{author}{Tafreshi, S.},
  \bibinfo{author}{Sedoc, J.}, \bibinfo{author}{Alqahtani, S.},
  \bibinfo{year}{2022}.
\newblock \bibinfo{title}{{WASSA} 2022 shared task: Predicting empathy, emotion
  and personality in reaction to news stories}, in:
  \bibinfo{booktitle}{Proceedings of the 12th Workshop on Computational
  Approaches to Subjectivity, Sentiment {\&} Social Media Analysis},
  \bibinfo{publisher}{Association for Computational Linguistics},
  \bibinfo{address}{Dublin, Ireland}. pp. \bibinfo{pages}{214--227}.
\newblock \URLprefix \url{https://aclanthology.org/2022.wassa-1.20},
  \DOIprefix\doi{10.18653/v1/2022.wassa-1.20}.
\bibitem[{Barr{\'o}n-Cedeno et~al.(2019)Barr{\'o}n-Cedeno, Jaradat,
  Da~San~Martino and Nakov}]{barron2019proppy}
\bibinfo{author}{Barr{\'o}n-Cedeno, A.}, \bibinfo{author}{Jaradat, I.},
  \bibinfo{author}{Da~San~Martino, G.}, \bibinfo{author}{Nakov, P.},
  \bibinfo{year}{2019}.
\newblock \bibinfo{title}{Proppy: Organizing the news based on their
  propagandistic content}.
\newblock \bibinfo{journal}{Information Processing \& Management}
  \bibinfo{volume}{56}, \bibinfo{pages}{1849--1864}.
\bibitem[{Bender and Koller(2020)}]{bender-koller-2020-climbing}
\bibinfo{author}{Bender, E.M.}, \bibinfo{author}{Koller, A.},
  \bibinfo{year}{2020}.
\newblock \bibinfo{title}{Climbing towards {NLU}: {On} meaning, form, and
  understanding in the age of data}, in: \bibinfo{booktitle}{Proceedings of the
  58th Annual Meeting of the Association for Computational Linguistics},
  \bibinfo{publisher}{Association for Computational Linguistics},
  \bibinfo{address}{Online}. pp. \bibinfo{pages}{5185--5198}.
\newblock \URLprefix \url{https://aclanthology.org/2020.acl-main.463},
  \DOIprefix\doi{10.18653/v1/2020.acl-main.463}.
\bibitem[{Bengio et~al.(2009)Bengio, Louradour, Collobert and
  Weston}]{curriculum_learning_bengio}
\bibinfo{author}{Bengio, Y.}, \bibinfo{author}{Louradour, J.},
  \bibinfo{author}{Collobert, R.}, \bibinfo{author}{Weston, J.},
  \bibinfo{year}{2009}.
\newblock \bibinfo{title}{Curriculum learning}, in:
  \bibinfo{booktitle}{Proceedings of the 26th Annual International Conference
  on Machine Learning}, \bibinfo{publisher}{Association for Computing
  Machinery}, \bibinfo{address}{New York, NY, USA}. p.
  \bibinfo{pages}{41–48}.
\newblock \URLprefix \url{https://doi.org/10.1145/1553374.1553380},
  \DOIprefix\doi{10.1145/1553374.1553380}.
\bibitem[{Bernays(2004)}]{Bernays2004-fp}
\bibinfo{author}{Bernays, E.}, \bibinfo{year}{2004}.
\newblock \bibinfo{title}{Propaganda}.
\newblock \bibinfo{publisher}{Ig Publishing}, \bibinfo{address}{Brooklyn, NY}.
\bibitem[{Biotechnology and Council(2009)}]{sciencedaily_2009}
\bibinfo{author}{Biotechnology}, \bibinfo{author}{Council, B.S.R.},
  \bibinfo{year}{2009}.
\newblock \bibinfo{title}{Past experience is invaluable for complex decision
  making, brain research shows}.
\newblock \URLprefix
  \url{https://www.sciencedaily.com/releases/2009/05/090513130930.htm}.
\bibitem[{Bowman et~al.(2015)Bowman, Angeli, Potts and
  Manning}]{bowman2015large}
\bibinfo{author}{Bowman, S.R.}, \bibinfo{author}{Angeli, G.},
  \bibinfo{author}{Potts, C.}, \bibinfo{author}{Manning, C.D.},
  \bibinfo{year}{2015}.
\newblock \bibinfo{title}{A large annotated corpus for learning natural
  language inference}.
\newblock \bibinfo{journal}{arXiv preprint arXiv:1508.05326} .
\bibitem[{Brown et~al.(2020)Brown, Mann, Ryder, Subbiah, Kaplan, Dhariwal,
  Neelakantan, Shyam, Sastry, Askell, Agarwal, Herbert-Voss, Krueger, Henighan,
  Child, Ramesh, Ziegler, Wu, Winter, Hesse, Chen, Sigler, Litwin, Gray, Chess,
  Clark, Berner, McCandlish, Radford, Sutskever and
  Amodei}]{https://doi.org/10.48550/arxiv.2005.14165}
\bibinfo{author}{Brown, T.B.}, \bibinfo{author}{Mann, B.},
  \bibinfo{author}{Ryder, N.}, \bibinfo{author}{Subbiah, M.},
  \bibinfo{author}{Kaplan, J.}, \bibinfo{author}{Dhariwal, P.},
  \bibinfo{author}{Neelakantan, A.}, \bibinfo{author}{Shyam, P.},
  \bibinfo{author}{Sastry, G.}, \bibinfo{author}{Askell, A.},
  \bibinfo{author}{Agarwal, S.}, \bibinfo{author}{Herbert-Voss, A.},
  \bibinfo{author}{Krueger, G.}, \bibinfo{author}{Henighan, T.},
  \bibinfo{author}{Child, R.}, \bibinfo{author}{Ramesh, A.},
  \bibinfo{author}{Ziegler, D.M.}, \bibinfo{author}{Wu, J.},
  \bibinfo{author}{Winter, C.}, \bibinfo{author}{Hesse, C.},
  \bibinfo{author}{Chen, M.}, \bibinfo{author}{Sigler, E.},
  \bibinfo{author}{Litwin, M.}, \bibinfo{author}{Gray, S.},
  \bibinfo{author}{Chess, B.}, \bibinfo{author}{Clark, J.},
  \bibinfo{author}{Berner, C.}, \bibinfo{author}{McCandlish, S.},
  \bibinfo{author}{Radford, A.}, \bibinfo{author}{Sutskever, I.},
  \bibinfo{author}{Amodei, D.}, \bibinfo{year}{2020}.
\newblock \bibinfo{title}{Language models are few-shot learners}.
\newblock \URLprefix \url{https://arxiv.org/abs/2005.14165},
  \DOIprefix\doi{10.48550/ARXIV.2005.14165}.
\bibitem[{Br{\"u}ninghaus and Ashley(2006)}]{bruninghaus2006progress}
\bibinfo{author}{Br{\"u}ninghaus, S.}, \bibinfo{author}{Ashley, K.D.},
  \bibinfo{year}{2006}.
\newblock \bibinfo{title}{Progress in textual case-based reasoning: predicting
  the outcome of legal cases from text}, in: \bibinfo{booktitle}{AAAI}, pp.
  \bibinfo{pages}{1577--1580}.
\bibitem[{Caruana et~al.(2015)Caruana, Lou, Gehrke, Koch, Sturm and
  Elhadad}]{Caruana2015IntelligibleMF}
\bibinfo{author}{Caruana, R.}, \bibinfo{author}{Lou, Y.},
  \bibinfo{author}{Gehrke, J.}, \bibinfo{author}{Koch, P.},
  \bibinfo{author}{Sturm, M.}, \bibinfo{author}{Elhadad, N.},
  \bibinfo{year}{2015}.
\newblock \bibinfo{title}{Intelligible models for healthcare: Predicting
  pneumonia risk and hospital 30-day readmission}.
\newblock \bibinfo{journal}{Proceedings of the 21th ACM SIGKDD International
  Conference on Knowledge Discovery and Data Mining} .
\bibitem[{Chen et~al.(2021)Chen, Tworek, Jun, Yuan, Pinto, Kaplan, Edwards,
  Burda, Joseph, Brockman et~al.}]{chen2021evaluating}
\bibinfo{author}{Chen, M.}, \bibinfo{author}{Tworek, J.}, \bibinfo{author}{Jun,
  H.}, \bibinfo{author}{Yuan, Q.}, \bibinfo{author}{Pinto, H.P.d.O.},
  \bibinfo{author}{Kaplan, J.}, \bibinfo{author}{Edwards, H.},
  \bibinfo{author}{Burda, Y.}, \bibinfo{author}{Joseph, N.},
  \bibinfo{author}{Brockman, G.}, et~al., \bibinfo{year}{2021}.
\newblock \bibinfo{title}{Evaluating large language models trained on code}.
\newblock \bibinfo{journal}{arXiv preprint arXiv:2107.03374} .
\bibitem[{Chen and S.~Savage(2014)}]{https://doi.org/10.1111/1467-9817.12022}
\bibinfo{author}{Chen, V.}, \bibinfo{author}{S.~Savage, R.},
  \bibinfo{year}{2014}.
\newblock \bibinfo{title}{Evidence for a simplicity principle: teaching common
  complex grapheme-to-phonemes improves reading and motivation in at-risk
  readers}.
\newblock \bibinfo{journal}{Journal of Research in Reading}
  \bibinfo{volume}{37}, \bibinfo{pages}{196--214}.
\newblock \URLprefix
  \url{https://onlinelibrary.wiley.com/doi/abs/10.1111/1467-9817.12022},
  \DOIprefix\doi{https://doi.org/10.1111/1467-9817.12022}.
\bibitem[{Choi et~al.(2016)Choi, Bahadori, Kulas, Schuetz, Stewart and
  Sun}]{https://doi.org/10.48550/arxiv.1608.05745}
\bibinfo{author}{Choi, E.}, \bibinfo{author}{Bahadori, M.T.},
  \bibinfo{author}{Kulas, J.A.}, \bibinfo{author}{Schuetz, A.},
  \bibinfo{author}{Stewart, W.F.}, \bibinfo{author}{Sun, J.},
  \bibinfo{year}{2016}.
\newblock \bibinfo{title}{Retain: An interpretable predictive model for
  healthcare using reverse time attention mechanism}.
\newblock \URLprefix \url{https://arxiv.org/abs/1608.05745},
  \DOIprefix\doi{10.48550/ARXIV.1608.05745}.
\bibitem[{Chu(2021)}]{distilbert-nli}
\bibinfo{author}{Chu, D.}, \bibinfo{year}{2021}.
\newblock \bibinfo{title}{Typeform/distilbert-base-uncased-mnli · hugging
  face}.
\newblock \URLprefix
  \url{https://huggingface.co/typeform/distilbert-base-uncased-mnli}.
\bibitem[{Clark et~al.(2020a)Clark, Luong, Le and Manning}]{electra}
\bibinfo{author}{Clark, K.}, \bibinfo{author}{Luong, M.T.},
  \bibinfo{author}{Le, Q.V.}, \bibinfo{author}{Manning, C.D.},
  \bibinfo{year}{2020}a.
\newblock \bibinfo{title}{{ELECTRA}: Pre-training text encoders as
  discriminators rather than generators}, in: \bibinfo{booktitle}{ICLR}.
\newblock \URLprefix \url{https://openreview.net/pdf?id=r1xMH1BtvB}.
\bibitem[{Clark et~al.(2020b)Clark, Tafjord and
  Richardson}]{https://doi.org/10.48550/arxiv.2002.05867}
\bibinfo{author}{Clark, P.}, \bibinfo{author}{Tafjord, O.},
  \bibinfo{author}{Richardson, K.}, \bibinfo{year}{2020}b.
\newblock \bibinfo{title}{Transformers as soft reasoners over language}.
\newblock \URLprefix \url{https://arxiv.org/abs/2002.05867},
  \DOIprefix\doi{10.48550/ARXIV.2002.05867}.
\bibitem[{Conneau et~al.(2017)Conneau, Kiela, Schwenk, Barrault and
  Bordes}]{nli_fine_tune}
\bibinfo{author}{Conneau, A.}, \bibinfo{author}{Kiela, D.},
  \bibinfo{author}{Schwenk, H.}, \bibinfo{author}{Barrault, L.},
  \bibinfo{author}{Bordes, A.}, \bibinfo{year}{2017}.
\newblock \bibinfo{title}{Supervised learning of universal sentence
  representations from natural language inference data}, in:
  \bibinfo{booktitle}{Proceedings of the 2017 Conference on Empirical Methods
  in Natural Language Processing}, \bibinfo{publisher}{Association for
  Computational Linguistics}, \bibinfo{address}{Copenhagen, Denmark}. pp.
  \bibinfo{pages}{670--680}.
\newblock \URLprefix \url{https://aclanthology.org/D17-1070},
  \DOIprefix\doi{10.18653/v1/D17-1070}.
\bibitem[{Copi(1954)}]{Copi1954-COPITL-6}
\bibinfo{author}{Copi, I.M.}, \bibinfo{year}{1954}.
\newblock \bibinfo{title}{Introduction to logic}.
\newblock \bibinfo{journal}{Philosophy} \bibinfo{volume}{29},
  \bibinfo{pages}{271--271}.
\bibitem[{{Council of the European Union}(2022)}]{eu-2022-2065}
\bibinfo{author}{{Council of the European Union}}, \bibinfo{year}{2022}.
\newblock \bibinfo{title}{{Regulation (EU) 2022/2065 of the European Parliament
  and of the Council of 19 October 2022 on a Single Market For Digital Services
  and amending Directive 2000/31/EC (Digital Services Act) (Text with EEA
  relevance)}}.
\newblock \URLprefix \url{http://data.europa.eu/eli/reg/2022/2065/oj}.
  \bibinfo{note}{document 32022R2065. Accessed: 2022-11-30}.
\bibitem[{Cui et~al.(2021)Cui, Wu, Liu, Yang and
  Zhang}]{cui-etal-2021-template}
\bibinfo{author}{Cui, L.}, \bibinfo{author}{Wu, Y.}, \bibinfo{author}{Liu, J.},
  \bibinfo{author}{Yang, S.}, \bibinfo{author}{Zhang, Y.},
  \bibinfo{year}{2021}.
\newblock \bibinfo{title}{Template-based named entity recognition using
  {BART}}, in: \bibinfo{booktitle}{Findings of the Association for
  Computational Linguistics: ACL-IJCNLP 2021}, \bibinfo{publisher}{Association
  for Computational Linguistics}, \bibinfo{address}{Online}. pp.
  \bibinfo{pages}{1835--1845}.
\newblock \URLprefix \url{https://aclanthology.org/2021.findings-acl.161},
  \DOIprefix\doi{10.18653/v1/2021.findings-acl.161}.
\bibitem[{Da~San~Martino et~al.(2019)Da~San~Martino, Yu, Barr{\'o}n-Cedeno,
  Petrov and Nakov}]{da2019fine}
\bibinfo{author}{Da~San~Martino, G.}, \bibinfo{author}{Yu, S.},
  \bibinfo{author}{Barr{\'o}n-Cedeno, A.}, \bibinfo{author}{Petrov, R.},
  \bibinfo{author}{Nakov, P.}, \bibinfo{year}{2019}.
\newblock \bibinfo{title}{Fine-grained analysis of propaganda in news article},
  in: \bibinfo{booktitle}{Proceedings of the 2019 conference on empirical
  methods in natural language processing and the 9th international joint
  conference on natural language processing (EMNLP-IJCNLP)}, pp.
  \bibinfo{pages}{5636--5646}.
\bibitem[{Daelemans and van~den Bosch(2005)}]{daelemans_van_den_bosch_2005}
\bibinfo{author}{Daelemans, W.}, \bibinfo{author}{van~den Bosch, A.},
  \bibinfo{year}{2005}.
\newblock \bibinfo{title}{Memory-Based Language Processing}.
\newblock Studies in Natural Language Processing, \bibinfo{publisher}{Cambridge
  University Press}.
\newblock \DOIprefix\doi{10.1017/CBO9780511486579}.
\bibitem[{Das et~al.(2022)Das, Gupta, Kovatchev, Lease and
  Li}]{das-etal-2022-prototex}
\bibinfo{author}{Das, A.}, \bibinfo{author}{Gupta, C.},
  \bibinfo{author}{Kovatchev, V.}, \bibinfo{author}{Lease, M.},
  \bibinfo{author}{Li, J.J.}, \bibinfo{year}{2022}.
\newblock \bibinfo{title}{{P}roto{TE}x: Explaining model decisions with
  prototype tensors}, in: \bibinfo{booktitle}{Proceedings of the 60th Annual
  Meeting of the Association for Computational Linguistics (Volume 1: Long
  Papers)}, \bibinfo{publisher}{Association for Computational Linguistics},
  \bibinfo{address}{Dublin, Ireland}. pp. \bibinfo{pages}{2986--2997}.
\newblock \URLprefix \url{https://aclanthology.org/2022.acl-long.213},
  \DOIprefix\doi{10.18653/v1/2022.acl-long.213}.
\bibitem[{De~Saussure(2005)}]{de2005manipulation}
\bibinfo{author}{De~Saussure, L.}, \bibinfo{year}{2005}.
\newblock \bibinfo{title}{Manipulation and cognitive pragmatics}.
\newblock \bibinfo{journal}{Manipulation and ideologies in the twentieth
  century} , \bibinfo{pages}{113--145}.
\bibitem[{Devlin et~al.(2018)Devlin, Chang, Lee and Toutanova}]{devlin2018bert}
\bibinfo{author}{Devlin, J.}, \bibinfo{author}{Chang, M.W.},
  \bibinfo{author}{Lee, K.}, \bibinfo{author}{Toutanova, K.},
  \bibinfo{year}{2018}.
\newblock \bibinfo{title}{Bert: Pre-training of deep bidirectional transformers
  for language understanding}.
\newblock \bibinfo{journal}{arXiv preprint arXiv:1810.04805} .
\bibitem[{Dimitrov et~al.(2021)Dimitrov, Bin~Ali, Shaar, Alam, Silvestri,
  Firooz, Nakov and Da~San~Martino}]{dimitrov-etal-2021-detecting}
\bibinfo{author}{Dimitrov, D.}, \bibinfo{author}{Bin~Ali, B.},
  \bibinfo{author}{Shaar, S.}, \bibinfo{author}{Alam, F.},
  \bibinfo{author}{Silvestri, F.}, \bibinfo{author}{Firooz, H.},
  \bibinfo{author}{Nakov, P.}, \bibinfo{author}{Da~San~Martino, G.},
  \bibinfo{year}{2021}.
\newblock \bibinfo{title}{Detecting propaganda techniques in memes}, in:
  \bibinfo{booktitle}{Proceedings of the 59th Annual Meeting of the Association
  for Computational Linguistics and the 11th International Joint Conference on
  Natural Language Processing (Volume 1: Long Papers)},
  \bibinfo{publisher}{Association for Computational Linguistics},
  \bibinfo{address}{Online}. pp. \bibinfo{pages}{6603--6617}.
\newblock \URLprefix \url{https://aclanthology.org/2021.acl-long.516},
  \DOIprefix\doi{10.18653/v1/2021.acl-long.516}.
\bibitem[{Edunov et~al.(2018)Edunov, Ott, Auli and
  Grangier}]{backtranslation_at_scale}
\bibinfo{author}{Edunov, S.}, \bibinfo{author}{Ott, M.}, \bibinfo{author}{Auli,
  M.}, \bibinfo{author}{Grangier, D.}, \bibinfo{year}{2018}.
\newblock \bibinfo{title}{Understanding back-translation at scale}, in:
  \bibinfo{booktitle}{Proceedings of the 2018 Conference on Empirical Methods
  in Natural Language Processing}, \bibinfo{publisher}{Association for
  Computational Linguistics}, \bibinfo{address}{Brussels, Belgium}. pp.
  \bibinfo{pages}{489--500}.
\newblock \URLprefix \url{https://aclanthology.org/D18-1045},
  \DOIprefix\doi{10.18653/v1/D18-1045}.
\bibitem[{Elman(1993)}]{ELMAN199371}
\bibinfo{author}{Elman, J.L.}, \bibinfo{year}{1993}.
\newblock \bibinfo{title}{Learning and development in neural networks: the
  importance of starting small}.
\newblock \bibinfo{journal}{Cognition} \bibinfo{volume}{48},
  \bibinfo{pages}{71--99}.
\newblock \URLprefix
  \url{https://www.sciencedirect.com/science/article/pii/0010027793900584},
  \DOIprefix\doi{https://doi.org/10.1016/0010-0277(93)90058-4}.
\bibitem[{Facione(1990)}]{facione1990critical}
\bibinfo{author}{Facione, P.}, \bibinfo{year}{1990}.
\newblock \bibinfo{title}{Critical thinking: A statement of expert consensus
  for purposes of educational assessment and instruction (the delphi report)} .
\bibitem[{Feldman(2003)}]{doi:10.1046/j.0963-7214.2003.01267.x}
\bibinfo{author}{Feldman, J.}, \bibinfo{year}{2003}.
\newblock \bibinfo{title}{The simplicity principle in human concept learning}.
\newblock \bibinfo{journal}{Current Directions in Psychological Science}
  \bibinfo{volume}{12}, \bibinfo{pages}{227--232}.
\newblock \URLprefix \url{https://doi.org/10.1046/j.0963-7214.2003.01267.x},
  \DOIprefix\doi{10.1046/j.0963-7214.2003.01267.x}.
\bibitem[{Ferreira~Cruz et~al.(2019)Ferreira~Cruz, Rocha and
  Lopes~Cardoso}]{ferreira-cruz-etal-2019-sentence}
\bibinfo{author}{Ferreira~Cruz, A.}, \bibinfo{author}{Rocha, G.},
  \bibinfo{author}{Lopes~Cardoso, H.}, \bibinfo{year}{2019}.
\newblock \bibinfo{title}{On sentence representations for propaganda detection:
  From handcrafted features to word embeddings}, in:
  \bibinfo{booktitle}{Proceedings of the Second Workshop on Natural Language
  Processing for Internet Freedom: Censorship, Disinformation, and Propaganda},
  \bibinfo{publisher}{Association for Computational Linguistics},
  \bibinfo{address}{Hong Kong, China}. pp. \bibinfo{pages}{107--112}.
\newblock \URLprefix \url{https://aclanthology.org/D19-5015},
  \DOIprefix\doi{10.18653/v1/D19-5015}.
\bibitem[{Ford et~al.(2020)Ford, Kenny and
  Keane}]{https://doi.org/10.48550/arxiv.2009.06349}
\bibinfo{author}{Ford, C.}, \bibinfo{author}{Kenny, E.M.},
  \bibinfo{author}{Keane, M.T.}, \bibinfo{year}{2020}.
\newblock \bibinfo{title}{Play mnist for me! user studies on the effects of
  post-hoc, example-based explanations \&; error rates on debugging a deep
  learning, black-box classifier} \URLprefix
  \url{https://arxiv.org/abs/2009.06349},
  \DOIprefix\doi{10.48550/ARXIV.2009.06349}.
\bibitem[{Ganesh and Bright(2020)}]{ganesh2020countering}
\bibinfo{author}{Ganesh, B.}, \bibinfo{author}{Bright, J.},
  \bibinfo{year}{2020}.
\newblock \bibinfo{title}{Countering extremists on social media: Challenges for
  strategic communication and content moderation}.
\bibitem[{Gao et~al.(2021)Gao, Yao and
  Chen}]{https://doi.org/10.48550/arxiv.2104.08821}
\bibinfo{author}{Gao, T.}, \bibinfo{author}{Yao, X.}, \bibinfo{author}{Chen,
  D.}, \bibinfo{year}{2021}.
\newblock \bibinfo{title}{Simcse: Simple contrastive learning of sentence
  embeddings}.
\newblock \URLprefix \url{https://arxiv.org/abs/2104.08821},
  \DOIprefix\doi{10.48550/ARXIV.2104.08821}.
\bibitem[{Ge et~al.(2022)Ge, Mao and Cambria}]{Ge_Mao_Cambria_2022}
\bibinfo{author}{Ge, M.}, \bibinfo{author}{Mao, R.}, \bibinfo{author}{Cambria,
  E.}, \bibinfo{year}{2022}.
\newblock \bibinfo{title}{Explainable metaphor identification inspired by
  conceptual metaphor theory}.
\newblock \bibinfo{journal}{Proceedings of the AAAI Conference on Artificial
  Intelligence} \bibinfo{volume}{36}, \bibinfo{pages}{10681--10689}.
\newblock \URLprefix
  \url{https://ojs.aaai.org/index.php/AAAI/article/view/21313},
  \DOIprefix\doi{10.1609/aaai.v36i10.21313}.
\bibitem[{Ghazal et~al.(2013)Ghazal, Rabl, Hu, Raab, Poess, Crolotte and
  Jacobsen}]{ghazal2013bigbench}
\bibinfo{author}{Ghazal, A.}, \bibinfo{author}{Rabl, T.}, \bibinfo{author}{Hu,
  M.}, \bibinfo{author}{Raab, F.}, \bibinfo{author}{Poess, M.},
  \bibinfo{author}{Crolotte, A.}, \bibinfo{author}{Jacobsen, H.A.},
  \bibinfo{year}{2013}.
\newblock \bibinfo{title}{Bigbench: Towards an industry standard benchmark for
  big data analytics}, in: \bibinfo{booktitle}{Proceedings of the 2013 ACM
  SIGMOD international conference on Management of data}, pp.
  \bibinfo{pages}{1197--1208}.
\bibitem[{Gibbs(2010)}]{Gibbs2010-dp}
\bibinfo{author}{Gibbs, N.M.}, \bibinfo{year}{2010}.
\newblock \bibinfo{title}{Formal and informal fallacies in anaesthesia}.
\newblock \bibinfo{journal}{Anaesth Intensive Care} \bibinfo{volume}{38},
  \bibinfo{pages}{639--646}.
\bibitem[{Gibson et~al.(2007)Gibson, Rowe and Reed}]{gibson2007computational}
\bibinfo{author}{Gibson, A.}, \bibinfo{author}{Rowe, G.},
  \bibinfo{author}{Reed, C.}, \bibinfo{year}{2007}.
\newblock \bibinfo{title}{A computational approach to identifying formal
  fallacy}.
\newblock \bibinfo{journal}{CMNA VII-Computational Models of Natural Argument}
  .
\bibitem[{Goffredo et~al.(2022)Goffredo, Haddadan, Vorakitphan, Cabrio and
  Villata}]{goffredo2022fallacious}
\bibinfo{author}{Goffredo, P.}, \bibinfo{author}{Haddadan, S.},
  \bibinfo{author}{Vorakitphan, V.}, \bibinfo{author}{Cabrio, E.},
  \bibinfo{author}{Villata, S.}, \bibinfo{year}{2022}.
\newblock \bibinfo{title}{Fallacious argument classification in political
  debates}, in: \bibinfo{booktitle}{Thirty-First International Joint Conference
  on Artificial Intelligence $\{$IJCAI-22$\}$},
  \bibinfo{organization}{International Joint Conferences on Artificial
  Intelligence Organization}. pp. \bibinfo{pages}{4143--4149}.
\bibitem[{Goodwin(1998)}]{Goodwin1998}
\bibinfo{author}{Goodwin, J.}, \bibinfo{year}{1998}.
\newblock \bibinfo{title}{Forms of authority and the real ad verecundiam}.
\newblock \bibinfo{journal}{Argumentation} \bibinfo{volume}{12},
  \bibinfo{pages}{267--280}.
\newblock \URLprefix \url{https://doi.org/10.1023/a:1007756117287},
  \DOIprefix\doi{10.1023/a:1007756117287}.
\bibitem[{Gundapu and Mamidi(2022)}]{gundapu}
\bibinfo{author}{Gundapu, S.}, \bibinfo{author}{Mamidi, R.},
  \bibinfo{year}{2022}.
\newblock \bibinfo{title}{Detection of propaganda techniques in visuo-lingual
  metaphor in memes}.
\newblock \URLprefix \url{https://arxiv.org/abs/2205.02937},
  \DOIprefix\doi{10.48550/ARXIV.2205.02937}.
\bibitem[{Gupta et~al.(2019)Gupta, Saxena, Yaseen, Runkler and
  Schütze}]{https://doi.org/10.48550/arxiv.1909.06162}
\bibinfo{author}{Gupta, P.}, \bibinfo{author}{Saxena, K.},
  \bibinfo{author}{Yaseen, U.}, \bibinfo{author}{Runkler, T.},
  \bibinfo{author}{Schütze, H.}, \bibinfo{year}{2019}.
\newblock \bibinfo{title}{Neural architectures for fine-grained propaganda
  detection in news}.
\newblock \URLprefix \url{https://arxiv.org/abs/1909.06162},
  \DOIprefix\doi{10.48550/ARXIV.1909.06162}.
\bibitem[{Hamilton(2021)}]{hamilton2021towards}
\bibinfo{author}{Hamilton, K.}, \bibinfo{year}{2021}.
\newblock \bibinfo{title}{Towards an ontology for propaganda detection in news
  articles}, in: \bibinfo{booktitle}{European Semantic Web Conference},
  \bibinfo{organization}{Springer}. pp. \bibinfo{pages}{230--241}.
\bibitem[{Han et~al.(2022)Han, Mao and
  Cambria}]{https://doi.org/10.48550/arxiv.2209.07494}
\bibinfo{author}{Han, S.}, \bibinfo{author}{Mao, R.}, \bibinfo{author}{Cambria,
  E.}, \bibinfo{year}{2022}.
\newblock \bibinfo{title}{Hierarchical attention network for explainable
  depression detection on twitter aided by metaphor concept mappings}.
\newblock \URLprefix \url{https://arxiv.org/abs/2209.07494},
  \DOIprefix\doi{10.48550/ARXIV.2209.07494}.
\bibitem[{Hansen(2020a)}]{sep-fallacies}
\bibinfo{author}{Hansen, H.}, \bibinfo{year}{2020}a.
\newblock \bibinfo{title}{{Fallacies}}, in: \bibinfo{editor}{Zalta, E.N.}
  (Ed.), \bibinfo{booktitle}{The {Stanford} Encyclopedia of Philosophy}.
  \bibinfo{edition}{{S}ummer 2020} ed.. \bibinfo{publisher}{Metaphysics
  Research Lab, Stanford University}.
\bibitem[{Hansen(2020b)}]{hansen_2020}
\bibinfo{author}{Hansen, H.}, \bibinfo{year}{2020}b.
\newblock \bibinfo{title}{Fallacies}.
\newblock \URLprefix \url{https://plato.stanford.edu/entries/fallacies}.
\bibitem[{He(2021)}]{electra-mnli}
\bibinfo{author}{He, H.}, \bibinfo{year}{2021}.
\newblock \bibinfo{title}{Howey/electra-base-mnli · hugging face}.
\newblock \URLprefix \url{https://huggingface.co/howey/electra-base-mnli}.
\bibitem[{Hendrycks and
  Gimpel(2016)}]{https://doi.org/10.48550/arxiv.1606.08415}
\bibinfo{author}{Hendrycks, D.}, \bibinfo{author}{Gimpel, K.},
  \bibinfo{year}{2016}.
\newblock \bibinfo{title}{Gaussian error linear units (gelus)}.
\newblock \URLprefix \url{https://arxiv.org/abs/1606.08415},
  \DOIprefix\doi{10.48550/ARXIV.1606.08415}.
\bibitem[{Heo et~al.(2018)Heo, Lee, Kim, Lee, Kim, Yang and
  Hwang}]{NEURIPS2018_285e19f2}
\bibinfo{author}{Heo, J.}, \bibinfo{author}{Lee, H.B.}, \bibinfo{author}{Kim,
  S.}, \bibinfo{author}{Lee, J.}, \bibinfo{author}{Kim, K.J.},
  \bibinfo{author}{Yang, E.}, \bibinfo{author}{Hwang, S.J.},
  \bibinfo{year}{2018}.
\newblock \bibinfo{title}{Uncertainty-aware attention for reliable
  interpretation and prediction}, in: \bibinfo{editor}{Bengio, S.},
  \bibinfo{editor}{Wallach, H.}, \bibinfo{editor}{Larochelle, H.},
  \bibinfo{editor}{Grauman, K.}, \bibinfo{editor}{Cesa-Bianchi, N.},
  \bibinfo{editor}{Garnett, R.} (Eds.), \bibinfo{booktitle}{Advances in Neural
  Information Processing Systems}, \bibinfo{publisher}{Curran Associates, Inc.}
\newblock \URLprefix
  \url{https://proceedings.neurips.cc/paper/2018/file/285e19f20beded7d215102b49d5c09a0-Paper.pdf}.
\bibitem[{Hitchcock(2017)}]{hitchcock2017fallacies}
\bibinfo{author}{Hitchcock, D.}, \bibinfo{year}{2017}.
\newblock \bibinfo{title}{Do the fallacies have a place in the teaching of
  reasoning skills or critical thinking?}, in: \bibinfo{booktitle}{On reasoning
  and argument}. \bibinfo{publisher}{Springer}, pp. \bibinfo{pages}{401--408}.
\bibitem[{Ilievski et~al.(2021a)Ilievski, Oltramari, Ma, Zhang, McGuinness and
  Szekely}]{ilievski2021dimensions}
\bibinfo{author}{Ilievski, F.}, \bibinfo{author}{Oltramari, A.},
  \bibinfo{author}{Ma, K.}, \bibinfo{author}{Zhang, B.},
  \bibinfo{author}{McGuinness, D.L.}, \bibinfo{author}{Szekely, P.},
  \bibinfo{year}{2021}a.
\newblock \bibinfo{title}{Dimensions of commonsense knowledge}.
\newblock \bibinfo{journal}{Knowledge-Based Systems} \bibinfo{volume}{229},
  \bibinfo{pages}{107347}.
\bibitem[{Ilievski et~al.(2021b)Ilievski, Szekely and Zhang}]{ilievski2021cskg}
\bibinfo{author}{Ilievski, F.}, \bibinfo{author}{Szekely, P.},
  \bibinfo{author}{Zhang, B.}, \bibinfo{year}{2021}b.
\newblock \bibinfo{title}{Cskg: The commonsense knowledge graph}, in:
  \bibinfo{booktitle}{European Semantic Web Conference},
  \bibinfo{organization}{Springer}. pp. \bibinfo{pages}{680--696}.
\bibitem[{Jiang et~al.(2020)Jiang, Metzger, Flanagin and Wilson}]{Jiang2020}
\bibinfo{author}{Jiang, S.}, \bibinfo{author}{Metzger, M.},
  \bibinfo{author}{Flanagin, A.}, \bibinfo{author}{Wilson, C.},
  \bibinfo{year}{2020}.
\newblock \bibinfo{title}{Modeling and measuring expressed (dis)belief in
  (mis)information}.
\newblock \bibinfo{journal}{Proceedings of the International AAAI Conference on
  Web and Social Media} \bibinfo{volume}{14}, \bibinfo{pages}{315--326}.
\newblock \URLprefix
  \url{https://ojs.aaai.org/index.php/ICWSM/article/view/7302}.
\bibitem[{Jin et~al.(2022)Jin, Lalwani, Vaidhya, Shen, Ding, Lyu, Sachan,
  Mihalcea and Schölkopf}]{logical_fallacy_main_paper}
\bibinfo{author}{Jin, Z.}, \bibinfo{author}{Lalwani, A.},
  \bibinfo{author}{Vaidhya, T.}, \bibinfo{author}{Shen, X.},
  \bibinfo{author}{Ding, Y.}, \bibinfo{author}{Lyu, Z.},
  \bibinfo{author}{Sachan, M.}, \bibinfo{author}{Mihalcea, R.},
  \bibinfo{author}{Schölkopf, B.}, \bibinfo{year}{2022}.
\newblock \bibinfo{title}{Logical fallacy detection}.
\newblock \URLprefix \url{https://arxiv.org/abs/2202.13758},
  \DOIprefix\doi{10.48550/ARXIV.2202.13758}.
\bibitem[{Johansen and Kruschke(2005)}]{Johansen2005-nu}
\bibinfo{author}{Johansen, M.K.}, \bibinfo{author}{Kruschke, J.K.},
  \bibinfo{year}{2005}.
\newblock \bibinfo{title}{Category representation for classification and
  feature inference}.
\newblock \bibinfo{journal}{J. Exp. Psychol. Learn. Mem. Cogn.}
  \bibinfo{volume}{31}, \bibinfo{pages}{1433--1458}.
\bibitem[{Khan(2021)}]{khan_2021}
\bibinfo{author}{Khan, I.}, \bibinfo{year}{2021}.
\newblock \bibinfo{title}{Disinformation and freedom of opinion and
  expression}.
\newblock \URLprefix
  \url{https://documents-dds-ny.un.org/doc/UNDOC/GEN/G21/085/64/PDF/G2108564.pdf}.
  \bibinfo{note}{undocs.org/en/A/HRC/47/25. Accessed: 2022-11-30}.
\bibitem[{Kiesel et~al.(2019)Kiesel, Mestre, Shukla, Vincent, Adineh, Corney,
  Stein and Potthast}]{kiesel-etal-2019-semeval}
\bibinfo{author}{Kiesel, J.}, \bibinfo{author}{Mestre, M.},
  \bibinfo{author}{Shukla, R.}, \bibinfo{author}{Vincent, E.},
  \bibinfo{author}{Adineh, P.}, \bibinfo{author}{Corney, D.},
  \bibinfo{author}{Stein, B.}, \bibinfo{author}{Potthast, M.},
  \bibinfo{year}{2019}.
\newblock \bibinfo{title}{{S}em{E}val-2019 task 4: Hyperpartisan news
  detection}, in: \bibinfo{booktitle}{Proceedings of the 13th International
  Workshop on Semantic Evaluation}, \bibinfo{publisher}{Association for
  Computational Linguistics}, \bibinfo{address}{Minneapolis, Minnesota, USA}.
  pp. \bibinfo{pages}{829--839}.
\newblock \URLprefix \url{https://aclanthology.org/S19-2145},
  \DOIprefix\doi{10.18653/v1/S19-2145}.
\bibitem[{Lakkaraju et~al.(2016)Lakkaraju, Bach and
  Leskovec}]{10.1145/2939672.2939874}
\bibinfo{author}{Lakkaraju, H.}, \bibinfo{author}{Bach, S.H.},
  \bibinfo{author}{Leskovec, J.}, \bibinfo{year}{2016}.
\newblock \bibinfo{title}{Interpretable decision sets: A joint framework for
  description and prediction}, in: \bibinfo{booktitle}{Proceedings of the 22nd
  ACM SIGKDD International Conference on Knowledge Discovery and Data Mining},
  \bibinfo{publisher}{Association for Computing Machinery},
  \bibinfo{address}{New York, NY, USA}. p. \bibinfo{pages}{1675–1684}.
\newblock \URLprefix \url{https://doi.org/10.1145/2939672.2939874},
  \DOIprefix\doi{10.1145/2939672.2939874}.
\bibitem[{Lazer et~al.(2018)Lazer, Baum, Benkler, Berinsky, Greenhill, Menczer,
  Metzger, Nyhan, Pennycook, Rothschild et~al.}]{lazer2018science}
\bibinfo{author}{Lazer, D.M.}, \bibinfo{author}{Baum, M.A.},
  \bibinfo{author}{Benkler, Y.}, \bibinfo{author}{Berinsky, A.J.},
  \bibinfo{author}{Greenhill, K.M.}, \bibinfo{author}{Menczer, F.},
  \bibinfo{author}{Metzger, M.J.}, \bibinfo{author}{Nyhan, B.},
  \bibinfo{author}{Pennycook, G.}, \bibinfo{author}{Rothschild, D.}, et~al.,
  \bibinfo{year}{2018}.
\newblock \bibinfo{title}{The science of fake news}.
\newblock \bibinfo{journal}{Science} \bibinfo{volume}{359},
  \bibinfo{pages}{1094--1096}.
\bibitem[{Lewis et~al.(2019)Lewis, Liu, Goyal, Ghazvininejad, Mohamed, Levy,
  Stoyanov and Zettlemoyer}]{lewis2019bart}
\bibinfo{author}{Lewis, M.}, \bibinfo{author}{Liu, Y.}, \bibinfo{author}{Goyal,
  N.}, \bibinfo{author}{Ghazvininejad, M.}, \bibinfo{author}{Mohamed, A.},
  \bibinfo{author}{Levy, O.}, \bibinfo{author}{Stoyanov, V.},
  \bibinfo{author}{Zettlemoyer, L.}, \bibinfo{year}{2019}.
\newblock \bibinfo{title}{Bart: Denoising sequence-to-sequence pre-training for
  natural language generation, translation, and comprehension}.
\newblock \bibinfo{journal}{arXiv preprint arXiv:1910.13461} .
\bibitem[{Li et~al.(2017)Li, Liu, Chen and
  Rudin}]{https://doi.org/10.48550/arxiv.1710.04806}
\bibinfo{author}{Li, O.}, \bibinfo{author}{Liu, H.}, \bibinfo{author}{Chen,
  C.}, \bibinfo{author}{Rudin, C.}, \bibinfo{year}{2017}.
\newblock \bibinfo{title}{Deep learning for case-based reasoning through
  prototypes: A neural network that explains its predictions}.
\newblock \URLprefix \url{https://arxiv.org/abs/1710.04806},
  \DOIprefix\doi{10.48550/ARXIV.1710.04806}.
\bibitem[{Lin et~al.(2019)Lin, Chen, Chen and Ren}]{lin-etal-2019-kagnet}
\bibinfo{author}{Lin, B.Y.}, \bibinfo{author}{Chen, X.}, \bibinfo{author}{Chen,
  J.}, \bibinfo{author}{Ren, X.}, \bibinfo{year}{2019}.
\newblock \bibinfo{title}{{K}ag{N}et: Knowledge-aware graph networks for
  commonsense reasoning}, in: \bibinfo{booktitle}{Proc. of EMNLP-IJCNLP}, pp.
  \bibinfo{pages}{2829--2839}.
\bibitem[{Liu et~al.(2019a)Liu, Zhou, Zhao, Wang, Ju, Deng and Wang}]{k-bert}
\bibinfo{author}{Liu, W.}, \bibinfo{author}{Zhou, P.}, \bibinfo{author}{Zhao,
  Z.}, \bibinfo{author}{Wang, Z.}, \bibinfo{author}{Ju, Q.},
  \bibinfo{author}{Deng, H.}, \bibinfo{author}{Wang, P.},
  \bibinfo{year}{2019}a.
\newblock \bibinfo{title}{K-bert: Enabling language representation with
  knowledge graph}.
\newblock \URLprefix \url{https://arxiv.org/abs/1909.07606},
  \DOIprefix\doi{10.48550/ARXIV.1909.07606}.
\bibitem[{Liu et~al.(2019b)Liu, Ott, Goyal, Du, Joshi, Chen, Levy, Lewis,
  Zettlemoyer and Stoyanov}]{https://doi.org/10.48550/arxiv.1907.11692}
\bibinfo{author}{Liu, Y.}, \bibinfo{author}{Ott, M.}, \bibinfo{author}{Goyal,
  N.}, \bibinfo{author}{Du, J.}, \bibinfo{author}{Joshi, M.},
  \bibinfo{author}{Chen, D.}, \bibinfo{author}{Levy, O.},
  \bibinfo{author}{Lewis, M.}, \bibinfo{author}{Zettlemoyer, L.},
  \bibinfo{author}{Stoyanov, V.}, \bibinfo{year}{2019}b.
\newblock \bibinfo{title}{Roberta: A robustly optimized bert pretraining
  approach}.
\newblock \URLprefix \url{https://arxiv.org/abs/1907.11692},
  \DOIprefix\doi{10.48550/ARXIV.1907.11692}.
\bibitem[{Locke(1997)}]{Locke1997-co}
\bibinfo{author}{Locke, J.}, \bibinfo{year}{1997}.
\newblock \bibinfo{title}{An Essay Concerning Human Understanding}.
\newblock Penguin classics, \bibinfo{publisher}{Penguin Classics},
  \bibinfo{address}{London, England}.
\bibitem[{Luceri et~al.(2020)Luceri, Giordano and Ferrara}]{Luceri2020}
\bibinfo{author}{Luceri, L.}, \bibinfo{author}{Giordano, S.},
  \bibinfo{author}{Ferrara, E.}, \bibinfo{year}{2020}.
\newblock \bibinfo{title}{Detecting troll behavior via inverse reinforcement
  learning: A case study of russian trolls in the 2016 us election}.
\newblock \bibinfo{journal}{Proceedings of the International AAAI Conference on
  Web and Social Media} \bibinfo{volume}{14}, \bibinfo{pages}{417--427}.
\newblock \URLprefix
  \url{https://ojs.aaai.org/index.php/ICWSM/article/view/7311}.
\bibitem[{Ma(2019)}]{nlpaug}
\bibinfo{author}{Ma, E.}, \bibinfo{year}{2019}.
\newblock \bibinfo{title}{Nlp augmentation}.
\newblock \bibinfo{howpublished}{https://github.com/makcedward/nlpaug}.
\bibitem[{Ma et~al.(2019)Ma, Francis, Lu, Nyberg and
  Oltramari}]{ma-etal-2019-towards}
\bibinfo{author}{Ma, K.}, \bibinfo{author}{Francis, J.}, \bibinfo{author}{Lu,
  Q.}, \bibinfo{author}{Nyberg, E.}, \bibinfo{author}{Oltramari, A.},
  \bibinfo{year}{2019}.
\newblock \bibinfo{title}{Towards generalizable neuro-symbolic systems for
  commonsense question answering}, in: \bibinfo{booktitle}{Proc. of the First
  Workshop on Commonsense Inference in Natural Language Processing}, pp.
  \bibinfo{pages}{22--32}.
\bibitem[{Ma et~al.(2021a)Ma, Ilievski, Francis, Bisk, Nyberg and
  Oltramari}]{ma2021knowledge}
\bibinfo{author}{Ma, K.}, \bibinfo{author}{Ilievski, F.},
  \bibinfo{author}{Francis, J.}, \bibinfo{author}{Bisk, Y.},
  \bibinfo{author}{Nyberg, E.}, \bibinfo{author}{Oltramari, A.},
  \bibinfo{year}{2021}a.
\newblock \bibinfo{title}{Knowledge-driven data construction for zero-shot
  evaluation in commonsense question answering}, in: \bibinfo{booktitle}{AAAI}.
\bibitem[{Ma et~al.(2021b)Ma, Ilievski, Francis, Ozaki, Nyberg and
  Oltramari}]{https://doi.org/10.48550/arxiv.2109.02837}
\bibinfo{author}{Ma, K.}, \bibinfo{author}{Ilievski, F.},
  \bibinfo{author}{Francis, J.}, \bibinfo{author}{Ozaki, S.},
  \bibinfo{author}{Nyberg, E.}, \bibinfo{author}{Oltramari, A.},
  \bibinfo{year}{2021}b.
\newblock \bibinfo{title}{Exploring strategies for generalizable commonsense
  reasoning with pre-trained models}.
\newblock \URLprefix \url{https://arxiv.org/abs/2109.02837},
  \DOIprefix\doi{10.48550/ARXIV.2109.02837}.
\bibitem[{Marques-Silva(2008)}]{marques2008practical}
\bibinfo{author}{Marques-Silva, J.}, \bibinfo{year}{2008}.
\newblock \bibinfo{title}{Practical applications of boolean satisfiability} .
\bibitem[{Martino et~al.(2020)Martino, Cresci, Barron-Cedeno, Yu, Di~Pietro and
  Nakov}]{https://doi.org/10.48550/arxiv.2007.08024}
\bibinfo{author}{Martino, G.D.S.}, \bibinfo{author}{Cresci, S.},
  \bibinfo{author}{Barron-Cedeno, A.}, \bibinfo{author}{Yu, S.},
  \bibinfo{author}{Di~Pietro, R.}, \bibinfo{author}{Nakov, P.},
  \bibinfo{year}{2020}.
\newblock \bibinfo{title}{A survey on computational propaganda detection}
  \URLprefix \url{https://arxiv.org/abs/2007.08024},
  \DOIprefix\doi{10.48550/ARXIV.2007.08024}.
\bibitem[{Medin and Schaffer(1978)}]{Medin1978-je}
\bibinfo{author}{Medin, D.L.}, \bibinfo{author}{Schaffer, M.M.},
  \bibinfo{year}{1978}.
\newblock \bibinfo{title}{Context theory of classification learning}.
\newblock \bibinfo{journal}{Psychol. Rev.} \bibinfo{volume}{85},
  \bibinfo{pages}{207--238}.
\bibitem[{Miller(1995)}]{miller1995wordnet}
\bibinfo{author}{Miller, G.A.}, \bibinfo{year}{1995}.
\newblock \bibinfo{title}{Wordnet: a lexical database for english}.
\newblock \bibinfo{journal}{Communications of the ACM} \bibinfo{volume}{38},
  \bibinfo{pages}{39--41}.
\bibitem[{Mitra et~al.(2019)Mitra, Banerjee, Pal, Mishra and
  Baral}]{mitra2019exploring}
\bibinfo{author}{Mitra, A.}, \bibinfo{author}{Banerjee, P.},
  \bibinfo{author}{Pal, K.K.}, \bibinfo{author}{Mishra, S.},
  \bibinfo{author}{Baral, C.}, \bibinfo{year}{2019}.
\newblock \bibinfo{title}{Exploring ways to incorporate additional knowledge to
  improve natural language commonsense question answering}.
\newblock \bibinfo{journal}{arXiv preprint arXiv:1909.08855} .
\bibitem[{Morris et~al.(2020)Morris, Lifland, Yoo and Qi}]{bert-mnli}
\bibinfo{author}{Morris, J.}, \bibinfo{author}{Lifland, E.},
  \bibinfo{author}{Yoo, J.Y.}, \bibinfo{author}{Qi, Y.}, \bibinfo{year}{2020}.
\newblock \bibinfo{title}{Textattack/bert-base-uncased-mnli · hugging face}.
\newblock \URLprefix
  \url{https://huggingface.co/textattack/bert-base-uncased-MNLI}.
\bibitem[{Morrow et~al.(2022)Morrow, Swire-Thompson, Polny, Kopec and
  Wihbey}]{morrow2022emerging}
\bibinfo{author}{Morrow, G.}, \bibinfo{author}{Swire-Thompson, B.},
  \bibinfo{author}{Polny, J.M.}, \bibinfo{author}{Kopec, M.},
  \bibinfo{author}{Wihbey, J.P.}, \bibinfo{year}{2022}.
\newblock \bibinfo{title}{The emerging science of content labeling:
  Contextualizing social media content moderation}.
\newblock \bibinfo{journal}{Journal of the Association for Information Science
  and Technology} \bibinfo{volume}{73}, \bibinfo{pages}{1365--1386}.
\bibitem[{Morstatter et~al.(2019)Morstatter, Galstyan, Satyukov, Benjamin,
  Abeliuk, Mirtaheri, Hossain, Szekely, Ferrara, Matsui
  et~al.}]{morstatter2019sage}
\bibinfo{author}{Morstatter, F.}, \bibinfo{author}{Galstyan, A.},
  \bibinfo{author}{Satyukov, G.}, \bibinfo{author}{Benjamin, D.},
  \bibinfo{author}{Abeliuk, A.}, \bibinfo{author}{Mirtaheri, M.},
  \bibinfo{author}{Hossain, K.T.}, \bibinfo{author}{Szekely, P.A.},
  \bibinfo{author}{Ferrara, E.}, \bibinfo{author}{Matsui, A.}, et~al.,
  \bibinfo{year}{2019}.
\newblock \bibinfo{title}{Sage: A hybrid geopolitical event forecasting
  system.}, in: \bibinfo{booktitle}{IJCAI}, pp. \bibinfo{pages}{6557--6559}.
\bibitem[{Nakpih and Santini(2020)}]{Ireneous_Nakpih_2020}
\bibinfo{author}{Nakpih, C.I.}, \bibinfo{author}{Santini, S.},
  \bibinfo{year}{2020}.
\newblock \bibinfo{title}{Automated discovery of logical fallacies in legal
  argumentation}.
\newblock \bibinfo{journal}{International Journal of Artificial Intelligence \&
  Applications} \bibinfo{volume}{11}, \bibinfo{pages}{37--48}.
\newblock \URLprefix \url{https://doi.org/10.5121\%2Fijaia.2020.11203},
  \DOIprefix\doi{10.5121/ijaia.2020.11203}.
\bibitem[{Narang et~al.(2020)Narang, Raffel, Lee, Roberts, Fiedel and
  Malkan}]{https://doi.org/10.48550/arxiv.2004.14546}
\bibinfo{author}{Narang, S.}, \bibinfo{author}{Raffel, C.},
  \bibinfo{author}{Lee, K.}, \bibinfo{author}{Roberts, A.},
  \bibinfo{author}{Fiedel, N.}, \bibinfo{author}{Malkan, K.},
  \bibinfo{year}{2020}.
\newblock \bibinfo{title}{Wt5?! training text-to-text models to explain their
  predictions}.
\newblock \URLprefix \url{https://arxiv.org/abs/2004.14546},
  \DOIprefix\doi{10.48550/ARXIV.2004.14546}.
\bibitem[{Ng et~al.(2019)Ng, Yee, Baevski, Ott, Auli and
  Edunov}]{backtranslation}
\bibinfo{author}{Ng, N.}, \bibinfo{author}{Yee, K.}, \bibinfo{author}{Baevski,
  A.}, \bibinfo{author}{Ott, M.}, \bibinfo{author}{Auli, M.},
  \bibinfo{author}{Edunov, S.}, \bibinfo{year}{2019}.
\newblock \bibinfo{title}{Facebook fair's wmt19 news translation task
  submission}, in: \bibinfo{booktitle}{Proc. of WMT}.
\bibitem[{Oliinyk et~al.(2020)Oliinyk, Vysotska, Burov, Mykich and
  Fernandes}]{oliinyk2020propaganda}
\bibinfo{author}{Oliinyk, V.A.}, \bibinfo{author}{Vysotska, V.},
  \bibinfo{author}{Burov, Y.}, \bibinfo{author}{Mykich, K.},
  \bibinfo{author}{Fernandes, V.B.}, \bibinfo{year}{2020}.
\newblock \bibinfo{title}{Propaganda detection in text data based on nlp and
  machine learning.}, in: \bibinfo{booktitle}{MoMLeT+ DS}, pp.
  \bibinfo{pages}{132--144}.
\bibitem[{Oyelade and Ezugwu(2020)}]{OYELADE2020100395}
\bibinfo{author}{Oyelade, O.N.}, \bibinfo{author}{Ezugwu, A.E.},
  \bibinfo{year}{2020}.
\newblock \bibinfo{title}{A case-based reasoning framework for early detection
  and diagnosis of novel coronavirus}.
\newblock \bibinfo{journal}{Informatics in Medicine Unlocked}
  \bibinfo{volume}{20}, \bibinfo{pages}{100395}.
\newblock \URLprefix
  \url{https://www.sciencedirect.com/science/article/pii/S2352914820303683},
  \DOIprefix\doi{https://doi.org/10.1016/j.imu.2020.100395}.
\bibitem[{Pantazi et~al.(2004)Pantazi, Arocha and Moehr}]{pantazi2004case}
\bibinfo{author}{Pantazi, S.V.}, \bibinfo{author}{Arocha, J.F.},
  \bibinfo{author}{Moehr, J.R.}, \bibinfo{year}{2004}.
\newblock \bibinfo{title}{Case-based medical informatics}.
\newblock \bibinfo{journal}{BMC Medical Informatics and Decision Making}
  \bibinfo{volume}{4}, \bibinfo{pages}{1--23}.
\bibitem[{Paraschiv et~al.(2020)Paraschiv, Cercel and
  Dascalu}]{https://doi.org/10.48550/arxiv.2009.05289}
\bibinfo{author}{Paraschiv, A.}, \bibinfo{author}{Cercel, D.C.},
  \bibinfo{author}{Dascalu, M.}, \bibinfo{year}{2020}.
\newblock \bibinfo{title}{Upb at semeval-2020 task 11: Propaganda detection
  with domain-specific trained bert}.
\newblock \URLprefix \url{https://arxiv.org/abs/2009.05289},
  \DOIprefix\doi{10.48550/ARXIV.2009.05289}.
\bibitem[{Peters et~al.(2019)Peters, Neumann, Logan, Schwartz, Joshi, Singh and
  Smith}]{peters-etal-2019-knowledge}
\bibinfo{author}{Peters, M.E.}, \bibinfo{author}{Neumann, M.},
  \bibinfo{author}{Logan, R.}, \bibinfo{author}{Schwartz, R.},
  \bibinfo{author}{Joshi, V.}, \bibinfo{author}{Singh, S.},
  \bibinfo{author}{Smith, N.A.}, \bibinfo{year}{2019}.
\newblock \bibinfo{title}{Knowledge enhanced contextual word representations},
  in: \bibinfo{booktitle}{Proc. of EMNLP-IJCNLP}, pp. \bibinfo{pages}{43--54}.
\bibitem[{QIN and REGLI(2003)}]{qin_regli_2003}
\bibinfo{author}{QIN, X.}, \bibinfo{author}{REGLI, W.C.}, \bibinfo{year}{2003}.
\newblock \bibinfo{title}{A study in applying case-based reasoning to
  engineering design: Mechanical bearing design}.
\newblock \bibinfo{journal}{Artificial Intelligence for Engineering Design,
  Analysis and Manufacturing} \bibinfo{volume}{17}, \bibinfo{pages}{235–252}.
\newblock \DOIprefix\doi{10.1017/S0890060403173064}.
\bibitem[{Raaheim(1965)}]{10.2307/1165776}
\bibinfo{author}{Raaheim, K.}, \bibinfo{year}{1965}.
\newblock \bibinfo{title}{Problem solving and past experience}.
\newblock \bibinfo{journal}{Monographs of the Society for Research in Child
  Development} \bibinfo{volume}{30}, \bibinfo{pages}{58--67}.
\newblock \URLprefix \url{http://www.jstor.org/stable/1165776}.
\bibitem[{Reimers(2021a)}]{deberta-nli}
\bibinfo{author}{Reimers, N.}, \bibinfo{year}{2021}a.
\newblock \bibinfo{title}{Cross-encoder/nli-deberta-base · hugging face}.
\newblock \URLprefix
  \url{https://huggingface.co/cross-encoder/nli-deberta-base}.
\bibitem[{Reimers(2021b)}]{reimers_2021}
\bibinfo{author}{Reimers, N.}, \bibinfo{year}{2021}b.
\newblock \bibinfo{title}{Cross-encoder/nli-roberta-base · hugging face}.
\newblock \URLprefix
  \url{https://huggingface.co/cross-encoder/nli-roberta-base}.
\bibitem[{Reimers and Gurevych(2019)}]{reimers-2019-sentence-bert}
\bibinfo{author}{Reimers, N.}, \bibinfo{author}{Gurevych, I.},
  \bibinfo{year}{2019}.
\newblock \bibinfo{title}{Sentence-bert: Sentence embeddings using siamese
  bert-networks}, in: \bibinfo{booktitle}{Proceedings of the 2019 Conference on
  Empirical Methods in Natural Language Processing},
  \bibinfo{publisher}{Association for Computational Linguistics}.
\newblock \URLprefix \url{http://arxiv.org/abs/1908.10084}.
\bibitem[{Renkl(2014)}]{https://doi.org/10.1111/cogs.12086}
\bibinfo{author}{Renkl, A.}, \bibinfo{year}{2014}.
\newblock \bibinfo{title}{Toward an instructionally oriented theory of
  example-based learning}.
\newblock \bibinfo{journal}{Cognitive Science} \bibinfo{volume}{38},
  \bibinfo{pages}{1--37}.
\newblock \URLprefix
  \url{https://onlinelibrary.wiley.com/doi/abs/10.1111/cogs.12086},
  \DOIprefix\doi{https://doi.org/10.1111/cogs.12086}.
\bibitem[{Ribeiro et~al.(2016)Ribeiro, Singh and
  Guestrin}]{https://doi.org/10.48550/arxiv.1602.04938}
\bibinfo{author}{Ribeiro, M.T.}, \bibinfo{author}{Singh, S.},
  \bibinfo{author}{Guestrin, C.}, \bibinfo{year}{2016}.
\newblock \bibinfo{title}{"why should i trust you?": Explaining the predictions
  of any classifier}.
\newblock \URLprefix \url{https://arxiv.org/abs/1602.04938},
  \DOIprefix\doi{10.48550/ARXIV.1602.04938}.
\bibitem[{Rohde and Plaut(1999)}]{Rohde1999}
\bibinfo{author}{Rohde, D.L.}, \bibinfo{author}{Plaut, D.C.},
  \bibinfo{year}{1999}.
\newblock \bibinfo{title}{Language acquisition in the absence of explicit
  negative evidence: how important is starting small?}
\newblock \bibinfo{journal}{Cognition} \bibinfo{volume}{72},
  \bibinfo{pages}{67--109}.
\newblock \URLprefix
  \url{https://www.sciencedirect.com/science/article/pii/S0010027799000311},
  \DOIprefix\doi{https://doi.org/10.1016/S0010-0277(99)00031-1}.
\bibitem[{Rohde and Plaut(2004)}]{rohde2004less}
\bibinfo{author}{Rohde, D.L.}, \bibinfo{author}{Plaut, D.C.},
  \bibinfo{year}{2004}.
\newblock \bibinfo{title}{Less is less in language acquisition}, in:
  \bibinfo{booktitle}{Connectionist models of development}.
  \bibinfo{publisher}{Psychology Press}, pp. \bibinfo{pages}{178--218}.
\bibitem[{Rosch(1973)}]{ROSCH1973328}
\bibinfo{author}{Rosch, E.H.}, \bibinfo{year}{1973}.
\newblock \bibinfo{title}{Natural categories}.
\newblock \bibinfo{journal}{Cognitive Psychology} \bibinfo{volume}{4},
  \bibinfo{pages}{328--350}.
\newblock \URLprefix
  \url{https://www.sciencedirect.com/science/article/pii/0010028573900170},
  \DOIprefix\doi{https://doi.org/10.1016/0010-0285(73)90017-0}.
\bibitem[{Saha et~al.(2021)Saha, Yadav, Bauer and Bansal}]{saha2021explagraphs}
\bibinfo{author}{Saha, S.}, \bibinfo{author}{Yadav, P.},
  \bibinfo{author}{Bauer, L.}, \bibinfo{author}{Bansal, M.},
  \bibinfo{year}{2021}.
\newblock \bibinfo{title}{{E}xpla{G}raphs: An explanation graph generation task
  for structured commonsense reasoning}, in: \bibinfo{booktitle}{Proceedings of
  the 2021 Conference on Empirical Methods in Natural Language Processing},
  \bibinfo{publisher}{Association for Computational Linguistics},
  \bibinfo{address}{Online and Punta Cana, Dominican Republic}. pp.
  \bibinfo{pages}{7716--7740}.
\newblock \URLprefix \url{https://aclanthology.org/2021.emnlp-main.609},
  \DOIprefix\doi{10.18653/v1/2021.emnlp-main.609}.
\bibitem[{Sahai et~al.(2021)Sahai, Balalau and
  Horincar}]{sahai-etal-2021-breaking}
\bibinfo{author}{Sahai, S.}, \bibinfo{author}{Balalau, O.},
  \bibinfo{author}{Horincar, R.}, \bibinfo{year}{2021}.
\newblock \bibinfo{title}{Breaking down the invisible wall of informal
  fallacies in online discussions}, in: \bibinfo{booktitle}{Proceedings of the
  59th Annual Meeting of the Association for Computational Linguistics and the
  11th International Joint Conference on Natural Language Processing (Volume 1:
  Long Papers)}, \bibinfo{publisher}{Association for Computational
  Linguistics}, \bibinfo{address}{Online}. pp. \bibinfo{pages}{644--657}.
\newblock \URLprefix \url{https://aclanthology.org/2021.acl-long.53},
  \DOIprefix\doi{10.18653/v1/2021.acl-long.53}.
\bibitem[{Sap et~al.(2019)Sap, Le~Bras, Allaway, Bhagavatula, Lourie, Rashkin,
  Roof, Smith and Choi}]{sap2019atomic}
\bibinfo{author}{Sap, M.}, \bibinfo{author}{Le~Bras, R.},
  \bibinfo{author}{Allaway, E.}, \bibinfo{author}{Bhagavatula, C.},
  \bibinfo{author}{Lourie, N.}, \bibinfo{author}{Rashkin, H.},
  \bibinfo{author}{Roof, B.}, \bibinfo{author}{Smith, N.A.},
  \bibinfo{author}{Choi, Y.}, \bibinfo{year}{2019}.
\newblock \bibinfo{title}{Atomic: An atlas of machine commonsense for if-then
  reasoning}, in: \bibinfo{booktitle}{Proceedings of the AAAI conference on
  artificial intelligence}, pp. \bibinfo{pages}{3027--3035}.
\bibitem[{Scheffer and Rubenfeld(2000)}]{scheffer2000consensus}
\bibinfo{author}{Scheffer, B.K.}, \bibinfo{author}{Rubenfeld, M.G.},
  \bibinfo{year}{2000}.
\newblock \bibinfo{title}{A consensus statement on critical thinking in
  nursing}.
\newblock \bibinfo{journal}{Journal of Nursing Education} \bibinfo{volume}{39},
  \bibinfo{pages}{352--359}.
\newblock \URLprefix
  \url{https://journals.healio.com/doi/abs/10.3928/0148-4834-20001101-06},
  \DOIprefix\doi{10.3928/0148-4834-20001101-06}.
\bibitem[{Sennrich et~al.(2016)Sennrich, Haddow and
  Birch}]{sennrich-etalbacktranslation}
\bibinfo{author}{Sennrich, R.}, \bibinfo{author}{Haddow, B.},
  \bibinfo{author}{Birch, A.}, \bibinfo{year}{2016}.
\newblock \bibinfo{title}{Improving neural machine translation models with
  monolingual data}, in: \bibinfo{booktitle}{Proceedings of the 54th Annual
  Meeting of the Association for Computational Linguistics (Volume 1: Long
  Papers)}, \bibinfo{publisher}{Association for Computational Linguistics},
  \bibinfo{address}{Berlin, Germany}. pp. \bibinfo{pages}{86--96}.
\newblock \URLprefix \url{https://aclanthology.org/P16-1009},
  \DOIprefix\doi{10.18653/v1/P16-1009}.
\bibitem[{Shannon(1949)}]{shannon1949communication}
\bibinfo{author}{Shannon, C.E.}, \bibinfo{year}{1949}.
\newblock \bibinfo{title}{Communication theory of secrecy systems}.
\newblock \bibinfo{journal}{The Bell system technical journal}
  \bibinfo{volume}{28}, \bibinfo{pages}{656--715}.
\bibitem[{Sourati et~al.(2023)Sourati, Ilievski, Sandlin and
  Mermoud}]{sourati-cbr}
\bibinfo{author}{Sourati, Z.}, \bibinfo{author}{Ilievski, F.},
  \bibinfo{author}{Sandlin, H.A.}, \bibinfo{author}{Mermoud, A.},
  \bibinfo{year}{2023}.
\newblock \bibinfo{title}{Case-based reasoning with language models for
  classification of logical fallacies}.
\bibitem[{Speer et~al.(2017)Speer, Chin and Havasi}]{speer2017conceptnet}
\bibinfo{author}{Speer, R.}, \bibinfo{author}{Chin, J.},
  \bibinfo{author}{Havasi, C.}, \bibinfo{year}{2017}.
\newblock \bibinfo{title}{Conceptnet 5.5: An open multilingual graph of general
  knowledge}, in: \bibinfo{booktitle}{Thirty-first AAAI conference on
  artificial intelligence}.
\bibitem[{Spensberger et~al.(2022)Spensberger, Kollar and
  Pankofer}]{spensberger2022effects}
\bibinfo{author}{Spensberger, F.}, \bibinfo{author}{Kollar, I.},
  \bibinfo{author}{Pankofer, S.}, \bibinfo{year}{2022}.
\newblock \bibinfo{title}{Effects of worked examples and external scripts on
  fallacy recognition skills: a randomized controlled trial}.
\newblock \bibinfo{journal}{Journal of Social Work Education}
  \bibinfo{volume}{58}, \bibinfo{pages}{622--639}.
\bibitem[{Tymbay(2022)}]{tymbay2022manipulative}
\bibinfo{author}{Tymbay, A.A.}, \bibinfo{year}{2022}.
\newblock \bibinfo{title}{Manipulative use of political headlines in western
  and russian online sources}.
\newblock \bibinfo{journal}{Discourse \& Communication} \bibinfo{volume}{16},
  \bibinfo{pages}{346--363}.
\newblock \URLprefix \url{https://doi.org/10.1177/17504813221101824},
  \DOIprefix\doi{10.1177/17504813221101824}.
\bibitem[{Vaswani et~al.(2017)Vaswani, Shazeer, Parmar, Uszkoreit, Jones,
  Gomez, Kaiser and Polosukhin}]{https://doi.org/10.48550/arxiv.1706.03762}
\bibinfo{author}{Vaswani, A.}, \bibinfo{author}{Shazeer, N.},
  \bibinfo{author}{Parmar, N.}, \bibinfo{author}{Uszkoreit, J.},
  \bibinfo{author}{Jones, L.}, \bibinfo{author}{Gomez, A.N.},
  \bibinfo{author}{Kaiser, L.}, \bibinfo{author}{Polosukhin, I.},
  \bibinfo{year}{2017}.
\newblock \bibinfo{title}{Attention is all you need}.
\newblock \URLprefix \url{https://arxiv.org/abs/1706.03762},
  \DOIprefix\doi{10.48550/ARXIV.1706.03762}.
\bibitem[{Vorakitphan et~al.(2021)Vorakitphan, Cabrio and
  Villata}]{vorakitphan-etal-2021-dont}
\bibinfo{author}{Vorakitphan, V.}, \bibinfo{author}{Cabrio, E.},
  \bibinfo{author}{Villata, S.}, \bibinfo{year}{2021}.
\newblock \bibinfo{title}{{``}don{'}t discuss{''}: Investigating semantic and
  argumentative features for supervised propagandist message detection and
  classification}, in: \bibinfo{booktitle}{Proceedings of the International
  Conference on Recent Advances in Natural Language Processing (RANLP 2021)},
  \bibinfo{publisher}{INCOMA Ltd.}, \bibinfo{address}{Held Online}. pp.
  \bibinfo{pages}{1498--1507}.
\newblock \URLprefix \url{https://aclanthology.org/2021.ranlp-1.168}.
\bibitem[{Walia et~al.(2019)Walia, Rana and Kansal}]{8776909}
\bibinfo{author}{Walia, H.}, \bibinfo{author}{Rana, A.},
  \bibinfo{author}{Kansal, V.}, \bibinfo{year}{2019}.
\newblock \bibinfo{title}{Case based interpretation model for word sense
  disambiguation in gurmukhi}, in: \bibinfo{booktitle}{2019 9th International
  Conference on Cloud Computing, Data Science \& Engineering (Confluence)}, pp.
  \bibinfo{pages}{359--364}.
\newblock \DOIprefix\doi{10.1109/CONFLUENCE.2019.8776909}.
\bibitem[{Wang et~al.(2021)Wang, Ilievski, Chen and
  Ren}]{https://doi.org/10.48550/arxiv.2106.11533}
\bibinfo{author}{Wang, P.}, \bibinfo{author}{Ilievski, F.},
  \bibinfo{author}{Chen, M.}, \bibinfo{author}{Ren, X.}, \bibinfo{year}{2021}.
\newblock \bibinfo{title}{Do language models perform generalizable commonsense
  inference?}
\newblock \URLprefix \url{https://arxiv.org/abs/2106.11533},
  \DOIprefix\doi{10.48550/ARXIV.2106.11533}.
\bibitem[{Wang et~al.(2020)Wang, Wei, Dong, Bao, Yang and
  Zhou}]{wang2020minilm}
\bibinfo{author}{Wang, W.}, \bibinfo{author}{Wei, F.}, \bibinfo{author}{Dong,
  L.}, \bibinfo{author}{Bao, H.}, \bibinfo{author}{Yang, N.},
  \bibinfo{author}{Zhou, M.}, \bibinfo{year}{2020}.
\newblock \bibinfo{title}{Minilm: Deep self-attention distillation for
  task-agnostic compression of pre-trained transformers}.
\newblock \href{http://arxiv.org/abs/2002.10957}{\tt arXiv:2002.10957}.
\bibitem[{Wang et~al.(2019)Wang, McKee, Torbica and
  Stuckler}]{wang2019systematic}
\bibinfo{author}{Wang, Y.}, \bibinfo{author}{McKee, M.},
  \bibinfo{author}{Torbica, A.}, \bibinfo{author}{Stuckler, D.},
  \bibinfo{year}{2019}.
\newblock \bibinfo{title}{Systematic literature review on the spread of
  health-related misinformation on social media}.
\newblock \bibinfo{journal}{Social Science \& Medicine} \bibinfo{volume}{240},
  \bibinfo{pages}{112552}.
\newblock \URLprefix
  \url{https://www.sciencedirect.com/science/article/pii/S0277953619305465},
  \DOIprefix\doi{https://doi.org/10.1016/j.socscimed.2019.112552}.
\bibitem[{Watts(2021)}]{Watts2021-an}
\bibinfo{author}{Watts, I.}, \bibinfo{year}{2021}.
\newblock \bibinfo{title}{Logic}.
\newblock \bibinfo{publisher}{Soli Deo Gloria Publications},
  \bibinfo{address}{Morgan, PA}.
\bibitem[{Wei et~al.(2022)Wei, Wang, Schuurmans, Bosma, Ichter, Xia, Chi, Le
  and Zhou}]{https://doi.org/10.48550/arxiv.2201.11903}
\bibinfo{author}{Wei, J.}, \bibinfo{author}{Wang, X.},
  \bibinfo{author}{Schuurmans, D.}, \bibinfo{author}{Bosma, M.},
  \bibinfo{author}{Ichter, B.}, \bibinfo{author}{Xia, F.},
  \bibinfo{author}{Chi, E.}, \bibinfo{author}{Le, Q.}, \bibinfo{author}{Zhou,
  D.}, \bibinfo{year}{2022}.
\newblock \bibinfo{title}{Chain of thought prompting elicits reasoning in large
  language models}.
\newblock \URLprefix \url{https://arxiv.org/abs/2201.11903},
  \DOIprefix\doi{10.48550/ARXIV.2201.11903}.
\bibitem[{Williams et~al.(2018)Williams, Nangia and Bowman}]{N18-1101}
\bibinfo{author}{Williams, A.}, \bibinfo{author}{Nangia, N.},
  \bibinfo{author}{Bowman, S.}, \bibinfo{year}{2018}.
\newblock \bibinfo{title}{A broad-coverage challenge corpus for sentence
  understanding through inference}, in: \bibinfo{booktitle}{Proceedings of the
  2018 Conference of the North American Chapter of the Association for
  Computational Linguistics: Human Language Technologies, Volume 1 (Long
  Papers)}, \bibinfo{publisher}{Association for Computational Linguistics}. pp.
  \bibinfo{pages}{1112--1122}.
\newblock \URLprefix \url{http://aclweb.org/anthology/N18-1101}.
\bibitem[{Wu et~al.(2019)Wu, Morstatter, Carley and Liu}]{wu2019misinformation}
\bibinfo{author}{Wu, L.}, \bibinfo{author}{Morstatter, F.},
  \bibinfo{author}{Carley, K.M.}, \bibinfo{author}{Liu, H.},
  \bibinfo{year}{2019}.
\newblock \bibinfo{title}{Misinformation in social media: Definition,
  manipulation, and detection}.
\newblock \bibinfo{journal}{SIGKDD Explor. Newsl.} \bibinfo{volume}{21},
  \bibinfo{pages}{80–90}.
\newblock \URLprefix \url{https://doi.org/10.1145/3373464.3373475},
  \DOIprefix\doi{10.1145/3373464.3373475}.
\bibitem[{Yaskorska et~al.(2013)Yaskorska, Budzynska and
  Kacprzak}]{Yaskorska2013}
\bibinfo{author}{Yaskorska, O.}, \bibinfo{author}{Budzynska, K.},
  \bibinfo{author}{Kacprzak, M.}, \bibinfo{year}{2013}.
\newblock \bibinfo{title}{Proving propositional tautologies in a natural
  dialogue}.
\newblock \bibinfo{journal}{Fundamenta Informaticae} \bibinfo{volume}{128},
  \bibinfo{pages}{239--253}.
\newblock \URLprefix \url{https://doi.org/10.3233/FI-2013-944},
  \DOIprefix\doi{10.3233/FI-2013-944}. \bibinfo{note}{1-2}.
\bibitem[{Yoosuf and Yang(2019)}]{yoosuf-yang-2019-fine}
\bibinfo{author}{Yoosuf, S.}, \bibinfo{author}{Yang, Y.}, \bibinfo{year}{2019}.
\newblock \bibinfo{title}{Fine-grained propaganda detection with fine-tuned
  {BERT}}, in: \bibinfo{booktitle}{Proceedings of the Second Workshop on
  Natural Language Processing for Internet Freedom: Censorship, Disinformation,
  and Propaganda}, \bibinfo{publisher}{Association for Computational
  Linguistics}, \bibinfo{address}{Hong Kong, China}. pp.
  \bibinfo{pages}{87--91}.
\newblock \URLprefix \url{https://aclanthology.org/D19-5011},
  \DOIprefix\doi{10.18653/v1/D19-5011}.
\bibitem[{Yu et~al.(2021)Yu, Martino, Mohtarami, Glass and
  Nakov}]{https://doi.org/10.48550/arxiv.2108.12802}
\bibinfo{author}{Yu, S.}, \bibinfo{author}{Martino, G.D.S.},
  \bibinfo{author}{Mohtarami, M.}, \bibinfo{author}{Glass, J.},
  \bibinfo{author}{Nakov, P.}, \bibinfo{year}{2021}.
\newblock \bibinfo{title}{Interpretable propaganda detection in news articles}
  \URLprefix \url{https://arxiv.org/abs/2108.12802},
  \DOIprefix\doi{10.48550/ARXIV.2108.12802}.
\bibitem[{Zheng et~al.(2016)Zheng, Song, Leung and
  Goodfellow}]{https://doi.org/10.48550/arxiv.1604.04326}
\bibinfo{author}{Zheng, S.}, \bibinfo{author}{Song, Y.},
  \bibinfo{author}{Leung, T.}, \bibinfo{author}{Goodfellow, I.},
  \bibinfo{year}{2016}.
\newblock \bibinfo{title}{Improving the robustness of deep neural networks via
  stability training}.
\newblock \URLprefix \url{https://arxiv.org/abs/1604.04326},
  \DOIprefix\doi{10.48550/ARXIV.1604.04326}.
\bibitem[{Zhong et~al.(2019)Zhong, Tang, Duan, Zhou, Wang and
  Yin}]{10.1007/978-3-030-32233-5_2}
\bibinfo{author}{Zhong, W.}, \bibinfo{author}{Tang, D.}, \bibinfo{author}{Duan,
  N.}, \bibinfo{author}{Zhou, M.}, \bibinfo{author}{Wang, J.},
  \bibinfo{author}{Yin, J.}, \bibinfo{year}{2019}.
\newblock \bibinfo{title}{Improving question answering by commonsense-based
  pre-training}, in: \bibinfo{booktitle}{Natural Language Processing and
  Chinese Computing: 8th CCF International Conference, NLPCC 2019, Dunhuang,
  China, October 9–14, 2019, Proceedings, Part I},
  \bibinfo{publisher}{Springer-Verlag}, \bibinfo{address}{Berlin, Heidelberg}.
  p. \bibinfo{pages}{16–28}.
\newblock \URLprefix \url{https://doi.org/10.1007/978-3-030-32233-5_2},
  \DOIprefix\doi{10.1007/978-3-030-32233-5_2}.

\end{thebibliography}

\end{document}